\documentclass[final]{cvpr}

\usepackage{times}
\usepackage{epsfig}
\usepackage{graphicx}
\usepackage{amsmath}
\usepackage{amssymb}
\usepackage{subcaption}
\usepackage{float}
\usepackage{commath}
\usepackage{booktabs}
\usepackage{pifont}

\usepackage{multirow,enumitem}
\usepackage[table,x11names,dvipsnames,table]{xcolor}
\definecolor{myorchid}{RGB}{150,10,30}

\usepackage[pagebackref=true,breaklinks=true,colorlinks,bookmarks=false]{hyperref}

\renewcommand{\l}{\mathbf{l}}
\renewcommand{\v}{\mathbf{v}}
\newcommand{\n}{\mathbf{n}}
\newcommand{\nest}{\tilde{\n}}
\newcommand{\h}{\mathbf{h}}
\renewcommand{\a}{\pmb{\alpha}}
\newcommand{\aest}{\tilde{\a}}
\newcommand{\spec}{\rho}
\newcommand{\specest}{\tilde{\rho}}
\newcommand{\Iest}{I}
\newcommand{\Iobs}{\hat{I}}

\begin{document}

\title{A Dark Flash Normal Camera}

\author{Zhihao Xia$^1$\thanks{Work done while Zhihao Xia was an intern at Google.}~~~~~~Jason Lawrence$^2$~~~~~~Supreeth Achar$^2$\\
  $^1$Washington University in St. Louis~~~~~~~$^2$Google Research\\
}

\maketitle

\begin{abstract}

Casual photography is often performed in uncontrolled lighting that can result in low quality images and degrade the performance of downstream processing. We consider the problem of estimating surface normal and reflectance maps of scenes depicting people despite these conditions by supplementing the available visible illumination with a single near infrared (NIR) light source and camera, a so-called ``dark flash image''. Our method takes as input a single color image captured under arbitrary visible lighting and a single dark flash image captured under controlled front-lit NIR lighting at the same viewpoint, and computes a normal map, a diffuse albedo map, and a specular intensity map of the scene. Since ground truth normal and reflectance maps of faces are difficult to capture, we propose a novel training technique that combines information from two readily available and complementary sources: a stereo depth signal and photometric shading cues. We evaluate our method over a range of subjects and lighting conditions and describe two applications: optimizing stereo geometry and filling the shadows in an image.

\end{abstract}

\section{Introduction}

In casual mobile photography, images are often captured under poor lighting conditions. Controlling the visible lighting or supplementing it with a flash is often too difficult or too disruptive to be practical. On the other hand, the near infrared (NIR) lighting in a scene can be much more easily controlled and is invisible to the user. In this paper, we demonstrate how a single ``dark flash'' NIR image and a single visible image taken under uncontrolled lighting can be used to recover high quality maps of the surface normals, diffuse albedos, and specular intensities in the scene. We collectively refer to the albedo and specular intensity estimates as a ``reflectance map''. Exposing these signals within a photography pipeline opens up a range of applications from refining independent depth estimates to digitally manipulating the lighting in the scene. Although our method is applicable to many types of objects, we focus on faces - the most common photography subject at the short ranges over which active illumination is effective. 

\begin{figure}
    \centering
    \begin{subfigure}[b]{0.325\columnwidth}
        \centering
        \includegraphics[width=\textwidth]{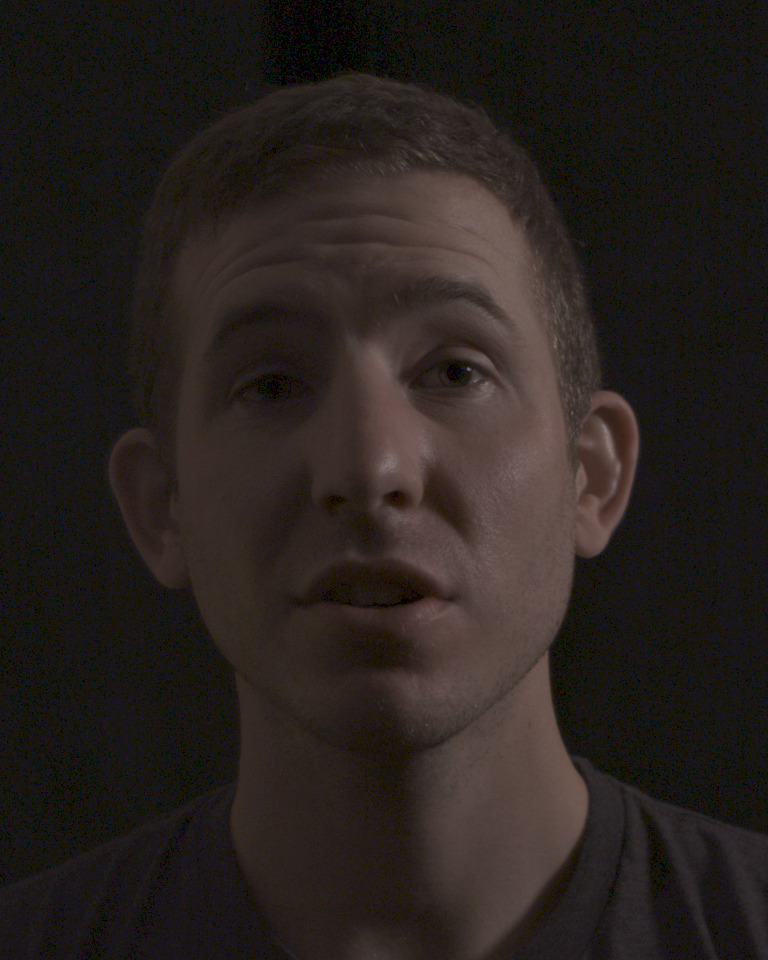} \\
        \includegraphics[width=\textwidth]{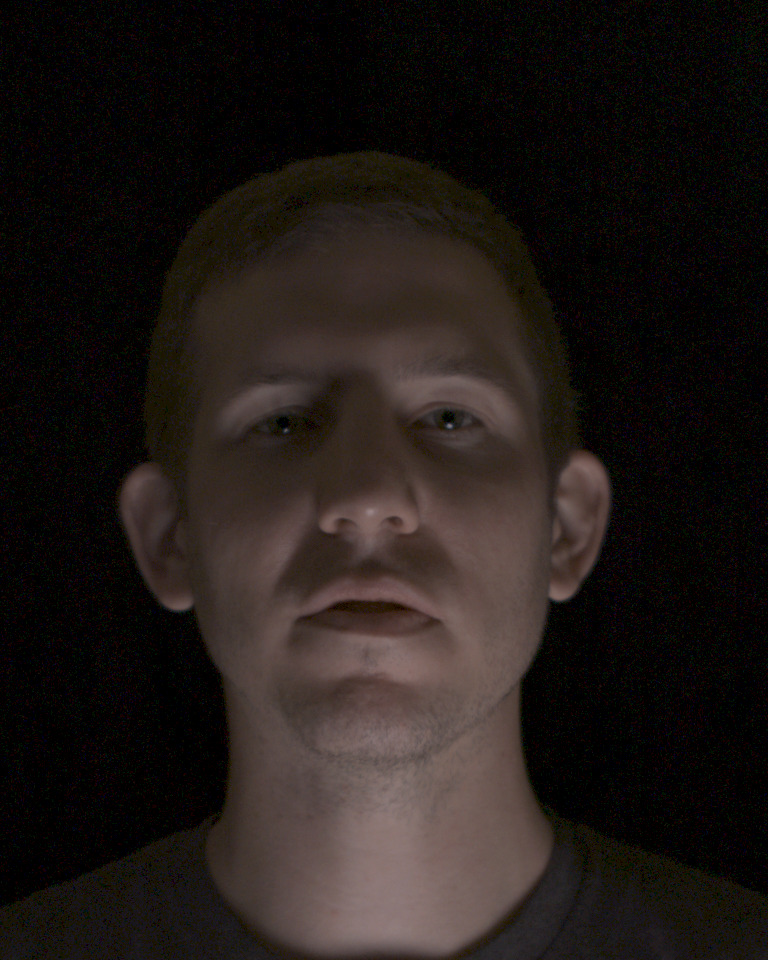} \\
        \includegraphics[width=\textwidth]{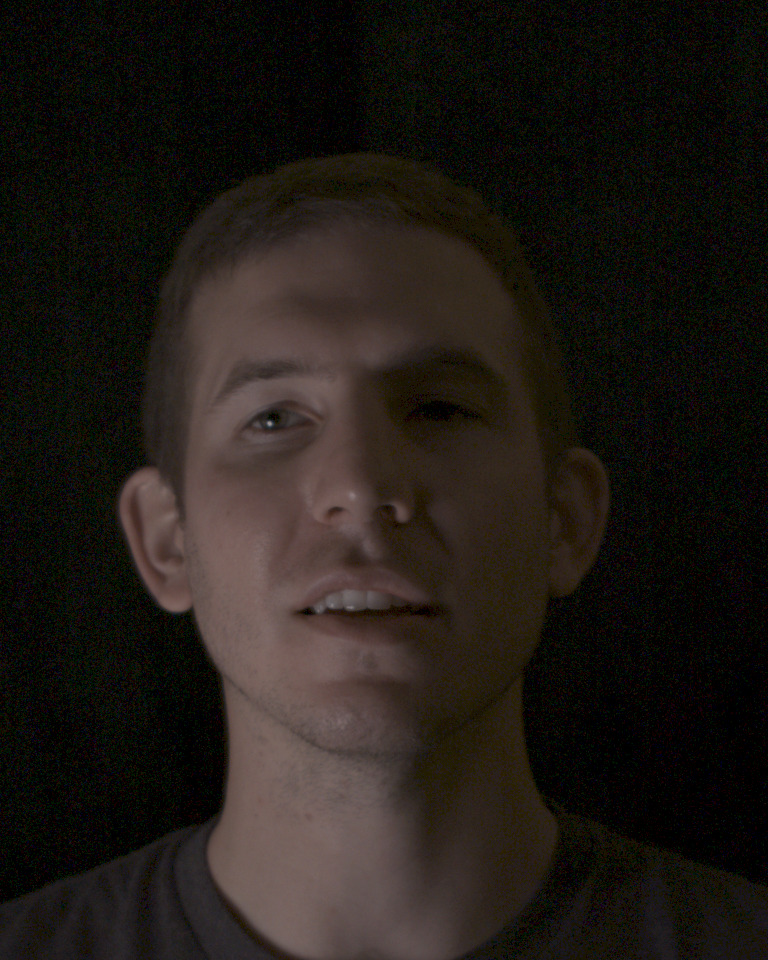}\vspace{-5pt}%
 		\caption*{\footnotesize{RGB Input}}
    \end{subfigure}
    \begin{subfigure}[b]{0.325\columnwidth}  
        \centering 
        \includegraphics[width=\textwidth]{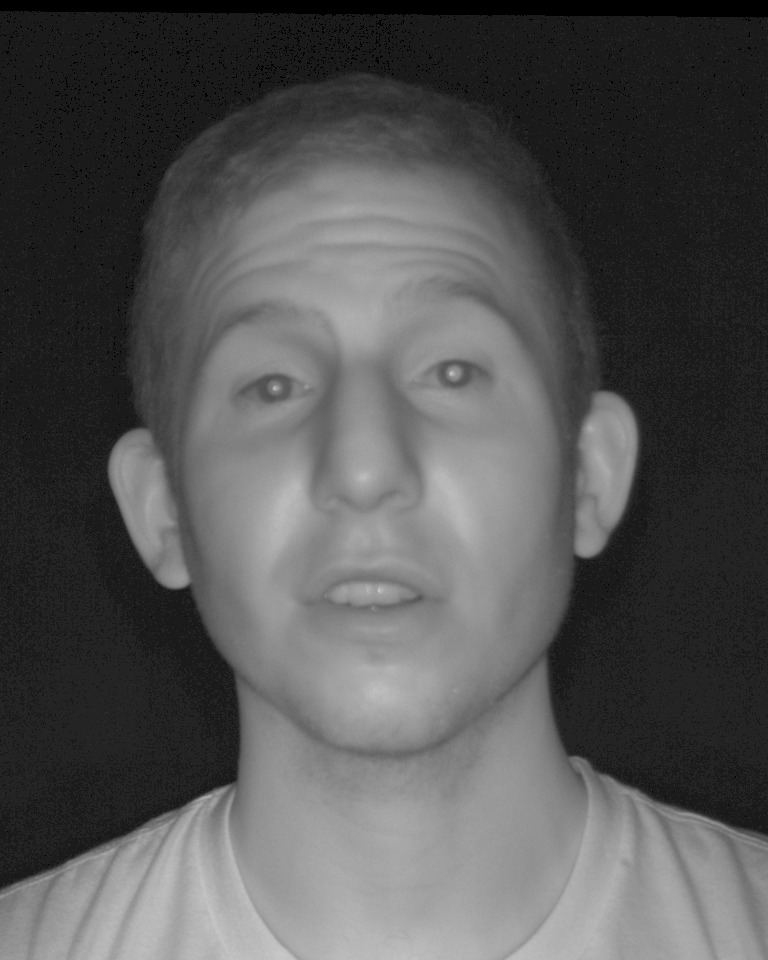} \\
        \includegraphics[width=\textwidth]{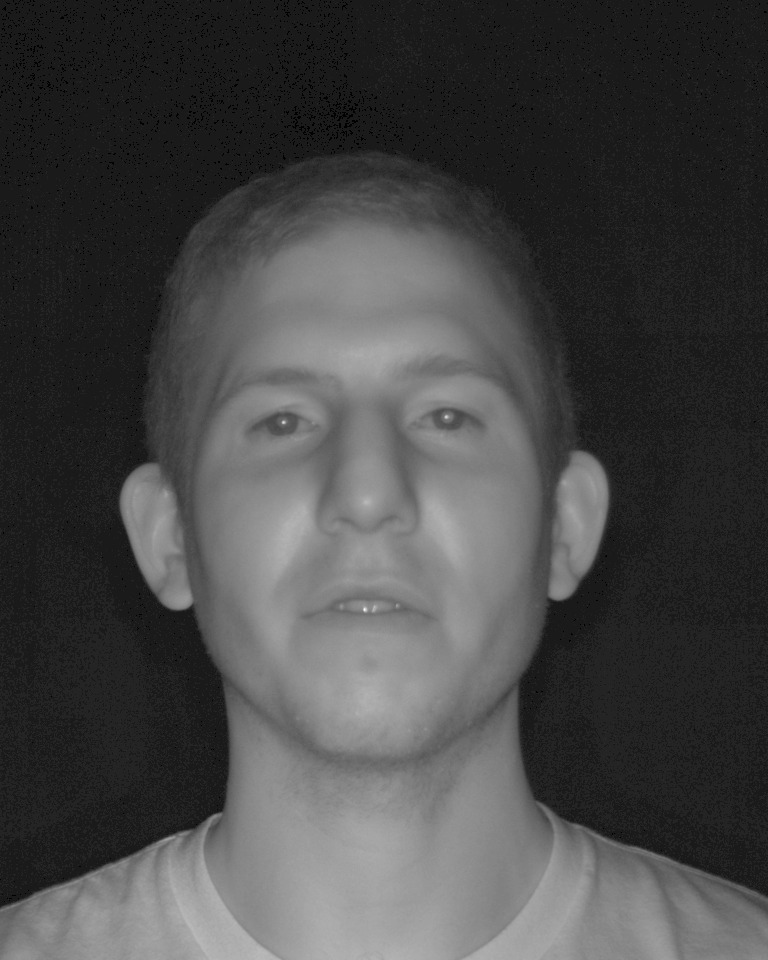} \\
        \includegraphics[width=\textwidth]{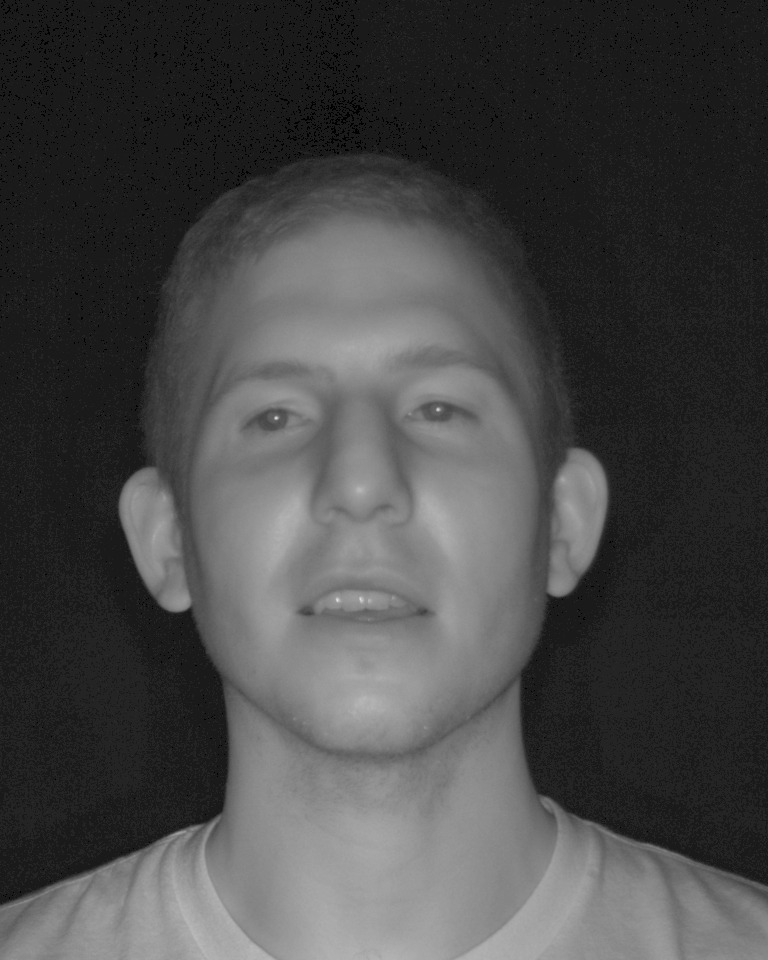}\vspace{-5pt}%
 		\caption*{\footnotesize{NIR Input}}
    \end{subfigure}
    \begin{subfigure}[b]{0.325\columnwidth}   
        \centering 
        \includegraphics[width=\textwidth]{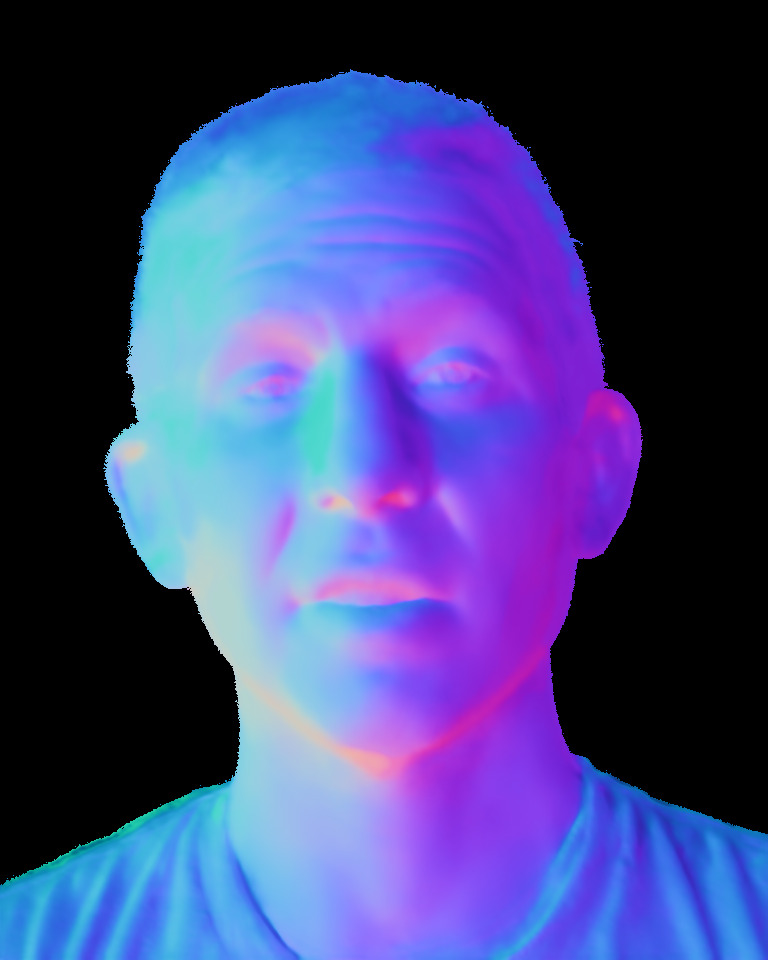} \\
        \includegraphics[width=\textwidth]{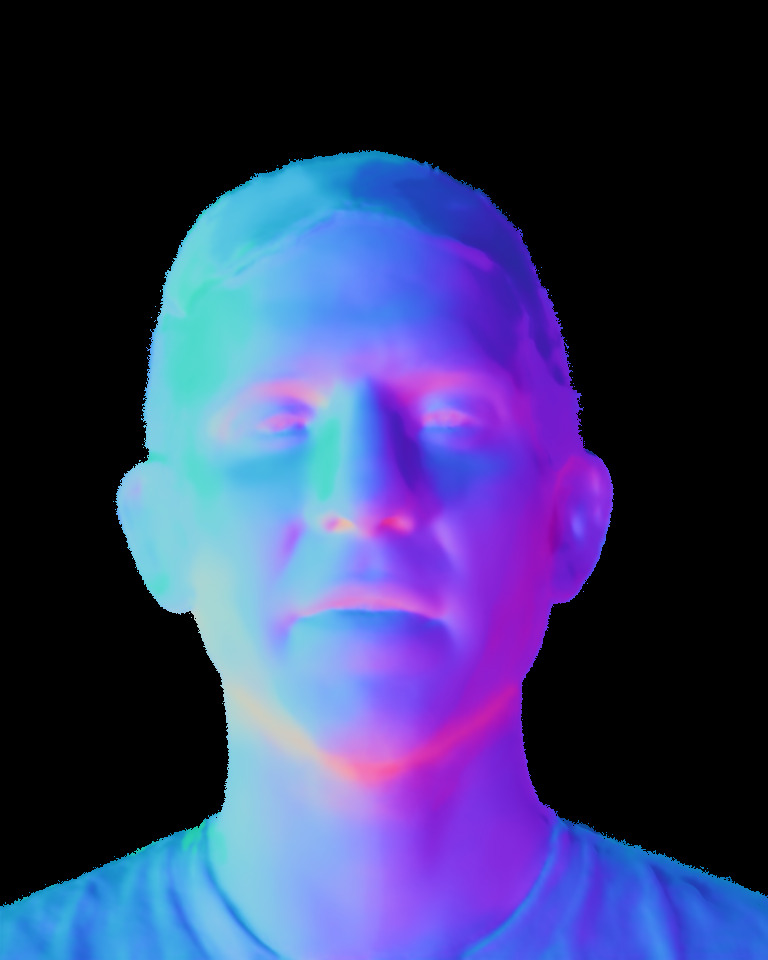} \\
        \includegraphics[width=\textwidth]{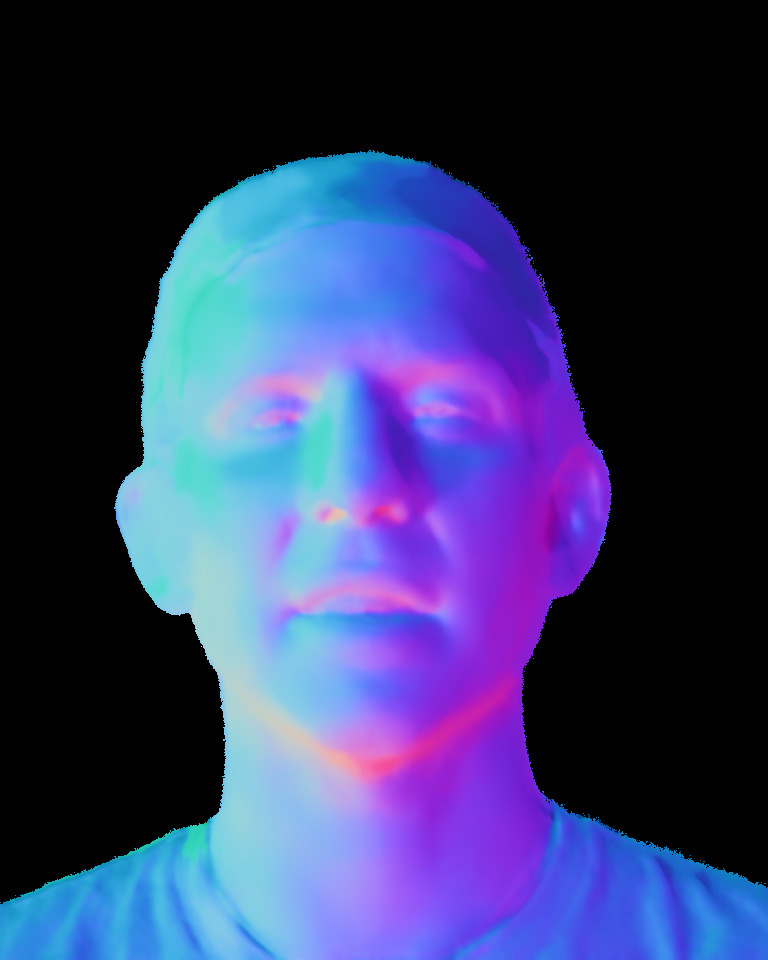}\vspace{-5pt}%
 		\caption*{\footnotesize{Normal Map}}
    \end{subfigure}
    \caption{Estimating surface geometry from a single RGB image is challenging. We augment this input with a single NIR ``dark flash'' image captured at the same time, and present a network that can estimate high quality normal maps and reflectance maps (not shown) under a wide range of visible lighting conditions.} 
    \label{fig:teaser}
\end{figure}

Our use of controlled NIR lighting provides a number of benefits.
First, the ambient NIR light in a scene is often weak or completely absent in indoor environments and is significantly attenuated by atmospheric absorption outdoors, making it practical to control this aspect of a scene. Second, it results in a more tractable estimation problem in contrast to single-image ``shape from shading'' and intrinsic image decomposition techniques that must simultaneously reason about shape, material properties, and lighting. Third, by locating the NIR light source near the camera we can minimize shadows in the scene which would otherwise complicate this estimation problem. This setup also produces specular highlights along surfaces that are nearly perpendicular to the viewing direction, giving a useful cue for determining surface orientations.

We present a deep neural network that takes as input one RGB image captured under uncontrolled visible lighting and one monochrome NIR image captured from the same viewpoint, but under controlled lighting provided by a single source located near the camera. The network generates a surface normal and reflectance estimate (diffuse albedo + specular intensity) at each pixel. We train this network by combining two imperfect but complementary cues: a stereo depth map that provides a reliable estimate of the low-frequency components of the scene's 3d shape along with photometric cues that convey higher-frequency geometric details. These measurements are far easier to obtain than ground truth geometry and appearance measurements. We explicitly model the specular reflectance of human skin in a photometric loss term that guides our training along with a prior on the albedo map that favors piecewise constant variation~\cite{Barron2015Sirfs}.

We compare our technique to a baseline learning approach that uses only a single RGB image as input and state-of-the-art methods for single image intrinsic image decomposition~\cite{Sengupta2018SFSNet} and relighting~\cite{Nestmeyer2020FaceRelighting}. We are able to produce overall more stable and more accurate outputs even in very challenging visible light conditions. We also present two applications of integrating our technique in a mobile photography pipeline. In all, this paper makes the following contributions:
\begin{itemize}
\item A new network architecture for estimating dense normal and reflectance maps from a single RGB+NIR image pair.
\item A new training strategy that combines two independent and complementary signals: one from stereo triangulation and the other from photometric cues in RGB and NIR, along with a hardware setup for collecting this data. Notably, our training is guided by a physically-based image formation model that reproduces both diffuse and surface reflectance.
\item We demonstrate two applications of our method in a modern photography pipeline: optimizing depths computed by an independent stereo technique and reducing shadows in an image post-capture.
\end{itemize}

\section{Related Work}

\noindent \textbf{Intrinsic imaging and shape from shading.} Decomposing a single image into its underlying shape and reflectance is a classical under-constrained problem in computer vision~\cite{Barrow1978Intrinsic,Horn1989ShapeShading}. Barron et al.~\cite{Barron2012Shape,Barron2015Sirfs} propose a solution, SIRFS, that applies carefully engineered priors to disambiguate these components.
Learning-based methods have been proposed more recently that 
train deep neural networks to perform this task using rendered datasets where ground truth is available~\cite{Shi2016LearningIntrinsics,Li2018ReconstructSVBRDF} or sparse human annotations
~\cite{Bell14IntrinsicWild}. Ye et al.~\cite{Yu2019InverseRenderNet} apply multiview stereo to images of buildings taken at different times and under different lighting conditions as a source of training data.
Whereas some learning approaches function as ``black boxes''~\cite{Shi2016LearningIntrinsics}, others apply a physically-based image formation model to the network outputs to form a reconstruction loss~\cite{Baslamisli2018CNNIntrinsic, Sengupta2018SFSNet} and others directly incorporate a rendering layer into the network~\cite{Li2018ReconstructSVBRDF,Tainai2018NeuralInverseRendering}. In work concurrent with ours, Qiu et al.~\cite{Qiu2020Towards} explore the use of a single visible flash image and a similar neural network architecture. Cheng et al.~\cite{Cheng2019Nonlocal} use an RGB+NIR image pair, although they don't attempt to control the NIR lighting in the scene as we do. Yoon et al.~\cite{Yoon2016FineScale} train a network that accepts a single NIR image using photometric normals computed from a multi-point-NIR-light dataset. In contrast, our network considers both a visible and front-lit NIR image and we use two independent signals to supervise the training.

A number of methods are specifically designed to work on images of faces. This includes 3D morphable models~\cite{Volker19993DMM}, which are commonly used as a prior on reflectance and geometry in learning-based approaches~\cite{Tewari2017MoFA, Shu2017NeuralFace, Sengupta2018SFSNet}. Sanyal et al.~\cite{Sanyal2019LearningToRegress} estimate the shape of a face within a single image in the form of blending weights over a parametric face model. Similar to our approach, other techniques estimate dense normal or displacement maps~\cite{Zeng2019DF2Net} including for faces partially hidden by occluders~\cite{Tran2018Extreme3DFace, Deng2019Accurate3DFace}. However these methods do not attempt to disentangle reflectance data from shading. Some portrait relighting techniques estimate surface normal and albedo maps as intermediate signals that guide a relighting process~\cite{Nestmeyer2020FaceRelighting}.

Distinct from these prior techniques is our use of a single front-lit NIR image in addition to a color input image, enabling our technique to perform well even in very harsh and low visible light conditions.
We compare to a baseline network that uses only a single RGB image as input in Section~\ref{subsection:Ablation}. \\

\begin{figure*}[!ht]
    \centering
    \includegraphics[width=0.96\textwidth]{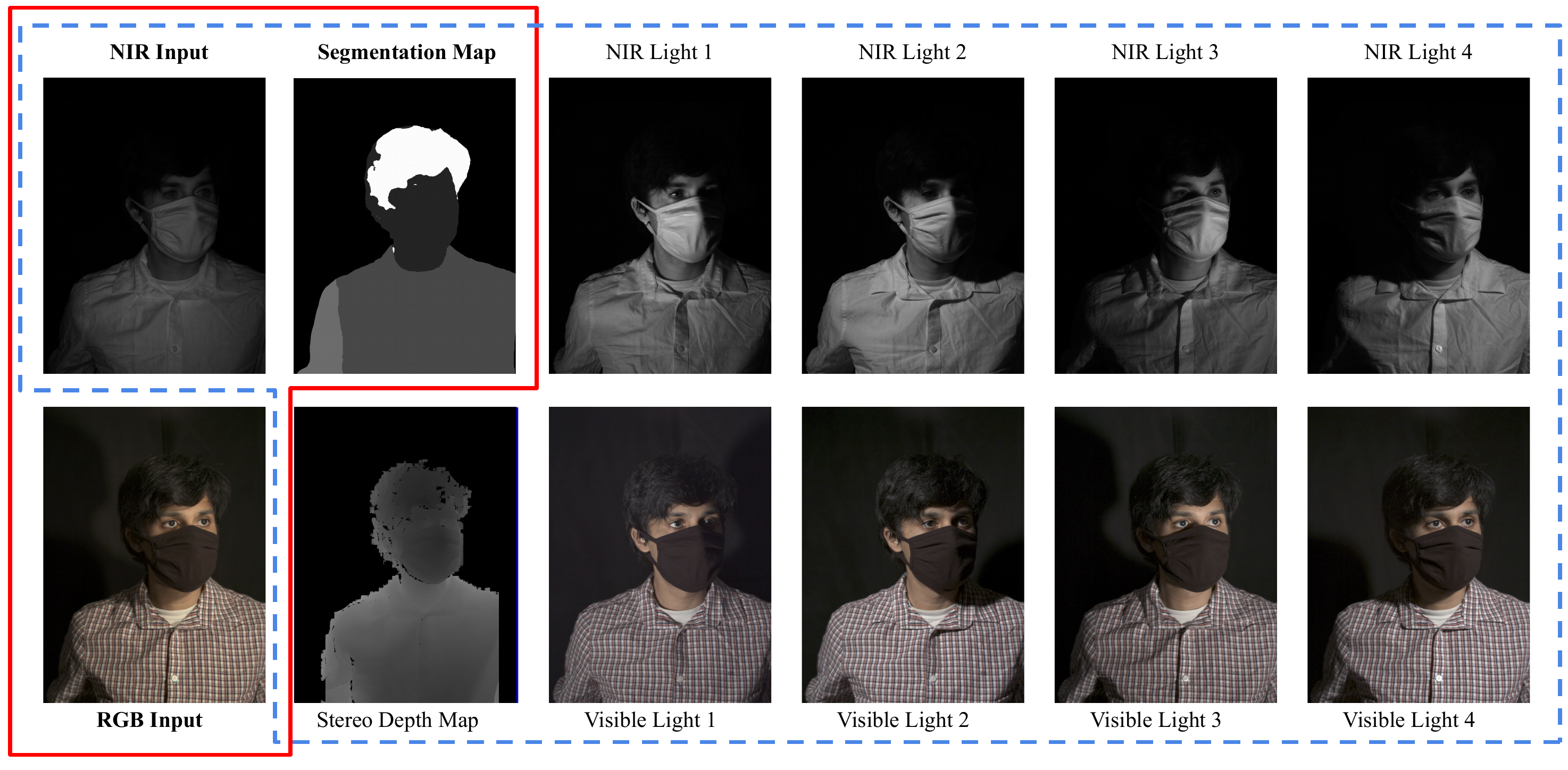}
    \caption{Our network learns to estimate shape and reflectance from a single front-lit NIR image, a single RGB image under arbitrary lighting, and a semantic segmentation map computed from the RGB image (inputs are enclosed by the red line). During training we also use a stereo depth map and replace the RGB image under arbitrary lighting with 4 RGB+NIR image pairs captured under calibrated point lights (the training inputs are inside the blue dashed line).}
    \label{fig:test_and_train_inputs}
\end{figure*}

\noindent \textbf{Monocular depth estimation.} Our work is also related to the growing number of techniques that estimate depth from a single image~\cite{Bhoi2019Monocular}. Deep neural networks have proven effective at performing this task when trained using ground truth depth data~\cite{Eigen2014Depth,Laina2016Deeper,Fu2018Deep, xia2020generating}. More recent work has focused on self-supervised training methods to overcome the difficulty of obtaining ground truth data~\cite{Zhou2017Unsupervised,Yang2017Unsupervised,Zhan2018Unsupervised,Godard2019Digging}. Sun et al.~\cite{Sun2012DepthEstimation} explored the use of a monocular infrared image for depth estimation and Li et al.~\cite{Li2020IVFuseNet} combined color and infrared images for depth estimation in the context of autonomous driving. Despite the progress in this field, state-of-the-art monocular depth estimation methods do not yet achieve the same level of accuracy as shape estimation methods that use some form of active lighting and constrain the input scene. \\

\noindent \textbf{Fusing depth and normals.} Depth estimated from methods like stereo triangulation and normals estimated from shading cues are complementary measurements for shape recovery. Nehab et al.~\cite{Nehab2005PositionNormals} describe a technique that seeks to combine the more accurate low-frequency information provided by direct depth measurement techniques with the higher-frequency geometric details provided by photometric measurements. We use their technique to evaluate how our approach could be used to improve a stereo pipeline (Section~\ref{subsection:results_refinement}). More recent work poses this as an optimization problem that seeks a surface that best agrees with these different signals~\cite{Barron2013SirfsKinect,Choe2014ShadingKinect,Yang2017Unsupervised,Xu2020StereoscopicFlash, Liang2020BetterTogether}. Our training method is similar to these approaches in that we also combine a stereo and photometric loss term. \\

\noindent \textbf{Face relighting.} 
Most single image face relighting methods include some representation of shape and reflectance as intermediate components. Our network architecture (Section~\ref{section:network}) is similar to the one proposed by Nestmeyer et al.~\cite{Nestmeyer2020FaceRelighting} for simulating lighting changes in a single image assumed to have been captured under a single directional light. Zhou et al.~\cite{Zhou2019DeepPortraitRelighting} present a dataset of relit portrait images generated using single-image normal and illumination estimates and a Lambertian reflectance model and then train a network with an adversarial loss to close the appearance gap between these synthetic images and real face images. Although surface geometry is fundamental to relighting, it is also possible to train an end-to-end network that does not explicitly reason about shape~\cite{Tiancheng2019PortraitRelighting}. We similarly use multiple images of a scene captured under varying controlled lighting to train our network in order to enable a much simpler set of inputs for inference. \\

\noindent \textbf{Combining infrared and color imagery.} A NIR (and/or ultraviolet) dark flash image can be used to denoise a color image captured in low visible light conditions~\cite{Krishnan2009DarkFlash}, or serve as a guide for correcting motion blur~\cite{Yamashita2017RGBNIRBracketing}. Techniques have also been developed that employ controlled NIR lighting to simulate better visible lighting in real-time video communication systems~\cite{Wang2008VideoRelighting, Gunawardane2010InvisibleLight}. We see these as compelling potential applications of this work.

\begin{figure*}
    \centering
    \includegraphics[width=0.95\textwidth]{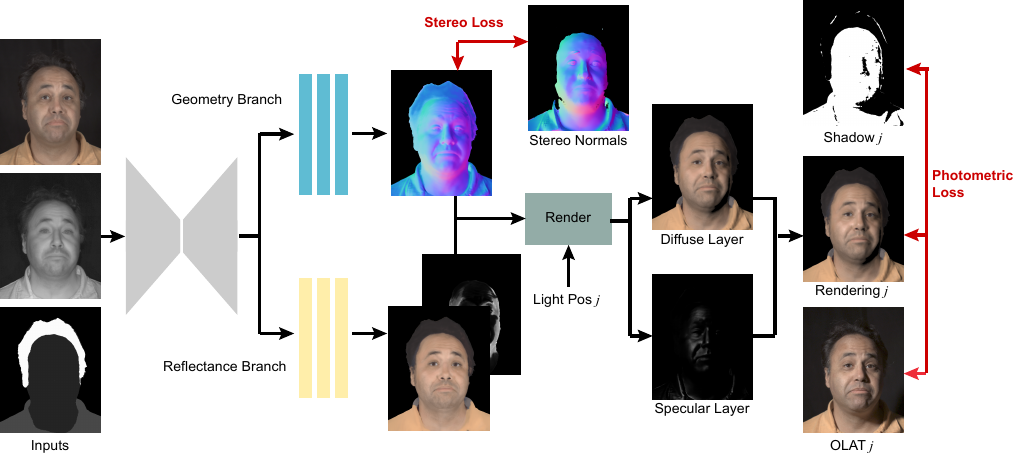}
    \caption{Illustration of our network and training strategy. We estimate network weights that minimize a photometric loss, computed between images rendered from our network outputs and ground truth images captured under known lighting, and a stereo loss, driven by differences between the output normals and those estimated using an independent stereo technique.}
    \label{fig:system}
\end{figure*}

\section{Network Design and Training}
\label{section:network}

Our goal is to estimate a normal map and a reflectance map from a single RGB image and a front-lit ``dark flash'' NIR image. We train a deep neural network to perform this task. As an auxiliary input, we use a 6-class semantic segmentation map computed from the RGB image (background, head, hair, body, upper arm and lower arm)~\cite{Chen2017DeepLab3}.
We found this segmentation map was a useful cue for helping the network reason about shape and reflectance. An example set of inputs are shown in Figure~\ref{fig:test_and_train_inputs} (red line).

Our training procedure is driven in part by a physically-based image formation model that connects the outputs of our network to images of a scene taken under known point lighting. This image formation model combines a standard Lambertian diffuse term with the Blinn-Phong BRDF~\cite{Blinn77ReflectionModel}, which has been used to model the specular reflectance of human skin~\cite{Weyrich2006SkinReflectance}.
Specifically, we introduce a reflectance function $f$ that gives the ratio of reflected light to incident light for a particular unit-length light vector $\l$, view vector $\v$, surface normal $\n$, four-channel (RGB+NIR) albedo $\a$, scalar specular intensity $\spec$, and specular exponent $m$:
\begin{equation}
    f(\l, \v, \n) = \a + \spec \frac{m + 2}{2 \pi} ( \n \cdot \h)^m,
    \label{eq:brdf}
\end{equation}
where $\h = (\n + \l)/\norm{\n + \l}$. The observed intensity at a pixel due to a point light is given by
\begin{equation}
    I(\cdot) = f(\l, \v, \n) (\n \cdot \l) L,
    \label{eq:intensity-at-pixel}
\end{equation}
the product of the reflectance, cosine term, and light intensity $L$. We do not observe the reflected intensity from enough unique light directions at each pixel to estimate all of the parameters in Equation~\ref{eq:brdf}. We therefore fix the specular exponent to $m=30$ based on prior measurements of human skin~\cite{Weyrich2006SkinReflectance} and our own observations, and estimate only $\n$, $\a$, and $\spec$. The geometric quantities $\l$ and $\v$, and light intensity $L$ are determined by the calibration procedures described in Section~\ref{section:dataset}.

Illustrated in Figure~\ref{fig:system}, we use a standard UNet with skip connections~\cite{Ronneberger2015UNet}. The encoder and decoder each consist of 5 blocks with 3 convolutional layers per block. The bottleneck has 256 channels. The output of this UNet is forwarded to two separate networks: a geometry branch that predicts a normal map $\nest$, and a reflectance branch that predicts an albedo map $\aest$ and log-scale specular intensity map, $log(\specest)$. Both branches have 3 convolutional layers with 32 channels and one final output layer.

We do not rely on ground truth normals or reflectance data to supervise training. Instead we combine a stereo loss and a photometric loss derived from data that is far easier to obtain: four one-light-at-a-time (OLAT) images in both RGB and NIR of the same subject, in the same exact pose, illuminated by a set of calibrated lights activated individually in rapid succession, and a stereo depth map (blue dashed line in Figure~\ref{fig:test_and_train_inputs}). These images are only used at training time.

A stereo loss encourages our estimated normals $\nest$ to agree with the gradients of a smoothed version of the stereo depth map $\n_s$. Specifically, it combines a L1 vector loss and angular loss:
\begin{equation}
    \mathcal{L}_{s}(\nest) = \|\nest - \n_s\|_{1} - (\nest \cdot \n_s).
\end{equation}

A photometric loss is computed between each of the OLAT images and an image rendered according to Equation~\ref{eq:intensity-at-pixel} and our network outputs for the corresponding lighting condition:
\begin{equation}
    \mathcal{L}_{p}^{j}(\nest, \aest, \specest) = \norm{ S_{j} \odot \left(\Iest(\l_{j}, \v, \nest, \aest, \specest) - \Iobs_{j} \right) }_{1},
    \label{eq:photoloss}
\end{equation}
where $\Iobs_{j}$ is the per-pixel color observed in the $j^\mathrm{th}$ OLAT image, and $S_{j}$ is a binary shadow map, computed using the stereo depth and calibrated light position (Section~\ref{section:dataset}). We also apply a prior to the albedo map that encourages piecewise constant variation~\cite{Barron2015Sirfs}:
\begin{equation}
    \mathcal{L}_{c}(\aest) = \sum_i\sum_{j \in \mathcal{N}(i)}\lVert \aest_i - \aest_j \rVert_1,
\end{equation}
where $\mathcal{N}(i)$ is the $5\times5$ neighborhood centered at pixel $i$. We apply this prior only to clothing pixels, those labeled as either body or arms in the segmentation mask. We found that other regions in the scene did not benefit from this regularization.

Our total loss function is a weighted sum of these terms:
\begin{multline}
    \mathcal{L}(\nest, \aest, \specest) = \\ \mathcal{L}_{s}(\nest) + \lambda_{p} \sum_j{\mathcal{L}_{p}^{j}(\nest, \aest, \specest)} + \lambda_{c} \mathcal{L}_{c}(\aest).
\end{multline}
We set the weight $\lambda_p$ to $10$ and $\lambda_c$ to $50$.

\paragraph{Data Augmentation and Training.}  To improve the robustness of our network, we apply a series of data augmentations to our captured OLATs to simulate a variety of different visible light conditions. Specifically, our training uses a combination of: evenly-lit RGB inputs obtained by adding together all of the OLAT images; inputs with strong shadows by selecting exactly one of the OLAT images; a mixture of two lights with different temperatures by applying randomly chosen color vectors to two randomly chosen OLAT images; low-light environments by adding Gaussian noise to a single OLAT; and saturated exposures by scaling and clipping a single OLAT. We sample evenly from these 5 lighting conditions during training. Further details on how these lighting conditions are simulated are provided in the supplementary material.

We train the network using the Adam optimizer~\cite{Kingma14AdamOptimizer} for $30$K iterations, with a learning rate of $10^{-3}$ and a batch size of $8$. Training takes around $12$ hours on a machine with 4 Tesla V100 GPUs.

\section{Hardware Setup and Data Collection}
\label{section:dataset}

\begin{figure}
    \centering
    \includegraphics[width=0.98\columnwidth]{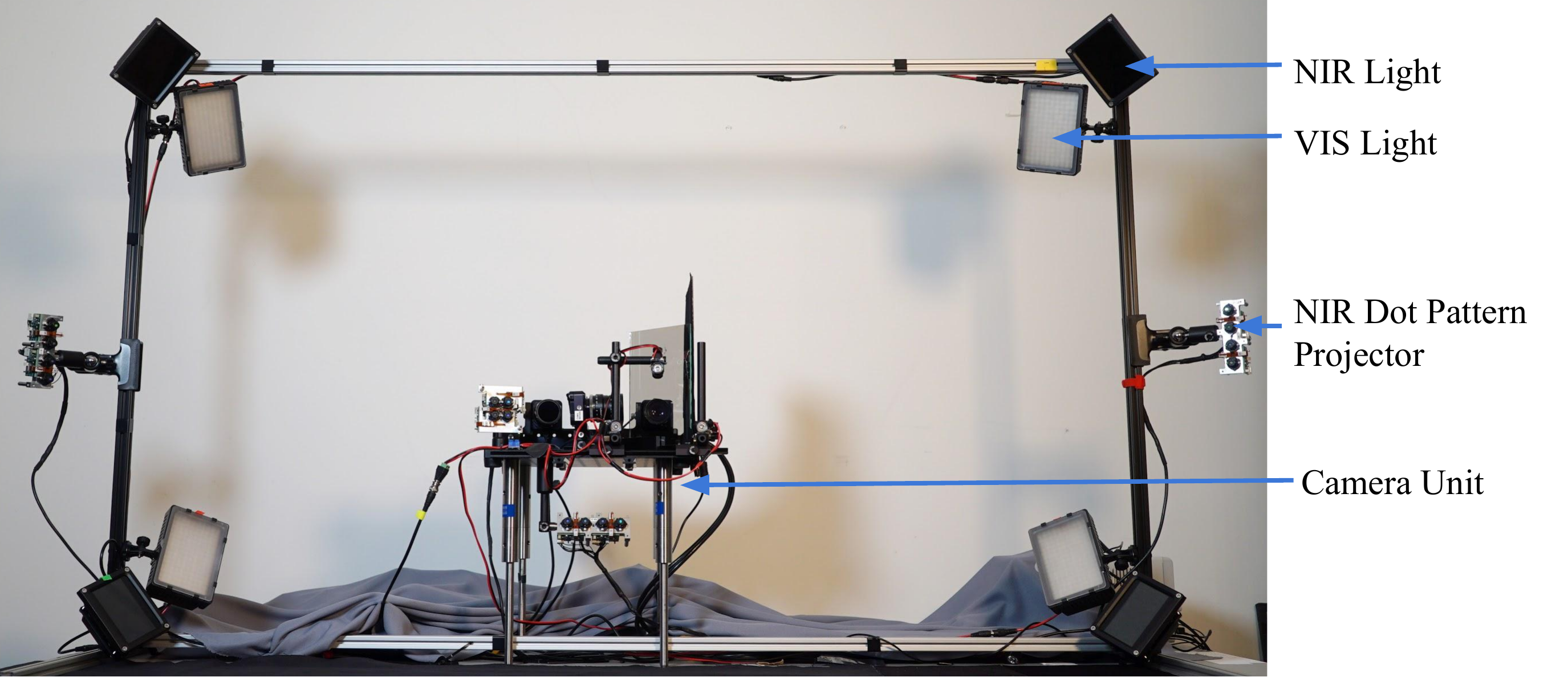}
    \includegraphics[width=0.98\columnwidth]{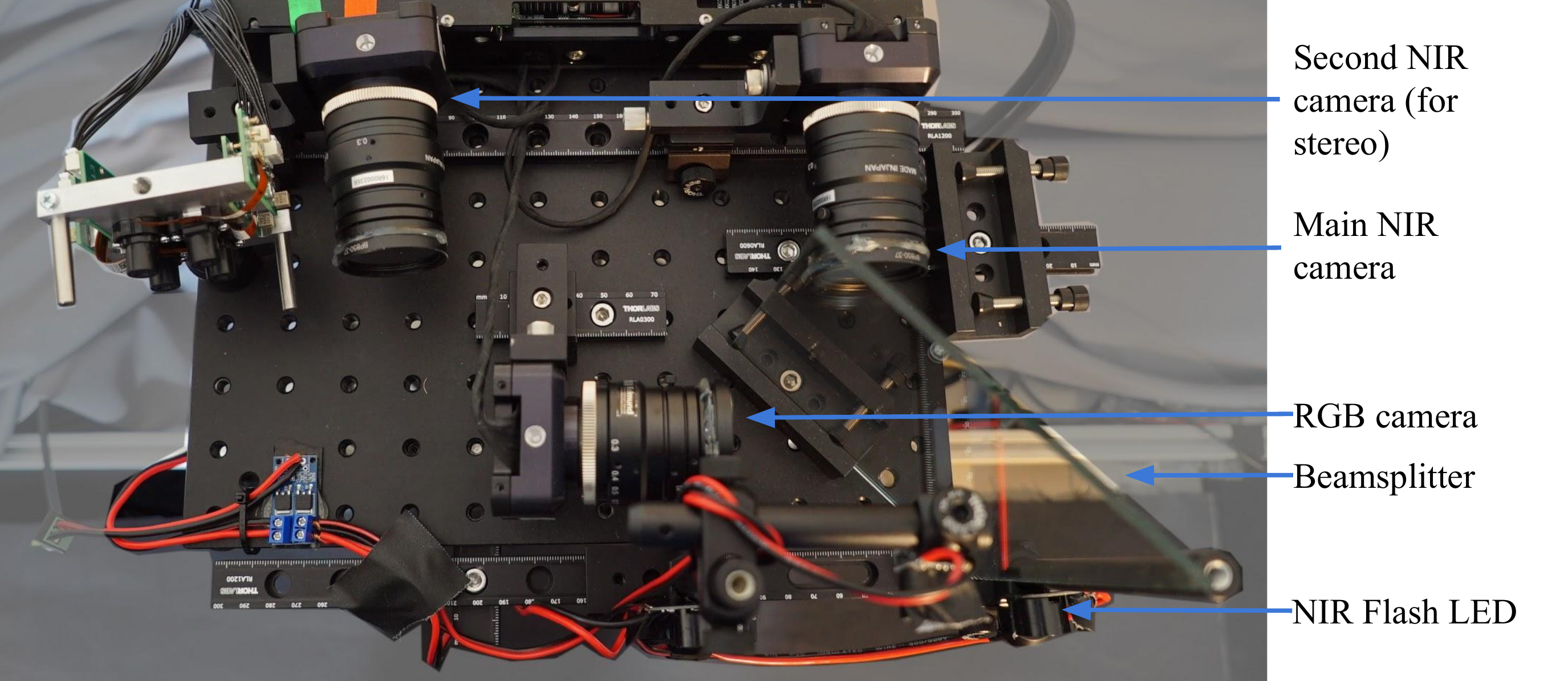}
    \caption{Our hardware setup consists of controllable NIR and visible spectrum light sources, an RGB camera, a stereo pair of NIR cameras, and two NIR dot projectors. One of the NIR cameras and the RGB camera are aligned with a beamsplitter and all of these components are triggered electronically to record the types of images shown in Figure~\ref{fig:test_and_train_inputs}.}
\label{fig:experimental_setup}
\end{figure}

Shown in Figure~\ref{fig:experimental_setup}, our setup combines a $7.0$MP RGB camera that operates at $66.67$ fps with a stereo pair of $2.8$MP NIR cameras that operate at $150$ fps. The RGB camera and one of the NIR cameras are co-located using a plate beamsplitter and a light trap. The RGB and NIR cameras have a linear photometric response and we downsample all of the images by a factor of 2 in each dimension and take a central crop that covers the face at a resolution of $960\times768$.

Visible spectrum lighting is provided by 4 wide-angle LED spotlights placed at the corners of a roughly $1.5m \times 0.8m$ (width x height) rectangle surrounding the cameras located approximately $1.1$m from the subject. NIR lighting is provided by 5 NIR spotlights, one adjacent to each of the visible lights, and a flash LED light located near the reference NIR camera to produce the ``dark flash'' input. These NIR light sources are temporally interleaved with projectors that emit NIR dot speckle patterns to assist stereo matching~\cite{Nover2018Espresso}. A microcontroller orchestrates triggering the lights and cameras to ensure that at any time only one visible light source and one NIR light source is active.
All light sources are calibrated for position and intensity and treated geometrically as point light sources. The light intensity term $L$ in Equation~\ref{eq:intensity-at-pixel} accounts for these calibrated colors. Note that the NIR and visible light sources are not colocated and so slightly different values of $\l$ are used in Equation~\ref{eq:intensity-at-pixel} between those two conditions.

\begin{table*}[!ht]\small
  \begin{center}
  \setlength\tabcolsep{4pt}
  \begin{tabular}{|c|l|ccccc|}
  \hline
  \multicolumn{2}{|c|}{Ablation} & Well lit  & Shadows  & Mixed colors  & Overexposure & Low light\\ \hline
  \multirow{3}{*}{Loss} & No Stereo Loss & 12.80 & 12.78 & 12.78 & 12.82 & 12.81 \\
  & No NIR Photometric Loss & 12.64 & 12.66 & 12.64 & 12.69 & 12.75 \\
  & No Photometric Loss & 12.77 & 12.77 & 12.81 & 12.79 & 12.77 \\ \hline
  Reflectance Model & No Specular Component & 12.44 & 12.43 & 12.44 & 12.51 & 12.47 \\ \hline
  \multirow{2}{*}{Inputs} & No RGB Input & 12.54 & 12.54 & 12.54 & 12.54 & 12.54 \\
  & No NIR Input & 13.13 & 15.19 & 16.43 & 19.82 & 19.39 \\ \hline
  \multicolumn{2}{|c|}{Ours (Full Method)} & \textbf{12.08} & \textbf{12.06} & \textbf{12.06} & \textbf{12.14} & \textbf{12.10} \\ \hline
  
  \end{tabular}
  \end{center}
  \vspace{-10pt}
  \caption{Mean absolute angular error in degrees of normal maps computed with modified versions of our full network. Results are reported for the five lighting conditions described in Section~\ref{section:Evaluation}.}
  \label{table:ablation}
\end{table*}


\begin{figure*}
\centering
    \setlength{\tabcolsep}{2pt}
    \newcommand{\imw}{0.19\textwidth}
    \centering
    \begin{tabular}{ccccc}
    \includegraphics[width=\imw]{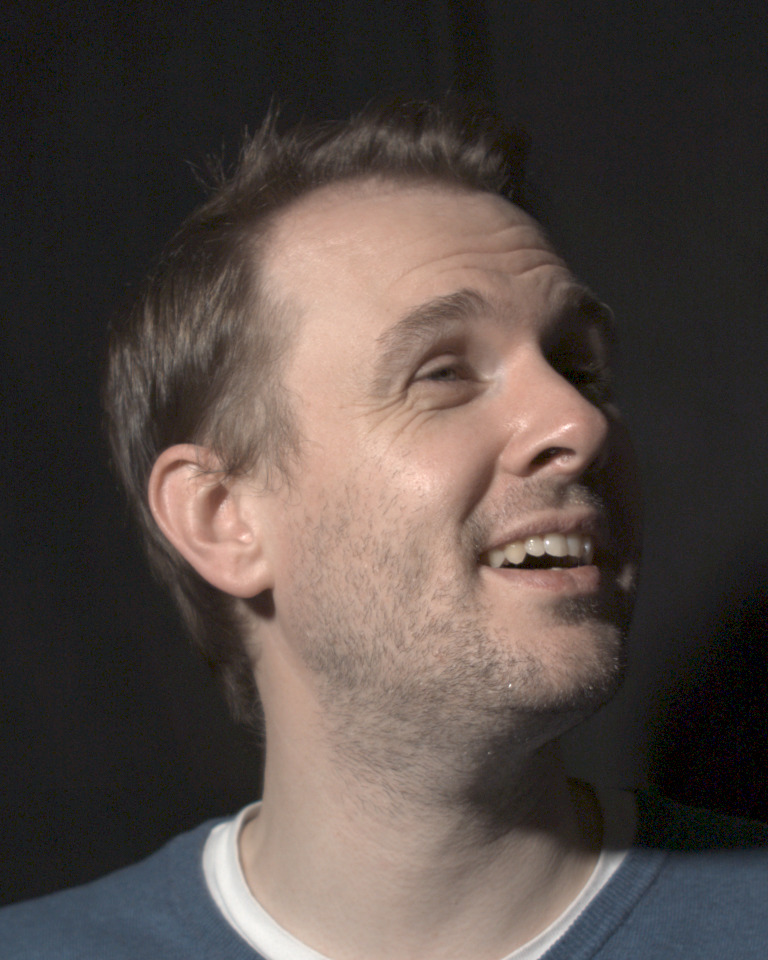} &
    \includegraphics[width=\imw]{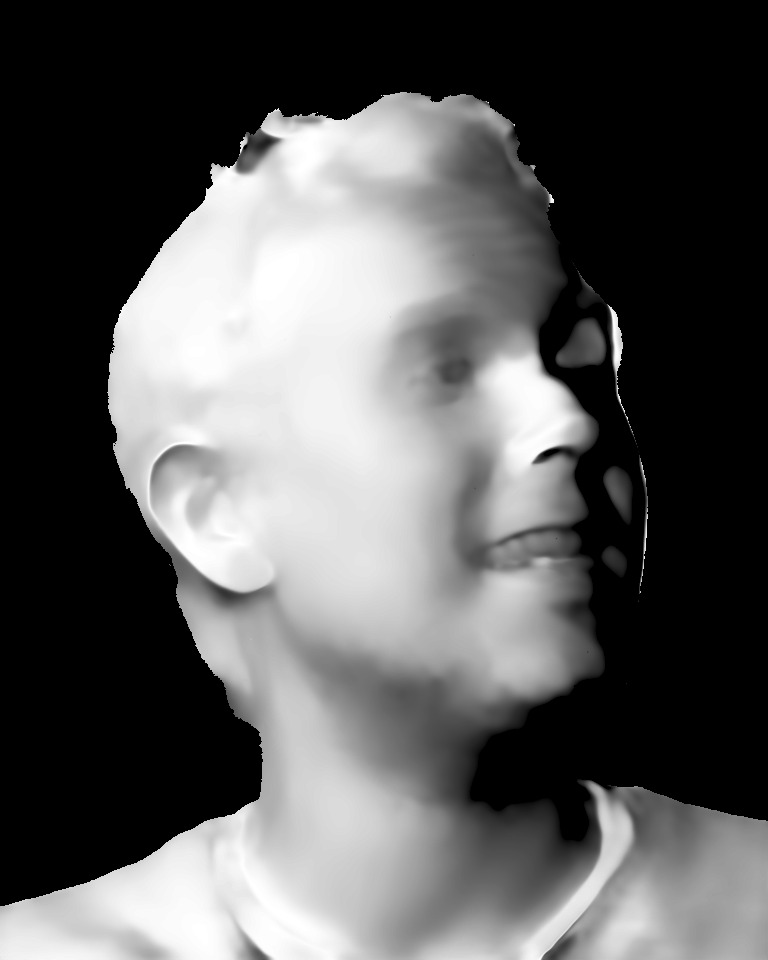} &
    \includegraphics[width=\imw]{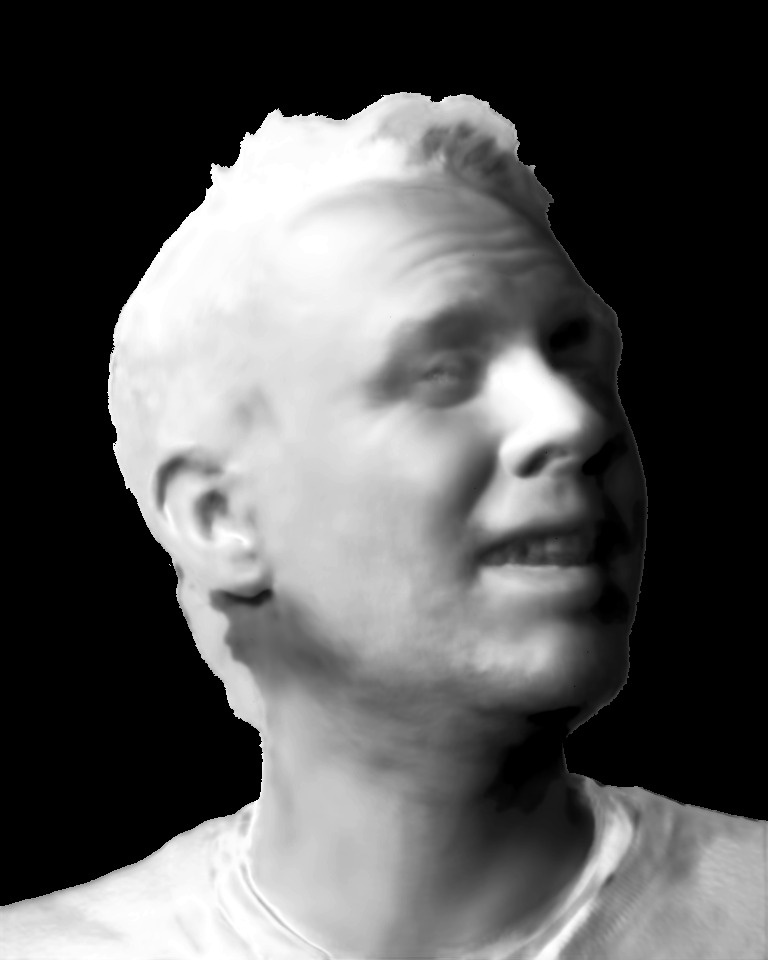} &
    \includegraphics[width=\imw]{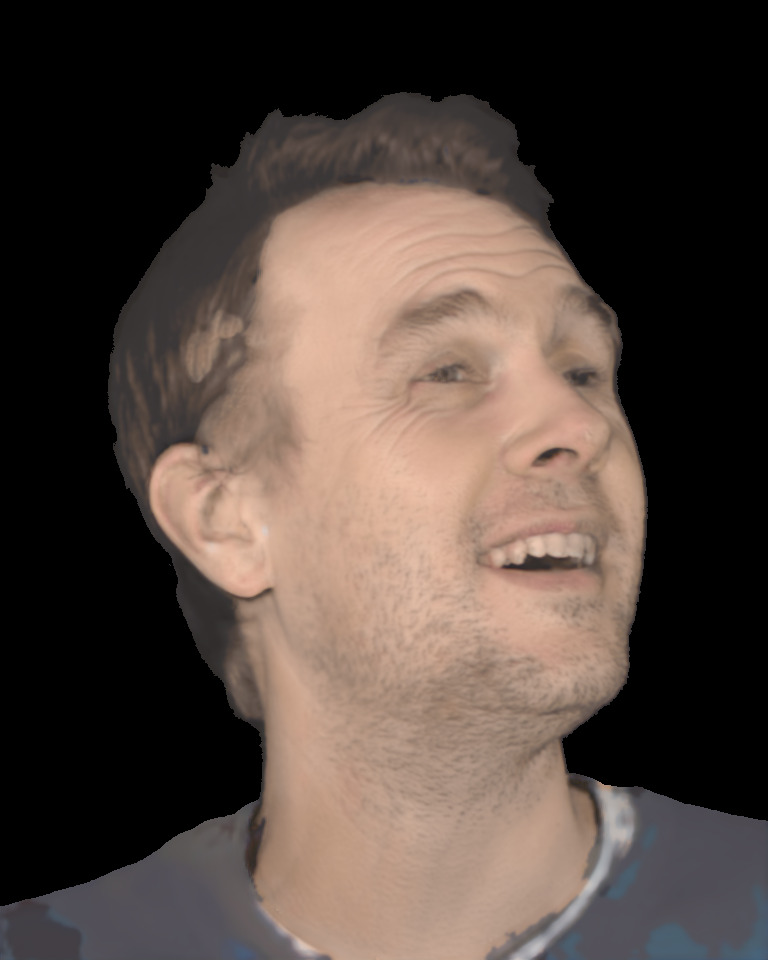} &
    \includegraphics[width=\imw]{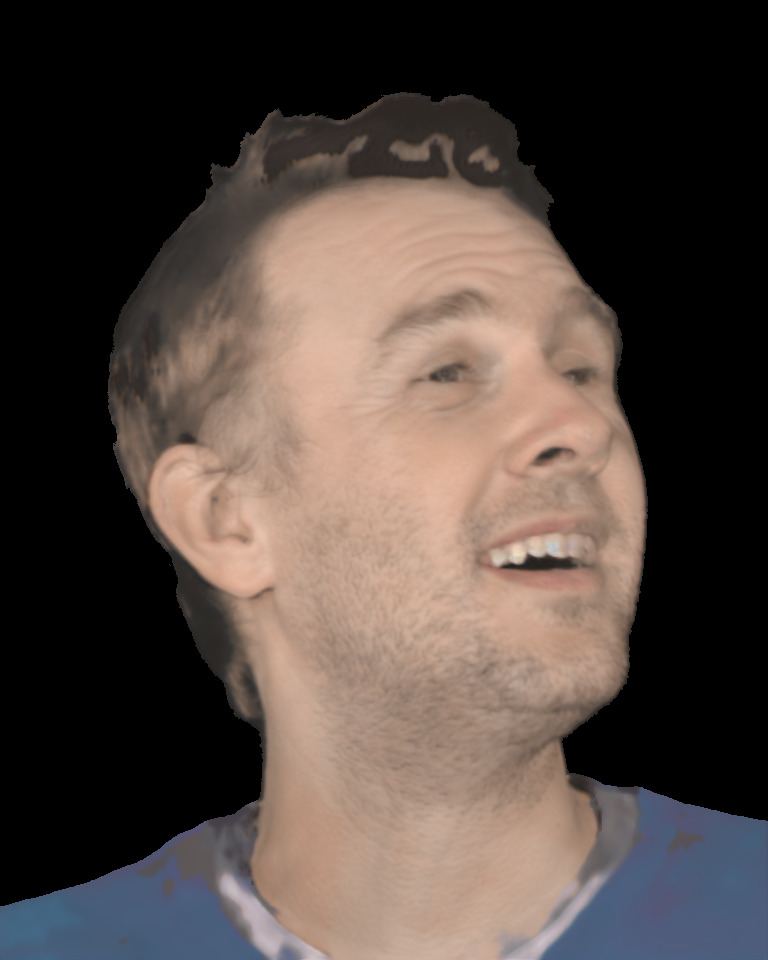} \\
    \small{RGB Input} & \small{Shading} & \small{Shading} & \small{Albedos} & \small{Albedos}\\
    &  \small{(w/o Photometric Loss)} & \small{(w/ Photometric Loss)} & \small{(w/o Blinn-Phong)} & \small{(w/ Blinn-Phong)} \vspace{-5pt}
    \end{tabular}
    \vspace{0pt}
    \caption{Impact of the photometric loss term in our training procedure and the Blinn-Phong BRDF in our image formation model, respectively. When trained without photometric loss, our network learns to output the stereo normals, which lack fine-scale details. This has a fairly small effect on the error measures in Table~\ref{table:ablation}, but is perceptually significant as seen in these ``n dot l'' shading renderings. Our full image formation model, which includes a Blinn-Phong specular term, produces more accurate albedos across the face than using a Lambertian model alone.}
    \label{figure:ablation}
\end{figure*}

The image acquisition rate is limited by the RGB camera's framerate and the total light output, but is fast enough for us to record video sequences of people who are gesturing and moving slowly. We compute optical flow~\cite{Jiangjian2006BilateralFlow} between consecutive frames captured under the same lighting condition to correct for the small amount of scene motion that occurs within a single round of exposures. Since the RGB and reference stereo NIR camera are co-located, we can generate pixel-aligned RGB, NIR, and depth images using scene-independent precomputed image warps.

Each recording in our dataset is $10$ seconds long and contains $166$ sets of frames. We recorded $9$ unique subjects, with between $5$ and $10$ sessions per subject, for a total of $61$ recordings.
We used recordings of 6 of the subjects for training and tested on recordings of the other 3.

\begin{figure*}
    \newcommand{\imh}{3.4cm}
    \setlength{\tabcolsep}{2pt}    
    \centering
    \begin{tabular}{cccccc}
    \includegraphics[height=\imh]{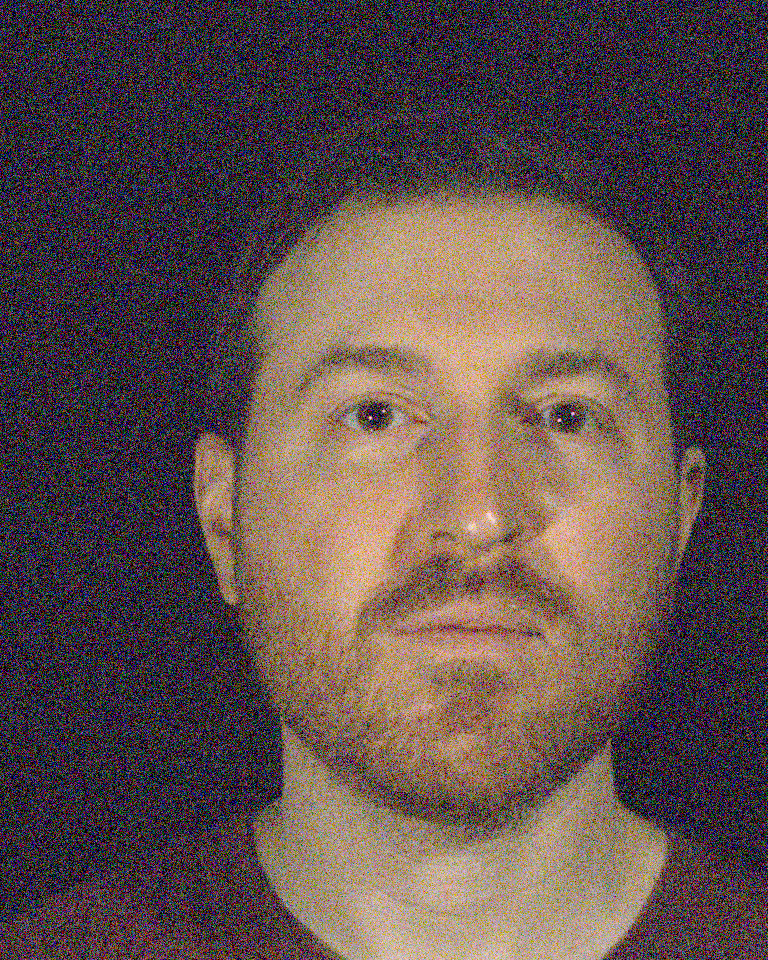} &
    \includegraphics[height=\imh]{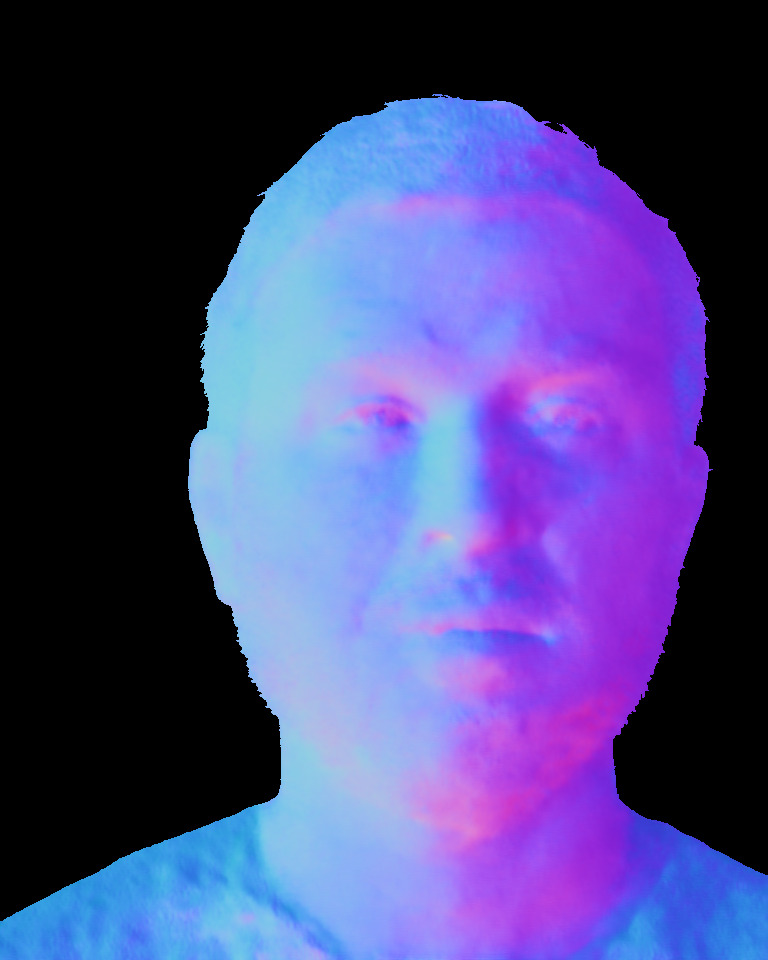} &
    \includegraphics[height=\imh]{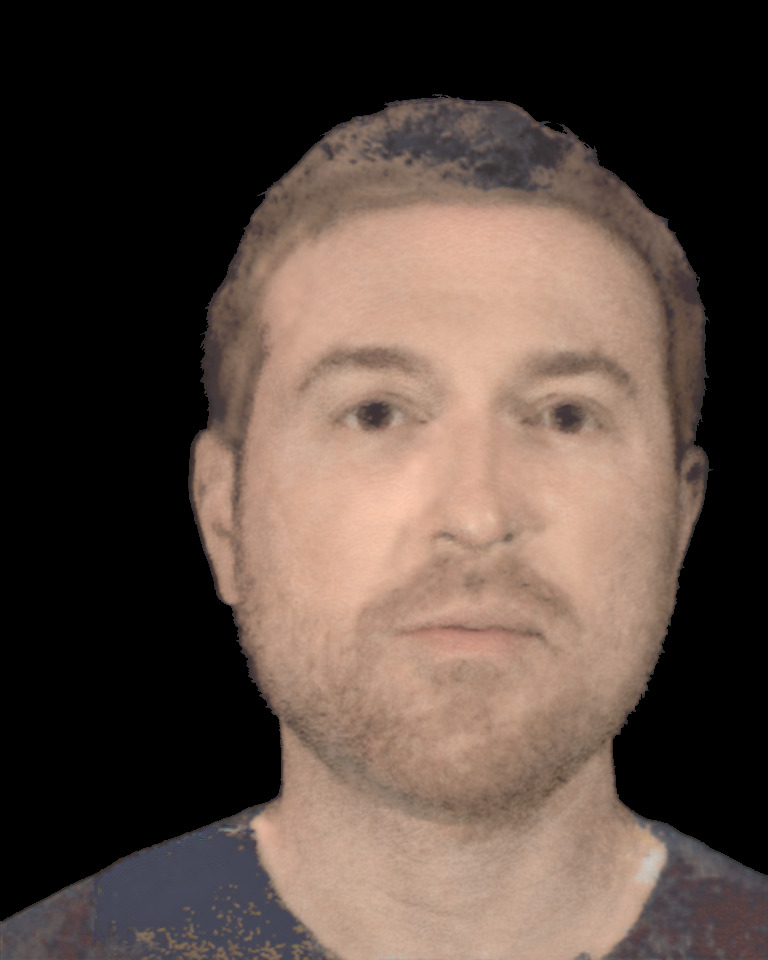} &
    \includegraphics[height=\imh]{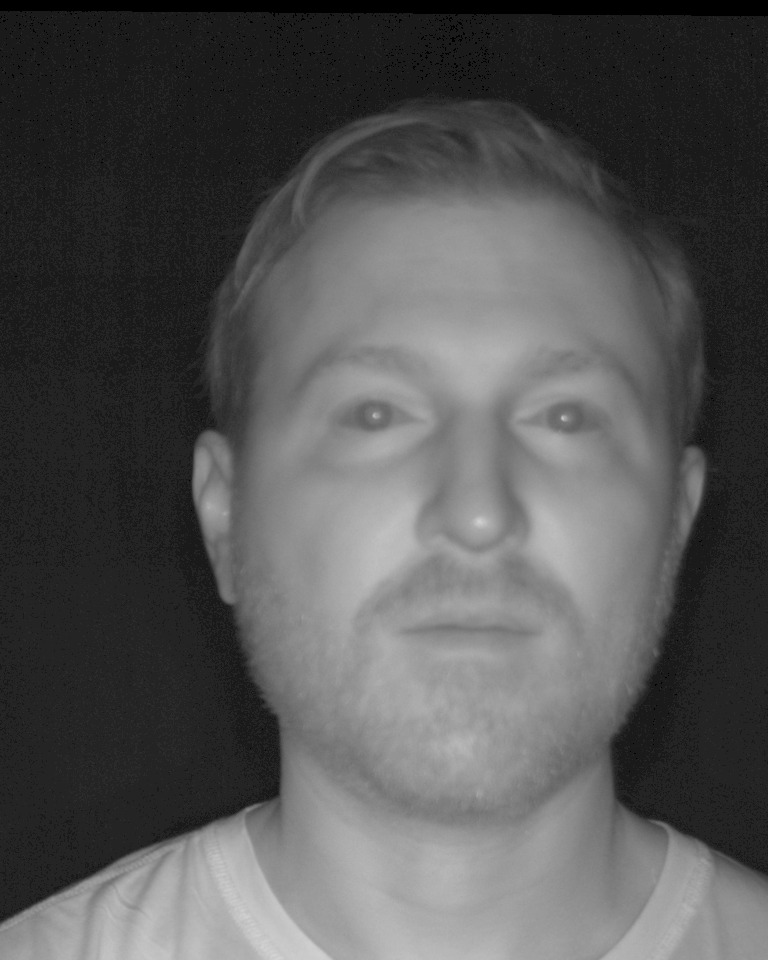} &
    \includegraphics[height=\imh]{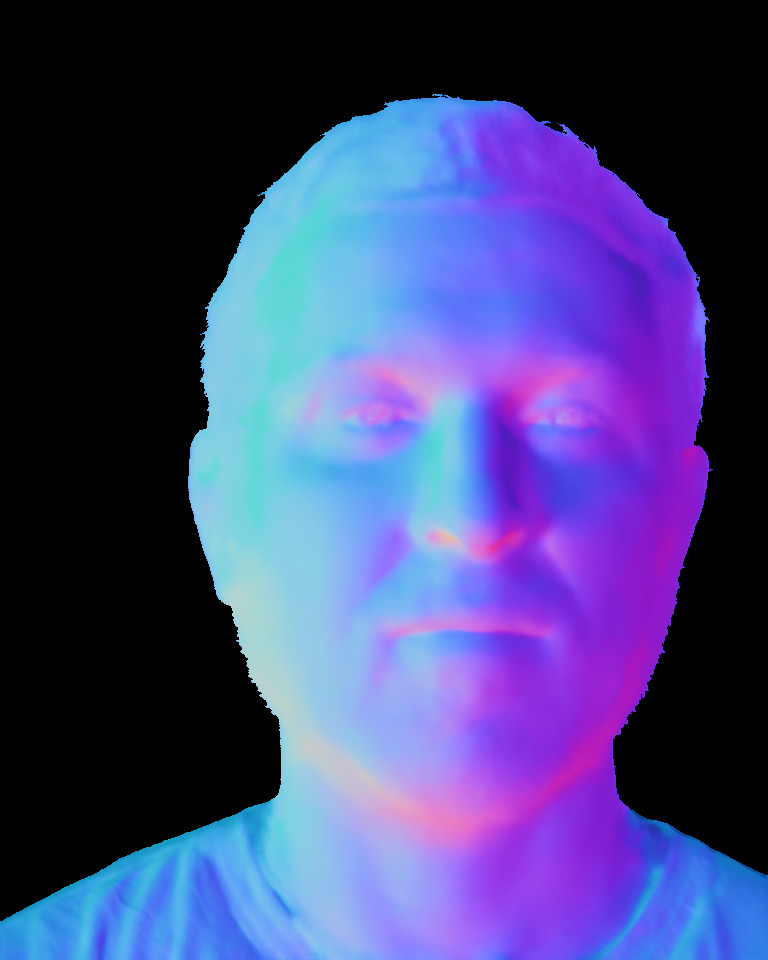} &
    \includegraphics[height=\imh]{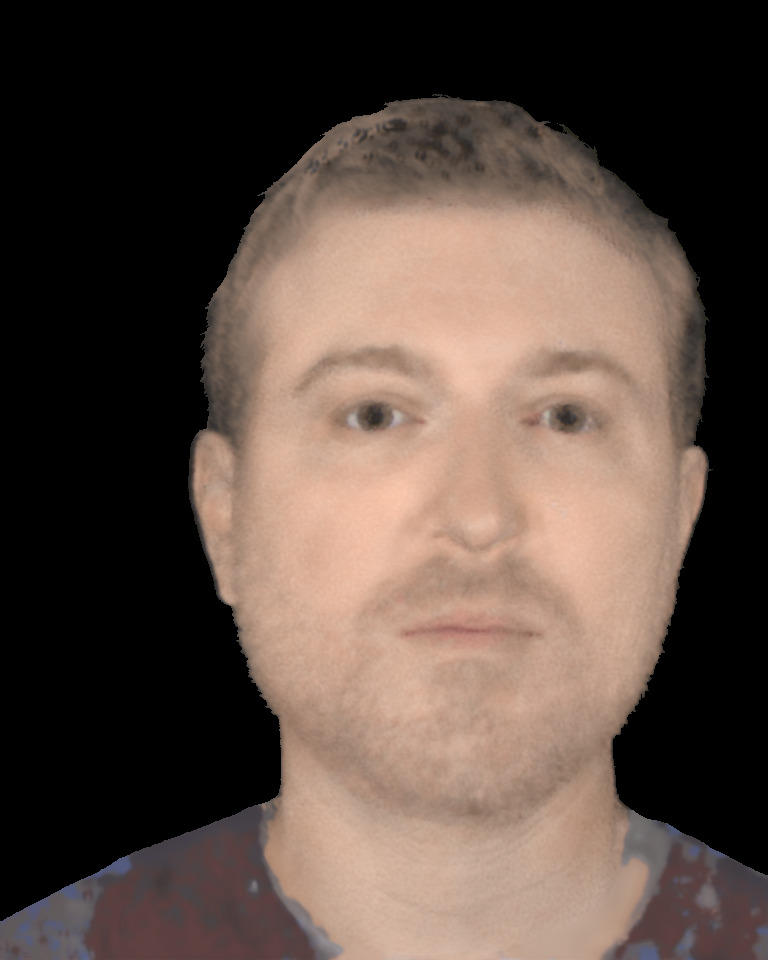} \\
    \includegraphics[height=\imh]{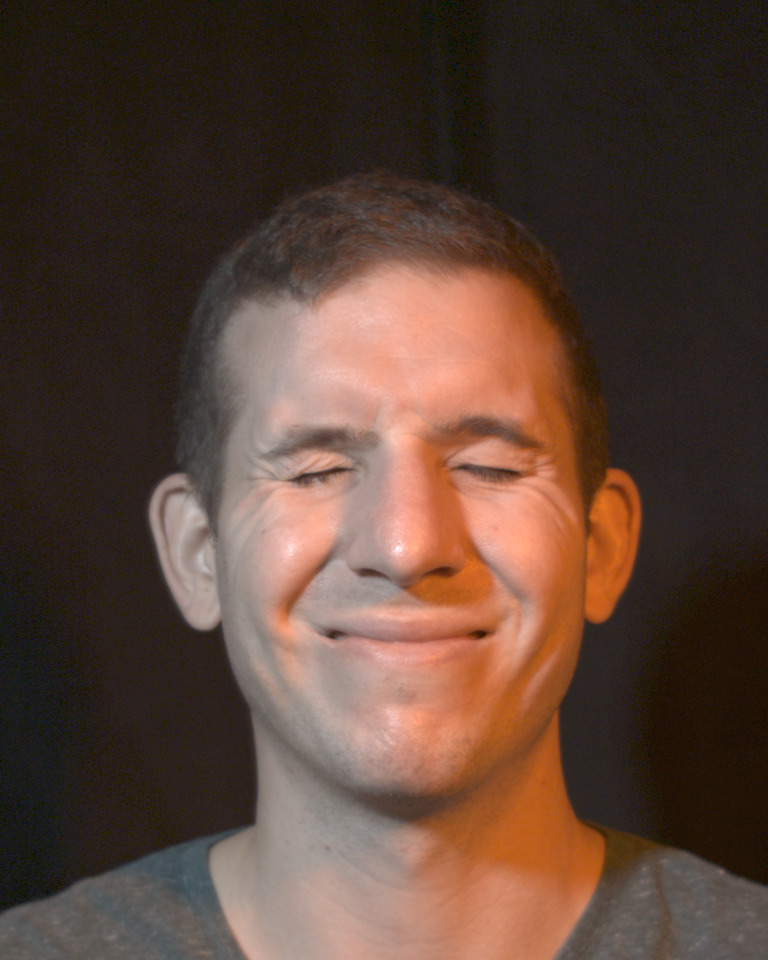} &
    \includegraphics[height=\imh]{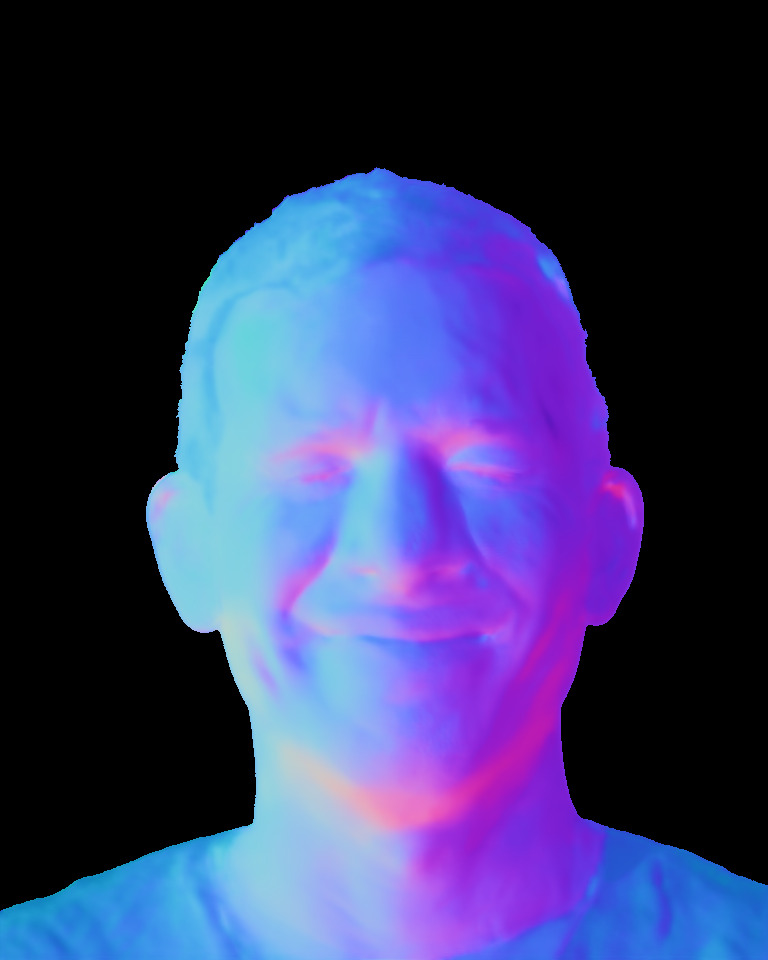} &
    \includegraphics[height=\imh]{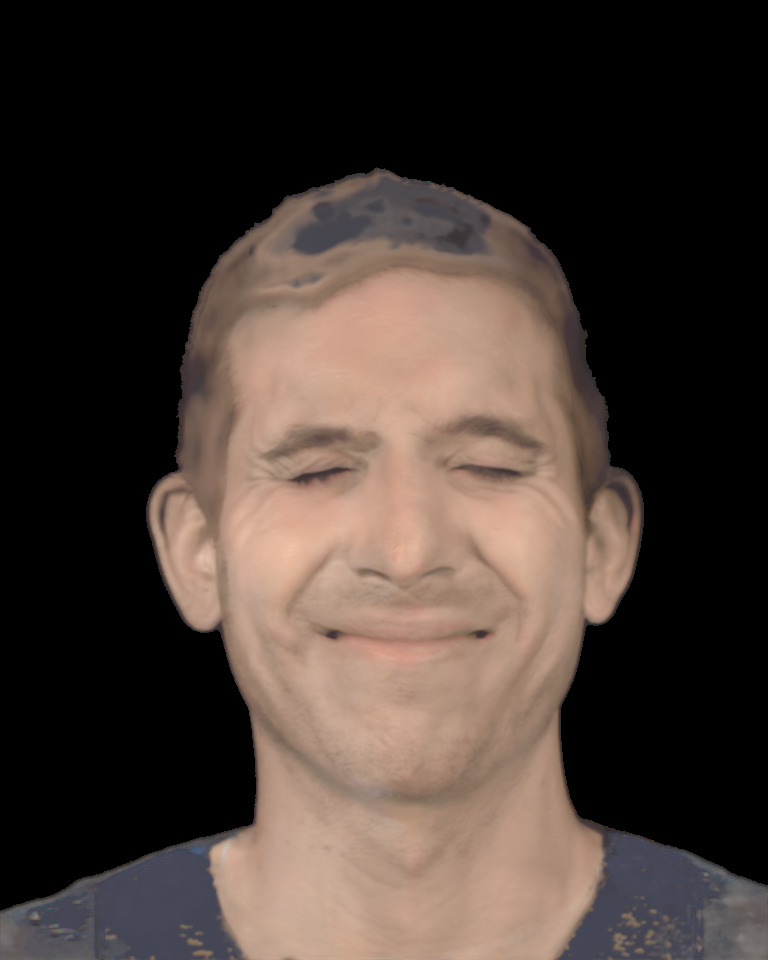} &
    \includegraphics[height=\imh]{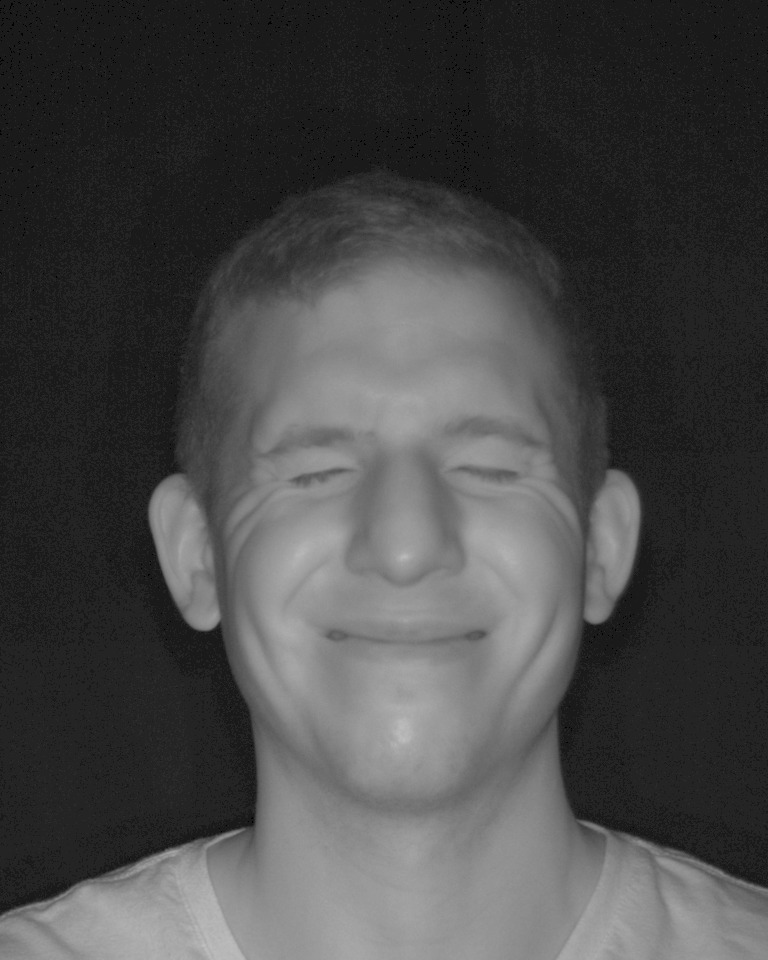} &
    \includegraphics[height=\imh]{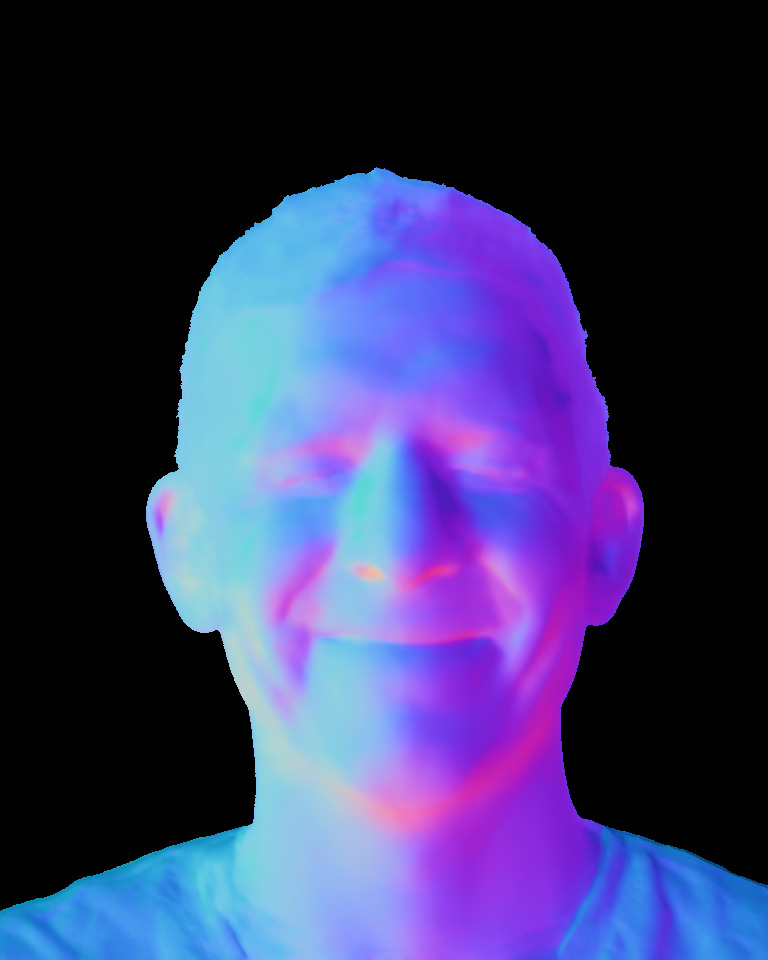} &
    \includegraphics[height=\imh]{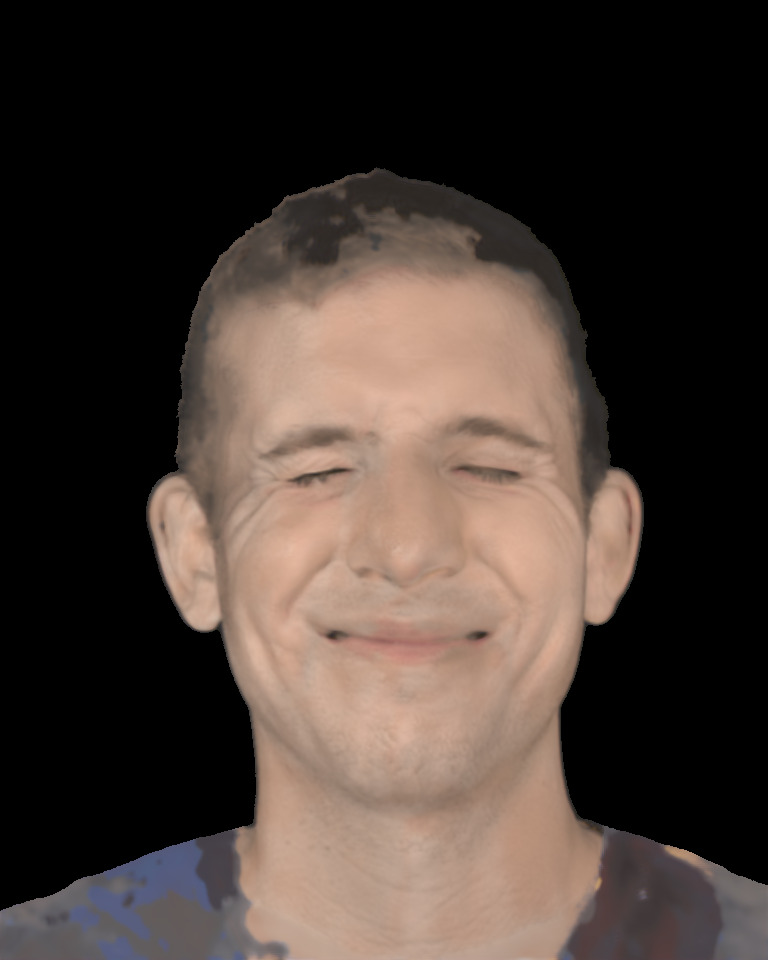} \\
    \includegraphics[height=\imh]{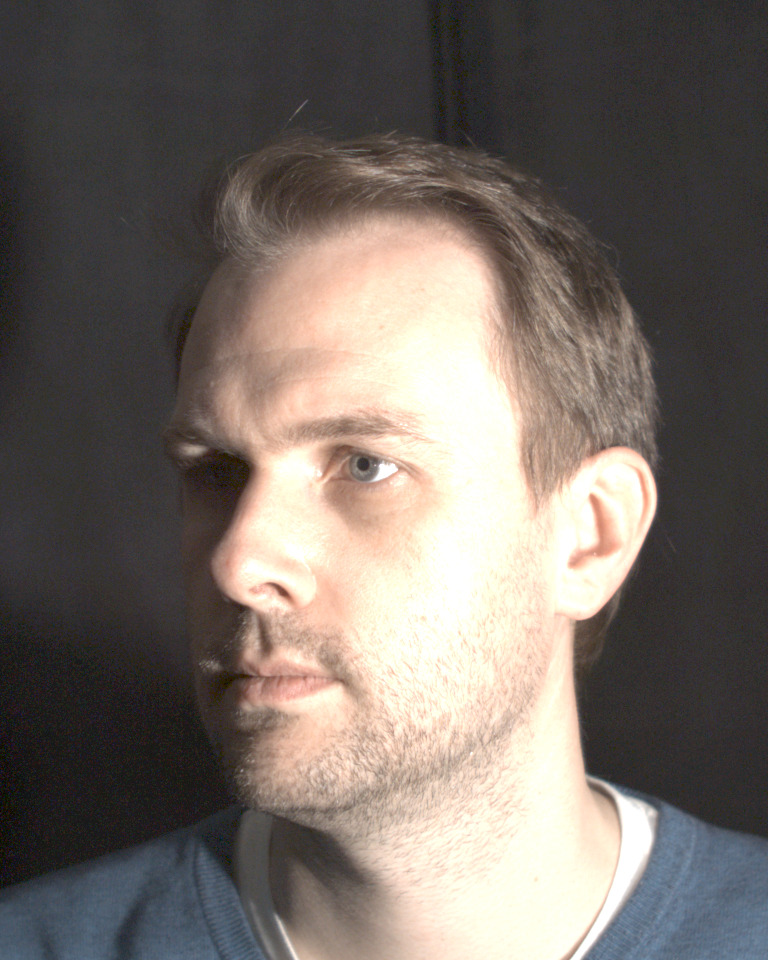} &
    \includegraphics[height=\imh]{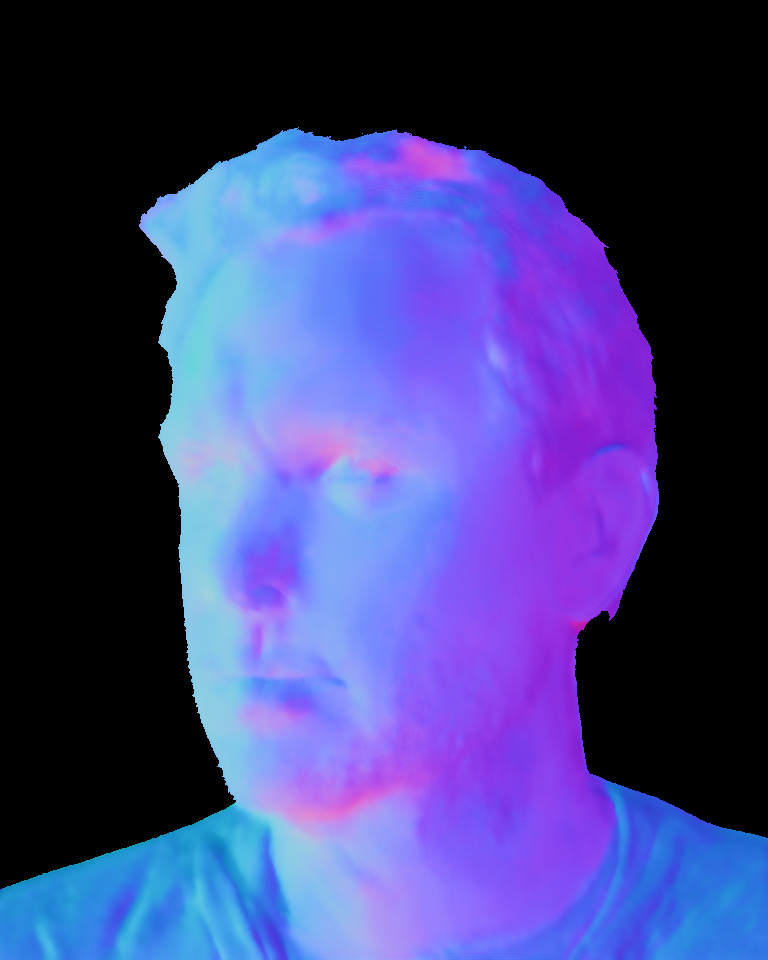} &
    \includegraphics[height=\imh]{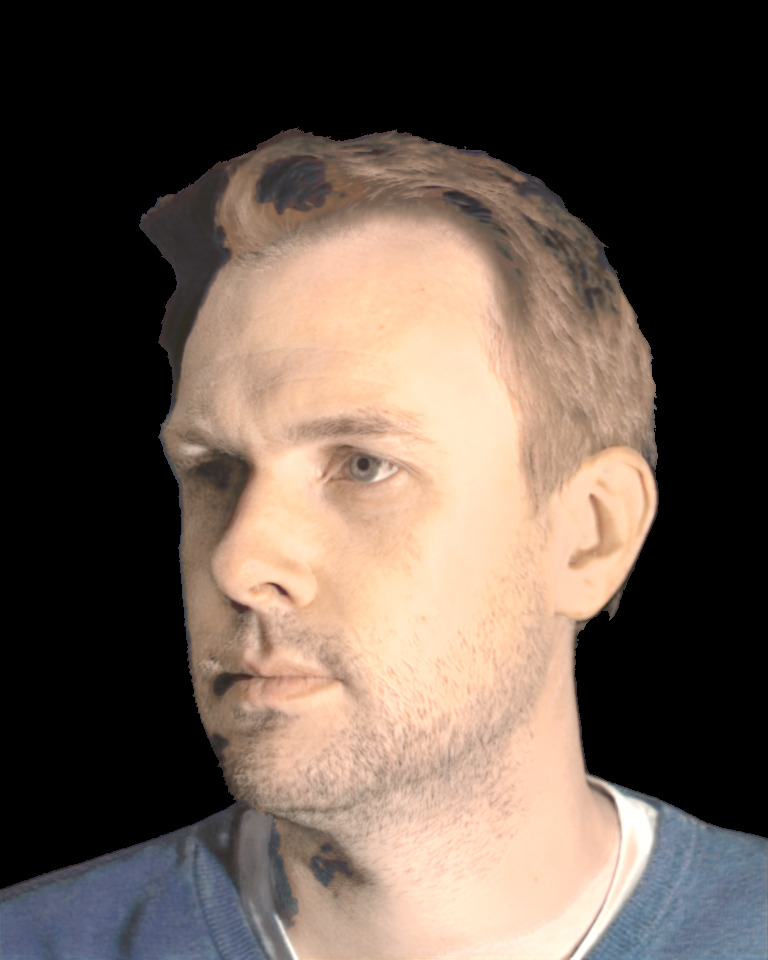} &
    \includegraphics[height=\imh]{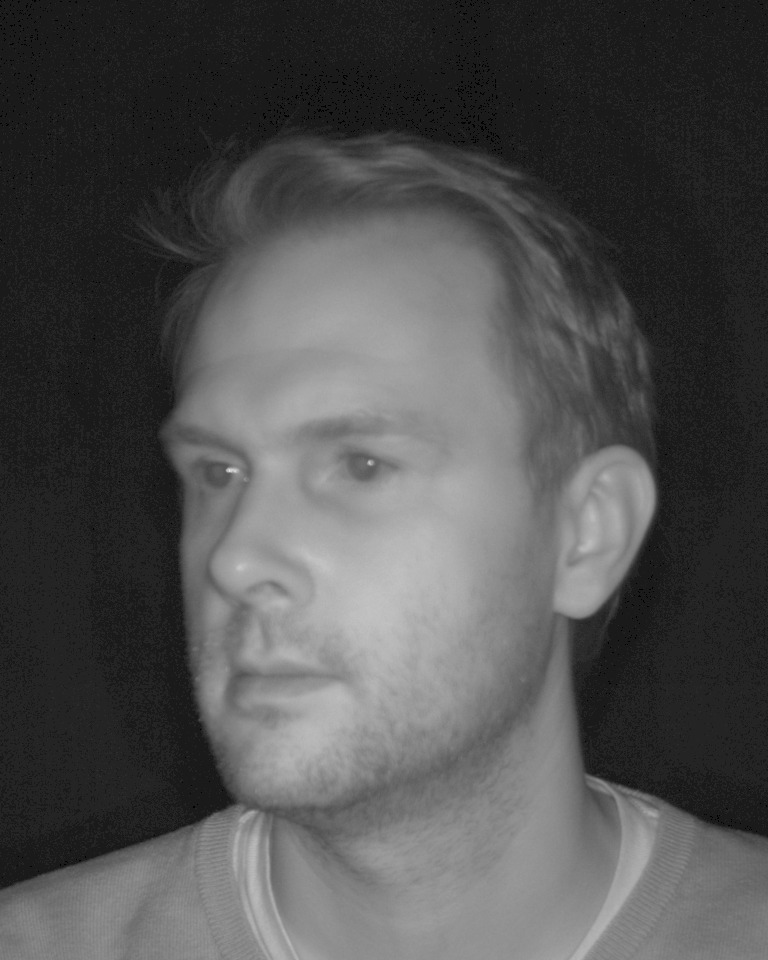} &
    \includegraphics[height=\imh]{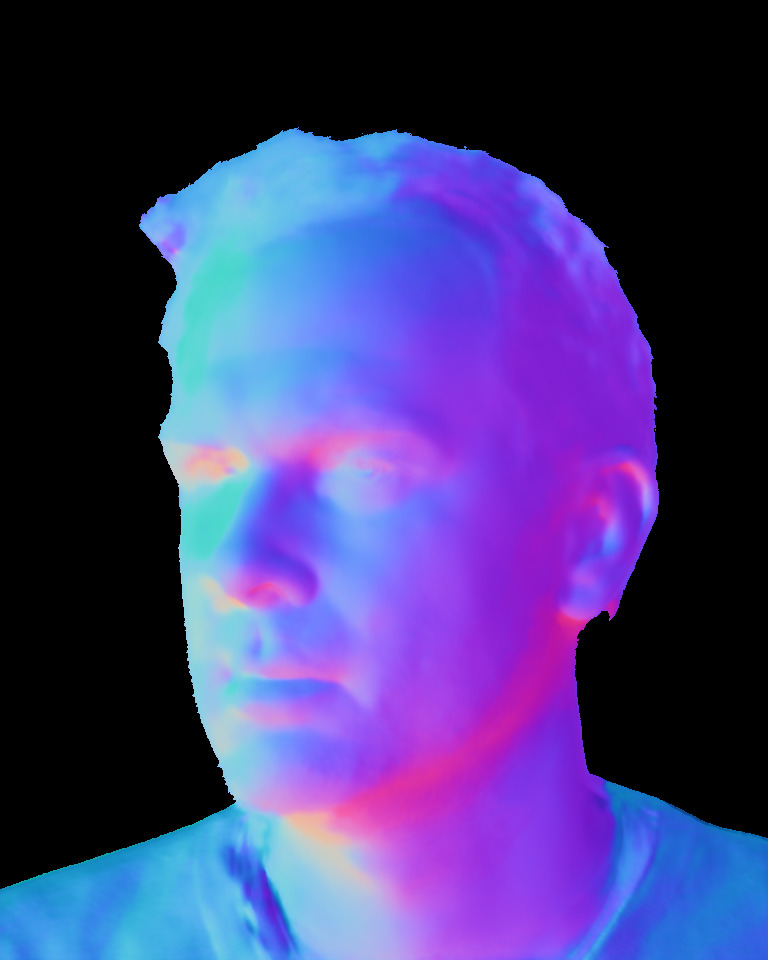} &
    \includegraphics[height=\imh]{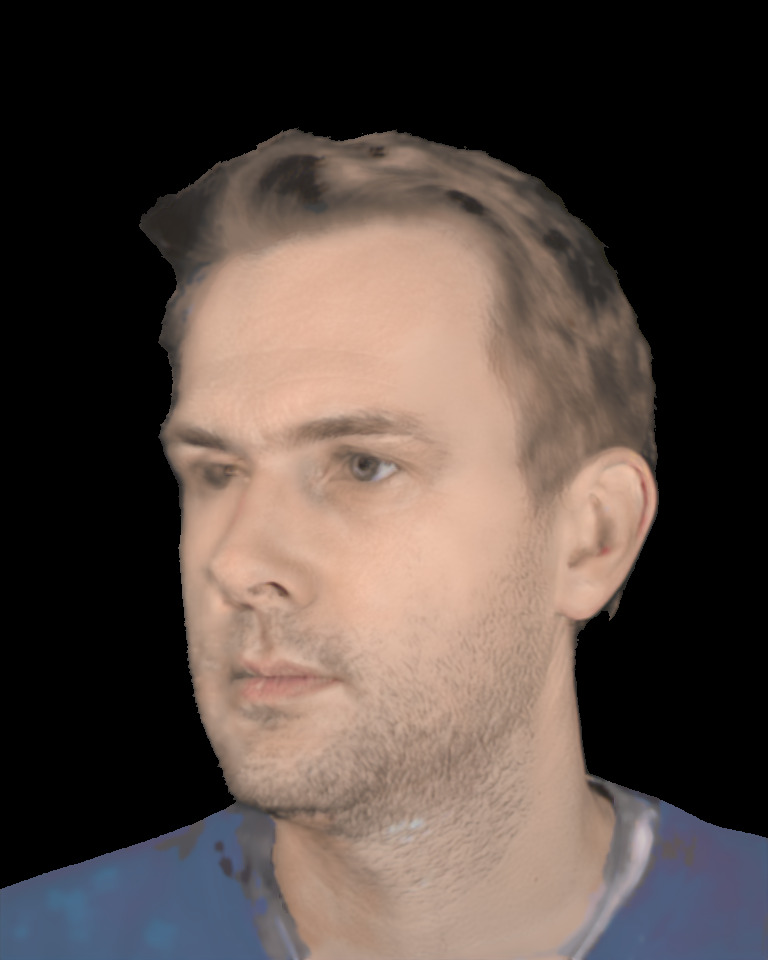} \\
    \small{RGB Input} & \small{Normals} & \small{Albedos} & \small{NIR Input} & \small{Normals} & \small{Albedos} \\
    & \small{(RGB Only)} & \small{(RGB Only)} & & \small{(Ours)} & \small{(Ours)} \vspace{-5pt}
    \end{tabular}
    \vspace{0pt}
    \caption{Comparison of our network to a modified version that takes only a single RGB image (``RGB Only'') as input. Example results for three common challenging lighting conditions. Top to bottom: low light / noisy inputs; mixed light colors; harsh directional lighting with saturated intensities. The ``RGB only'' network struggles to produce stable normal and reflectance estimates from these inputs in contrast to our method.}
    \label{fig:lighting_invariance_examples}
\end{figure*}

\section{Evaluation}
\label{section:Evaluation}

We evaluate our method through several ablation studies that measure the impact of key design decisions, including an in-depth comparison to a modified network that takes only a single RGB image as input. We also present two applications of our technique. Please see our supplemental video for results on image sequences and our supplemental document for additional comparisons to SfSNet~\cite{Sengupta2018SFSNet} and the face relighting network of Nestmemery et al.~\cite{Nestmeyer2020FaceRelighting}.

In our evaluations we consider five different visible lighting conditions: harsh lighting that produces strong cast shadows; a mixture of lights with different color temperatures; saturated/overexposed intensities; low-light conditions that produce noisy inputs; and a ``well lit'' condition that achieves largely shadow-free and well exposed inputs. Our process for synthesizing these different lighting conditions from the OLAT training images is detailed in our supplemental document. All of our results were produced using these five lighting conditions with the exception of Figure~\ref{fig:teaser}, which shows the performance of our method in lighting conditions that do not lie in the span of our OLAT images. That sequence was captured by casually moving a handheld light source around the scene. Please see our supplemental video for an expanded version of that result. None of the subjects shown in any of our results were included in our training set.

In lieu of ground truth geometry for quantitative assessments, we construct a baseline using the technique of Nehab et al.~\cite{Nehab2005PositionNormals} to refine our stereo depth maps according to normals computed by applying Lambertian photometric stereo to the RGB OLAT training images.

\subsection{Ablation Studies}
\label{subsection:Ablation}

Table~\ref{table:ablation} reports the mean absolute angular errors with respect to our baseline for normal maps computed with variants of our network that have different loss terms, image formation models, and inputs. Figures~\ref{figure:ablation} and~\ref{fig:lighting_invariance_examples} show examples of the perceptual impact of some of these design decisions. \\

\noindent \textbf{Loss terms.} As expected, using both stereo and photometric loss terms during training outperforms using either one alone. We consider two types of photometric loss - one computed on only the RGB training images (``No NIR Photometric Loss'' in Table~\ref{table:ablation}) and the second computed on both the NIR and RGB training images (``Full Method''). As illustrated in the shading images in Figure~\ref{figure:ablation}, including the photometric loss enables estimating fine geometric details that are not captured in the stereo depth maps. \\

\noindent \textbf{Image formation model.} Including the Blinn-Phong BRDF in our image formation model improves the accuracy of the normals and diffuse albedo maps. It results in a modest improvement in the quantitative errors in Table~\ref{table:ablation}, and it produces more uniform diffuse albedo maps with fewer artifacts (Figure~\ref{figure:ablation}). We attribute this to the fact that this richer image formation model is better able to explain the observed intensities. We also found that including this BRDF in our model enables reconstructing the glossy appearance of skin (Section~\ref{subsection:relighting}). \\

\noindent \textbf{Network inputs.} Including the NIR input image improves accuracy across the board, especially in poor visible lighting conditions (Table~\ref{table:ablation}). The benefit of the RGB input is comparatively smaller, but making it available to the network enables estimating visible spectrum reflectance data, which is a requirement for many downstream applications such as lighting adjustment (Section~\ref{subsection:relighting}). Figure~\ref{fig:lighting_invariance_examples} illustrates the perceptual impact of including the NIR input in different lighting conditions. For these comparisons we modified our network to take only a single RGB image as input (``RGB Only''). The network architecture was otherwise unchanged, and we applied the same training procedure described in Section~\ref{section:network}. Note how the performance of this ``RGB Only'' network significantly degrades in challenging conditions, while our method is far more robust to these conditions due to the more stable NIR input. It's particularly noteworthy how well our method is able to reconstruct plausible diffuse albedos even for highly saturated RGB input images (bottom row of Figure~\ref{fig:lighting_invariance_examples}).

Please see our supplemental document for an expanded version of Figure~\ref{fig:lighting_invariance_examples} and our supplemental video that includes comparisons for animated image sequences.

\subsection{Application: Stereo Refinement}
\label{subsection:results_refinement}

\begin{figure}
    \newcommand{\imw}{0.45\columnwidth}
    \setlength{\tabcolsep}{5pt}
    \centering
    \begin{tabular}{cc}
    \includegraphics[width=\imw]{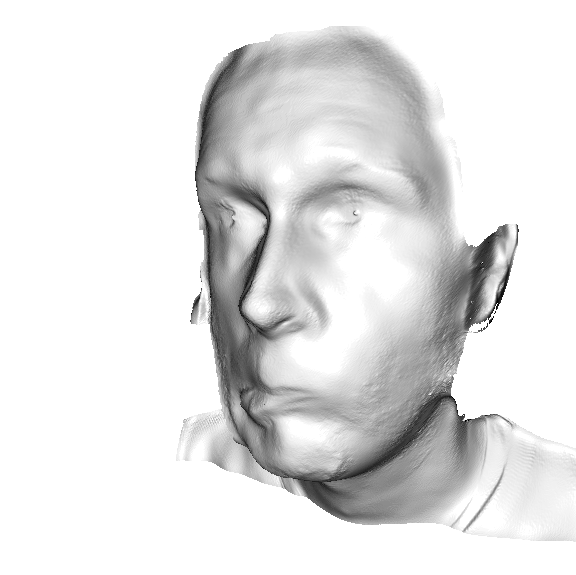} &
    \includegraphics[width=\imw]{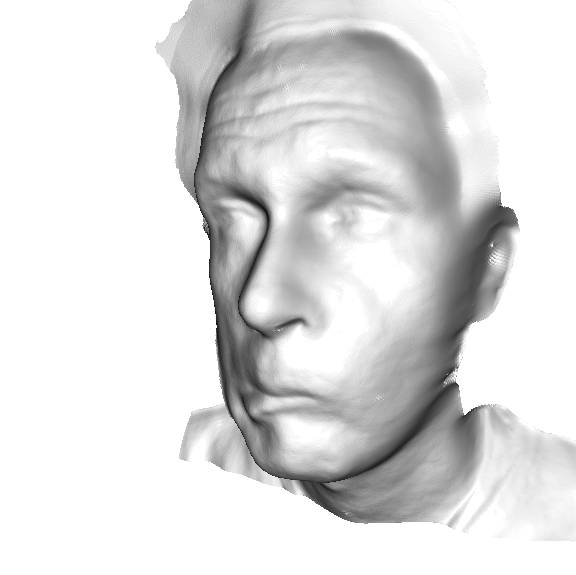} \\
    \includegraphics[width=\imw]{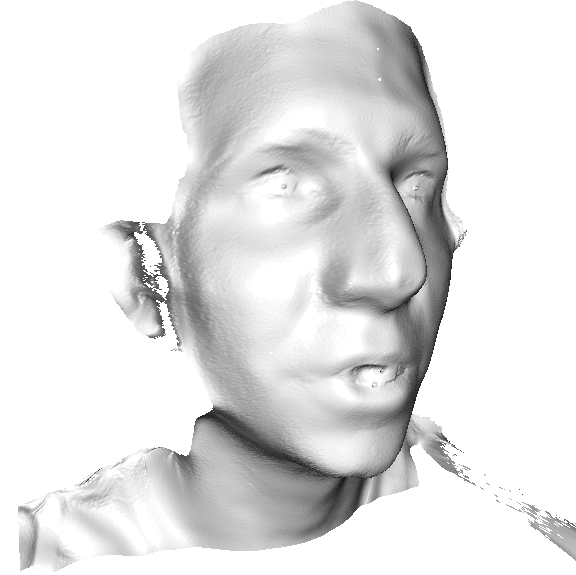} &
    \includegraphics[width=\imw]{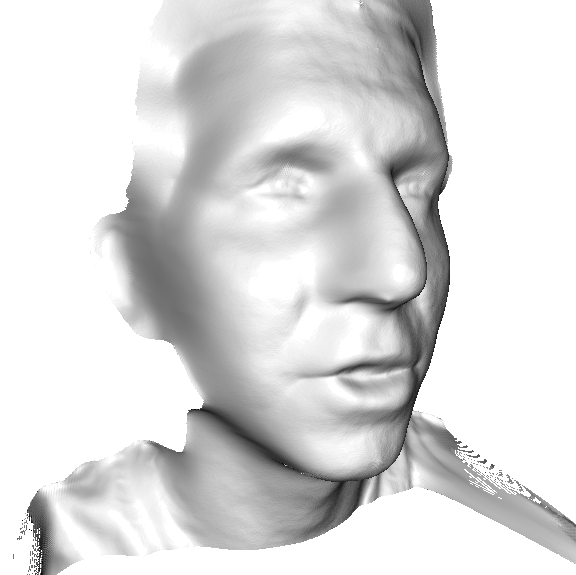} \\
    \small{Smoothed Stereo} & \small{Refined Stereo (Ours)} \\
    \end{tabular}
    \vspace{-5pt}
    \caption{Stereo methods often struggle to recover fine-scale surface details. \emph{Left:} Applying a guided bilateral filter to raw stereo depths yields a smoother surface but with distorted features (e.g. the nose is reduced and skin wrinkles are missing). \emph{Right:} We use the method of Nehab et al.~\cite{Nehab2005PositionNormals} to compute a refined surface according to normals estimated with our method. Note how details are better preserved around the eyes, nose, and mouth, along with fine wrinkles and creases.}
    \label{fig:refinement}
\end{figure}

Stereo methods excel at measuring coarse geometry, but often struggle to recover fine-scale surface details. This can be overcome by refining stereo depths according to accurate high-resolution normals typically estimated with a photometric approach~\cite{Nehab2005PositionNormals}. We evaluate using the normals produced by our method to refine depth measurements produced by an NIR space-time stereo algorithm~\cite{Nover2018Espresso} (Figure~\ref{fig:refinement}). In comparison to using a standard bilateral filter to smooth the stereo depths, refining them using our normals gives much higher quality reconstructions, most notably around the mouth, nose, and eyes and better recovery of fine wrinkles and creases in the skin. As our method works with a single NIR image it would be straightforward to integrate it into many existing stereo pipelines.

\subsection{Application: Lighting Adjustment}
\label{subsection:relighting}

\begin{figure}[!t]
    \newcommand{\imh}{3.4cm}
    \setlength{\tabcolsep}{2pt}    
    \centering
    \begin{tabular}{ccc}
    \includegraphics[height=\imh]{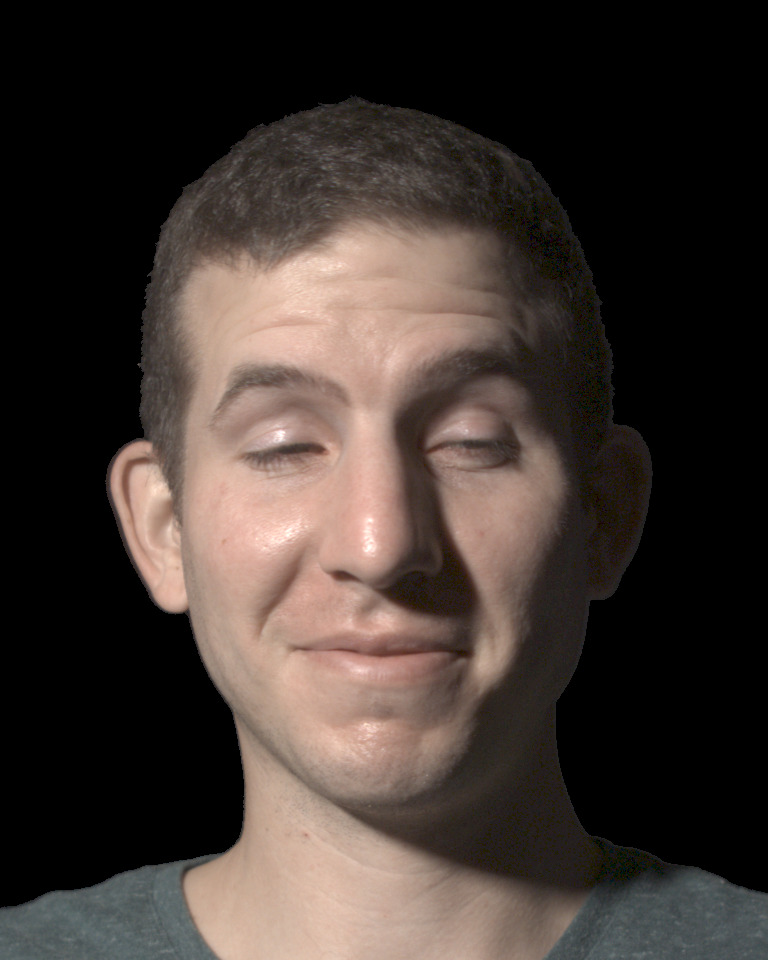} &
    \includegraphics[height=\imh]{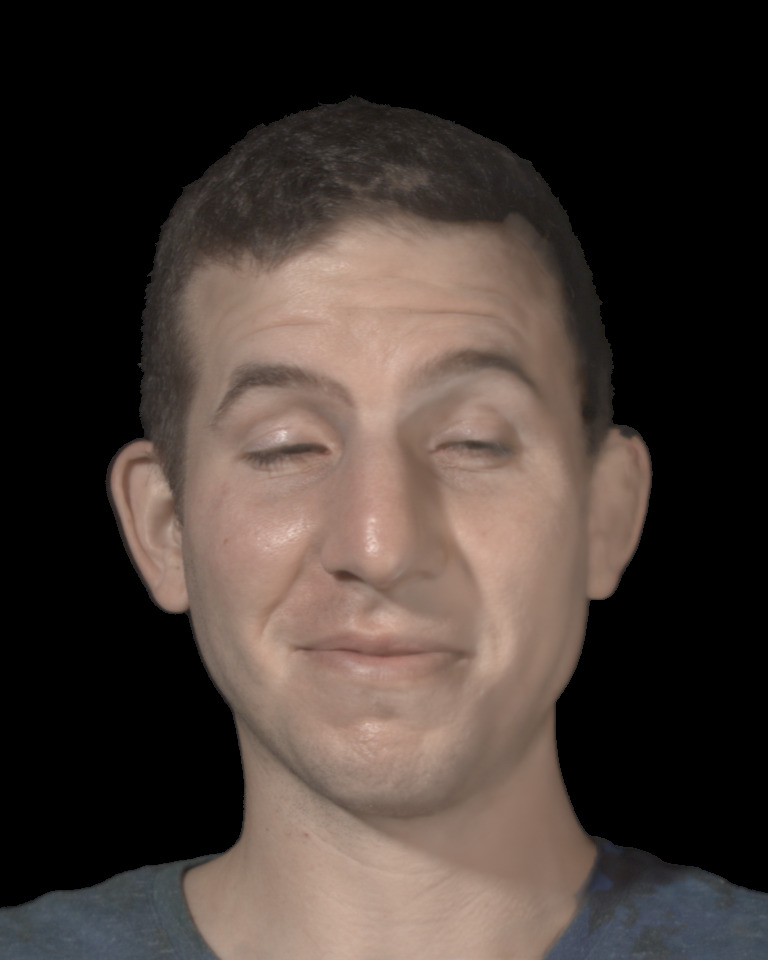} &
    \includegraphics[height=\imh]{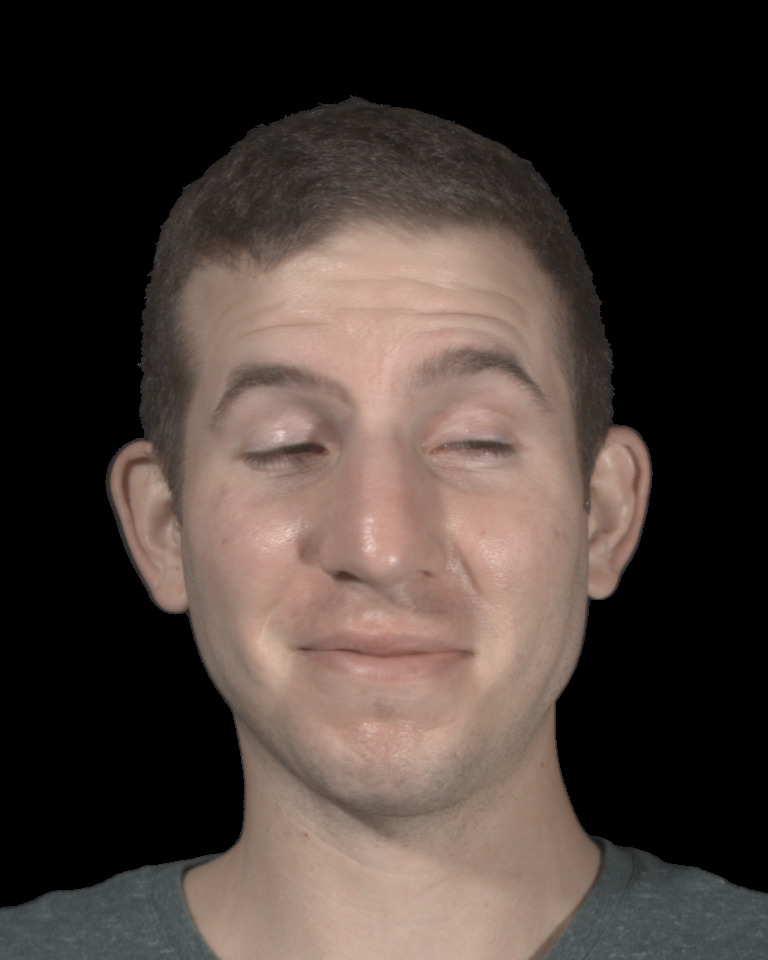} \\
    \includegraphics[height=\imh]{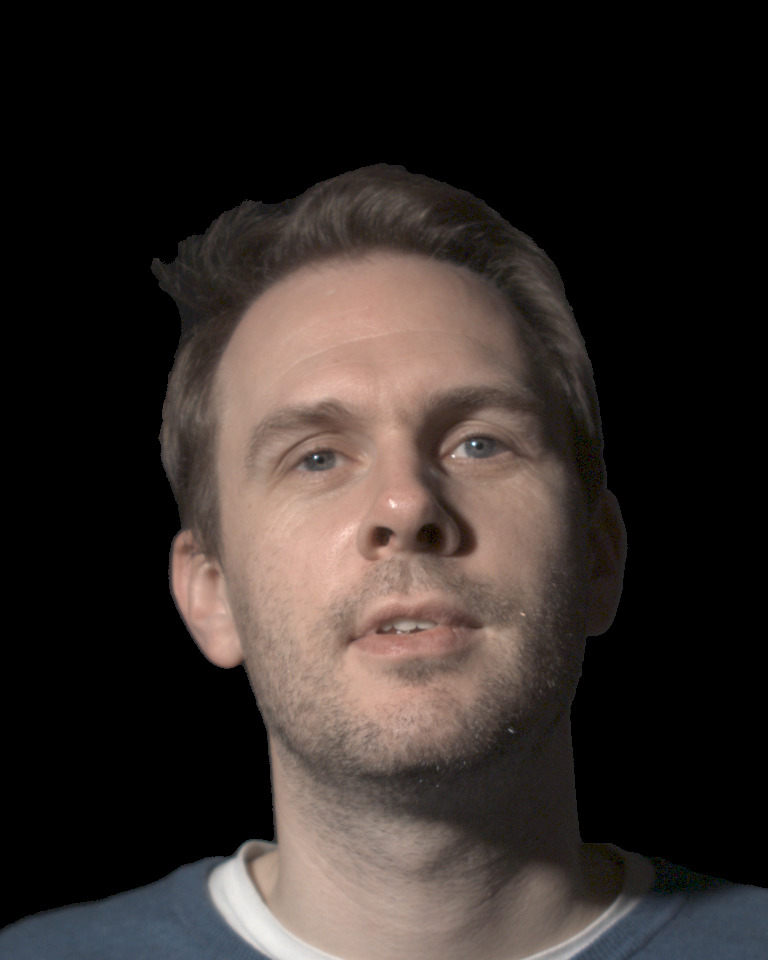} &
    \includegraphics[height=\imh]{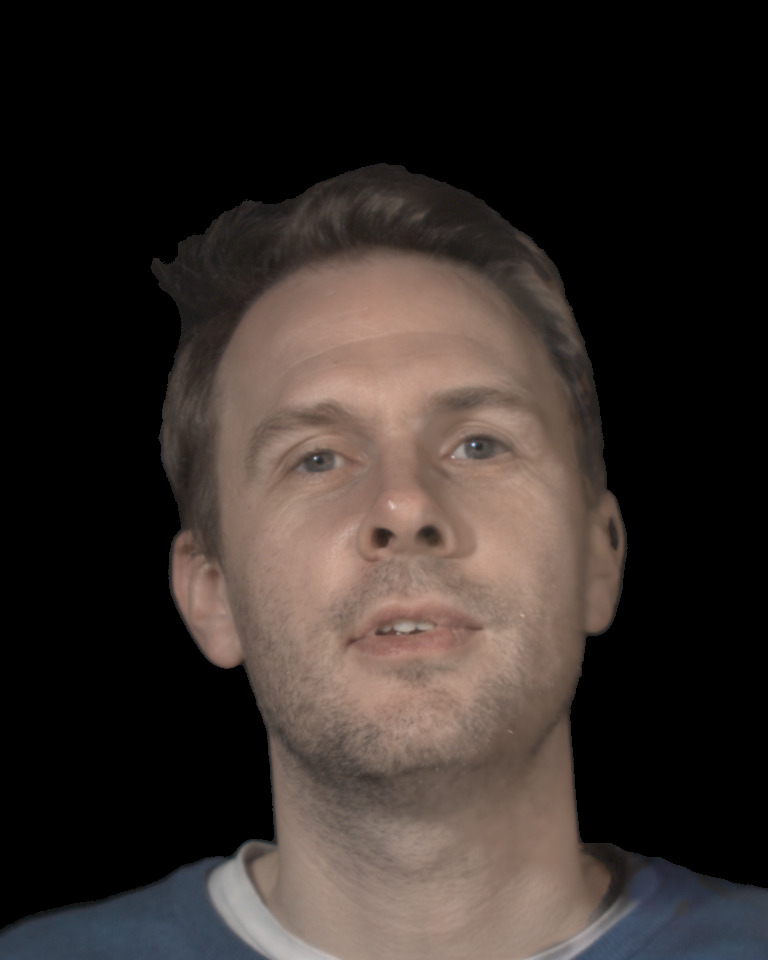} &
    \includegraphics[height=\imh]{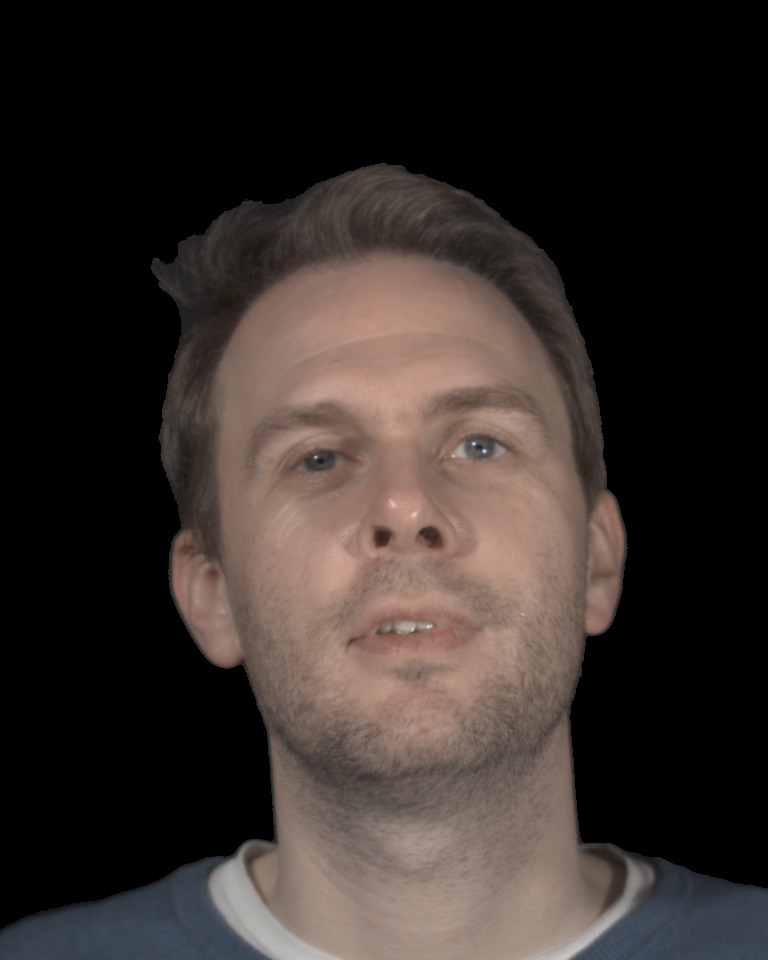} \\
    \small{RGB Input} &
    \small{Relit} &
    \small{Ground Truth}
    \end{tabular}
    \vspace{-5pt}
    \caption{Our method can be used to simulate adding lights to a scene to fill in shadows.
    }
    \label{fig:shadow_removal}
\end{figure}

We also explored using our approach to digitally improve the lighting in a portrait. Specifically, we evaluated adding a virtual fill light to brighten shadowed parts of the face (Figure~\ref{fig:shadow_removal}). We used normal and reflectance maps estimated by our method to render the contribution of a virtual point light located within view of the shadowed region, and then combined this with the original RGB image.
Our model enables a convincing effect, even producing realistic specular highlights along the nasolabial folds and the tip of the nose.

\section{Discussion and Future Work}
\label{sec:future_work}

We observed that the specular intensity maps produced by our method are not accurate across the entire face due to the sparse nature of our training images -- the specular component of the material appearance is simply not observed in many parts of the scene. We believe this could be overcome by including additional unique light positions in our training data or by introducing priors on this component.

We record co-axial color and NIR images using two cameras aligned with a beamsplitter. Future work could explore more practical hardware setups that potentially use integrated RGB+NIR image sensors~\cite{Tang2015HighResRGBIR, Hu2018SparseCodingRGBNIR} or RGB+NIR stereo setups that would allow placing the NIR and color cameras next to each other at a small baseline~\cite{Wang2019Stereoscopic}. This line of work could take advantage of the growing availability of NIR light sources and NIR sensors in modern smartphones.

Our approach assumes a single light located near the camera is the only source of NIR light in the scene. Although this is a safe assumption in many indoor environments, it is not always true, especially outdoors. It would be interesting to explore overcoming ambient NIR light, possibly through the use of flash/no-flash image pairs.

Another area of future work is improving our method's performance on hair and clothing. An effective strategy would be to expand the training data to include a larger number of subjects and more light positions in the OLAT training sequences. Finally, there is room to improve the temporal stability of our method by having the network explicitly consider consecutive frames. Please see the supplemental video that includes results for image sequences.

\section{Conclusion}

We have presented a dark flash normal camera that is capable of estimating high-quality normal and reflectance maps from a single RGB+NIR input image that can be recorded in a single exposure without distracting the subject. A key benefit of our method over prior work is its robustness. It performs well even in challenging lighting conditions that are commonly encountered in casual photography such as harsh shadows, saturated pixels, and in very low light environments. Our method could easily be integrated into existing smartphone camera hardware designs and software pipelines to enable a range of applications from refining the output of an auxiliary depth camera to improving the lighting of faces in still images and streaming video.

{\small

}

\clearpage
\onecolumn
\pagestyle{plain}
\pagenumbering{roman}

\appendix
\begin{center}{
    \LARGE \bf Supplementary Material\vspace{3em}
  }\end{center}

\appendix

\section{Visible Lighting Conditions Used in Our Evaluation}
\label{sec:light}
\begin{figure}[!h]
    \newcommand{\imh}{4.2cm}
    \setlength{\tabcolsep}{2pt}
    \centering
    \begin{tabular}{ccccc}
    \includegraphics[height=\imh]{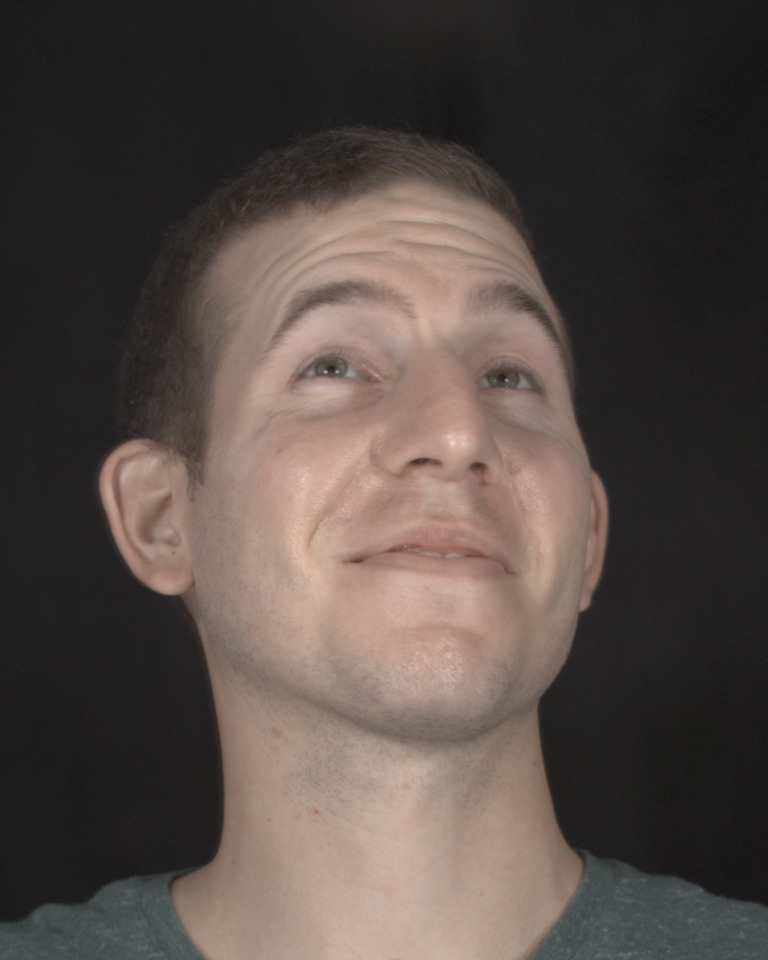} &
    \includegraphics[height=\imh]{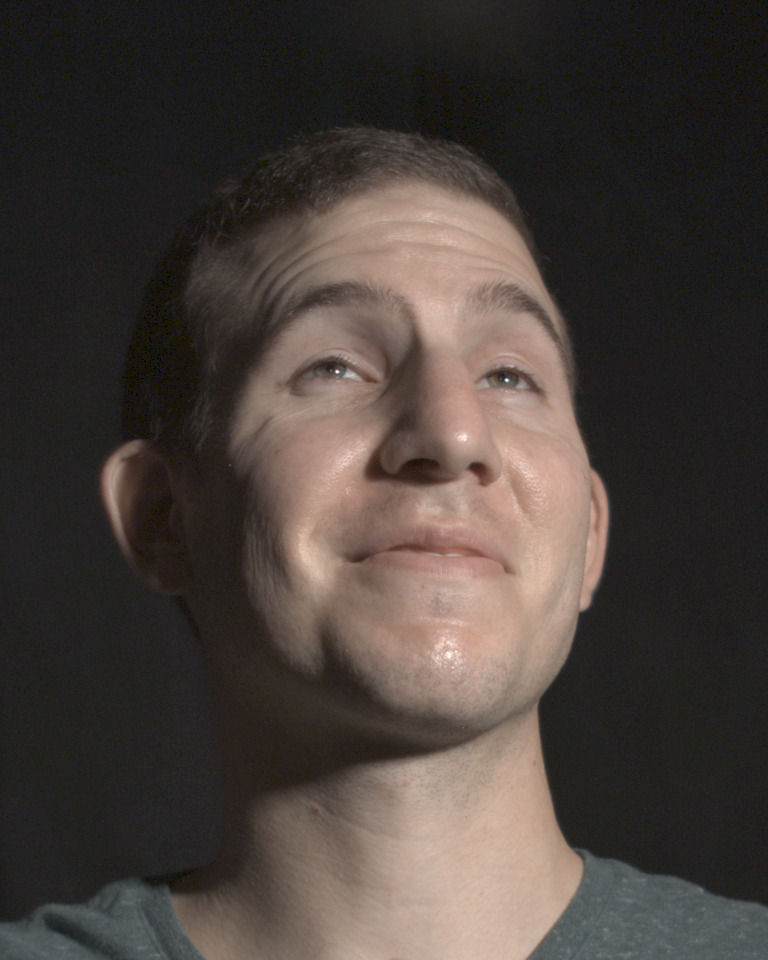} &
    \includegraphics[height=\imh]{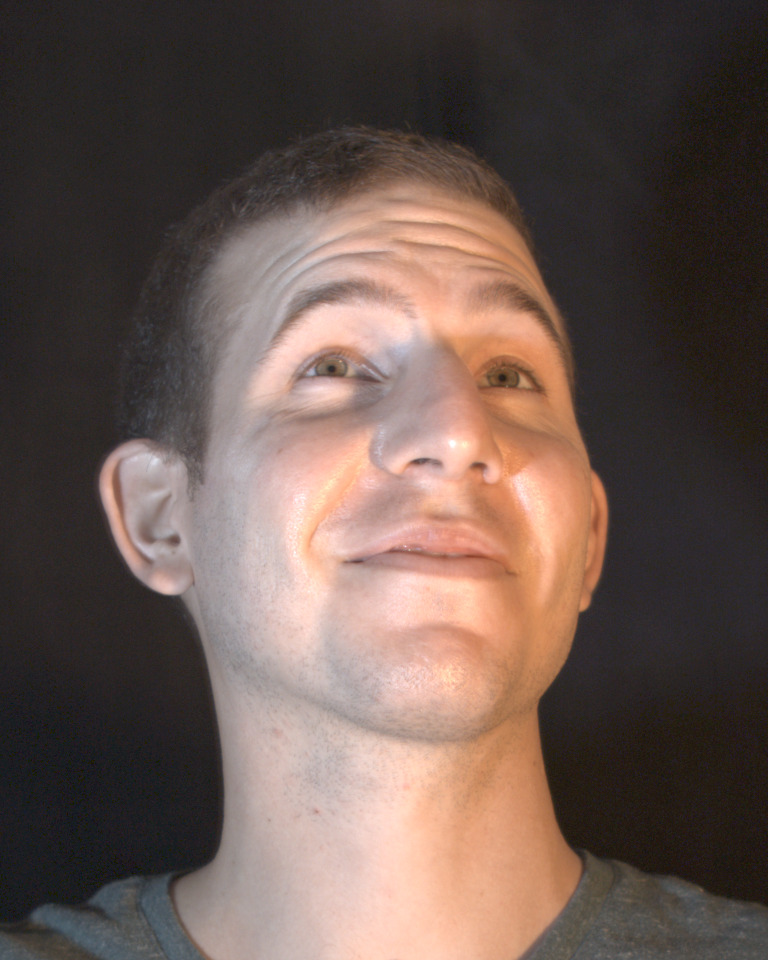} &
    \includegraphics[height=\imh]{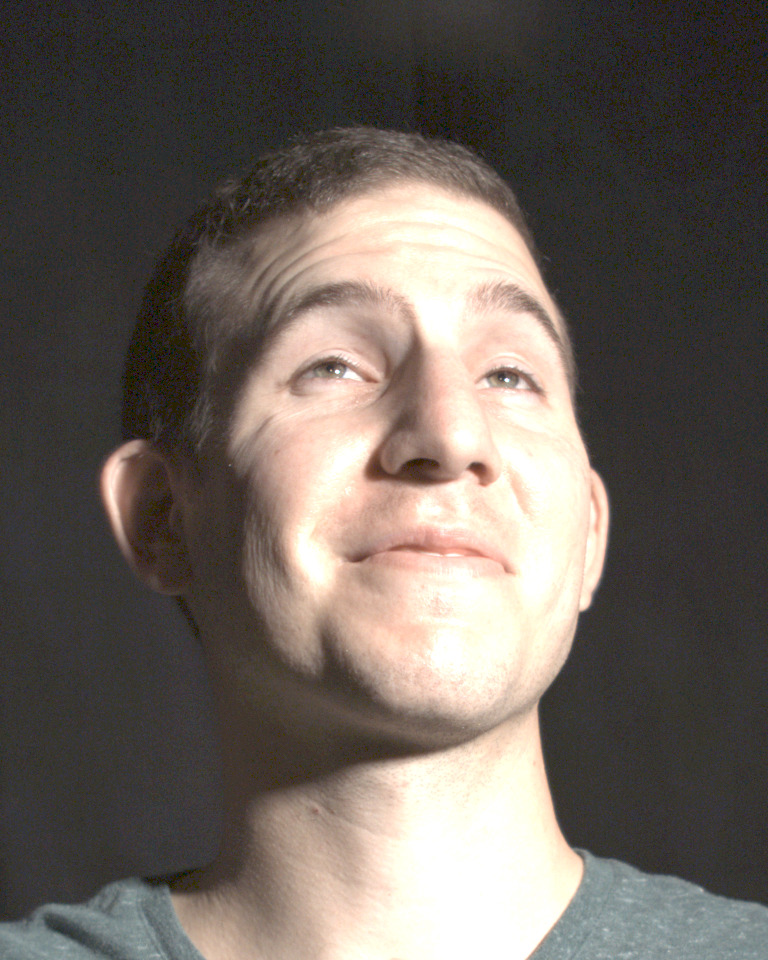} &
    \includegraphics[height=\imh]{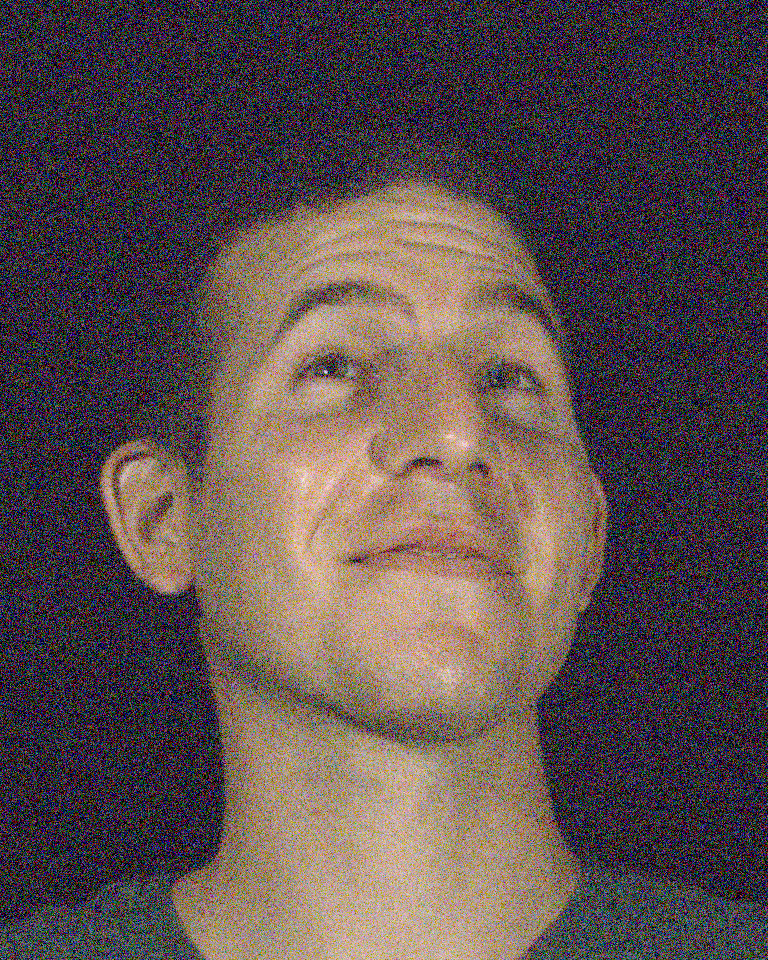} \\
    \small{(a) Well lit} & \small{(b) Shadows} & \small{(c) Mixed colors} & \small{(d) Overexposure} & \small{(e) Low light} \\
    \end{tabular}
    \vspace{-3pt}
    \caption{The five visible lighting conditions used in our evaluations. (a) A well lit image is created by averaging equally the four OLAT RGB images, using weights that avoid any under- or over-saturated pixels. (b) Picking a single OLAT image at random produces an RGB input with strong cast shadows. (c) We simulate mixing different colored lights by picking two OLAT images at random and remapping the color of each to a random color temperature in the range $[1900K, 2900K]$ and $[7000K, 20000K]$, respectively, and then averaging them together. (d) Scaling the intensities of one of the OLAT images with a random scale factor in the range $[1.8, 2.3]$ and then clipping the result yields an image with harsh lighting and saturated intensities. (e) Adding Gaussian white noise to each pixel ($\sigma=25$ 8-bit gray levels) simulates images captured under low light conditions.}
    \label{fig:lighting_type_examples}
\end{figure}

We evaluate our method using five different visible lighting conditions that range from favorable (``well lit'') to challenging (``shadows'', ``mixed colors'', ``overexposure'', and ``low light''). Figure~\ref{fig:lighting_type_examples} illustrates and details how these different lighting conditions are simulated from the RGB OLAT training images.

\section{Additional Ablation Studies}

\begin{figure*}[!h]
    \centering
    \newcommand{\imw}{0.32\textwidth}
    \setlength{\tabcolsep}{2pt}
    \begin{tabular}{ccc}
    \includegraphics[width=\imw]{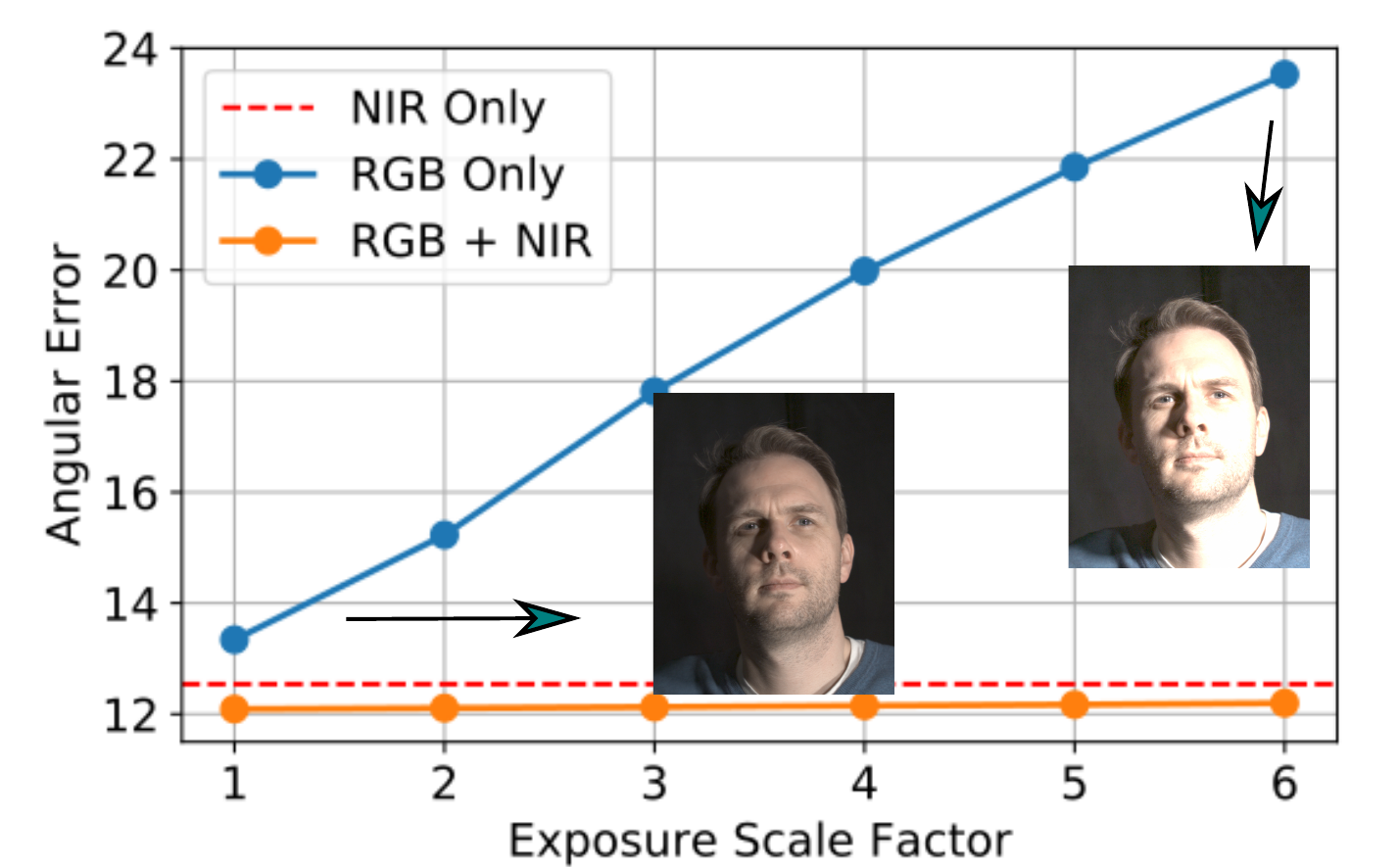} &
    \includegraphics[width=\imw]{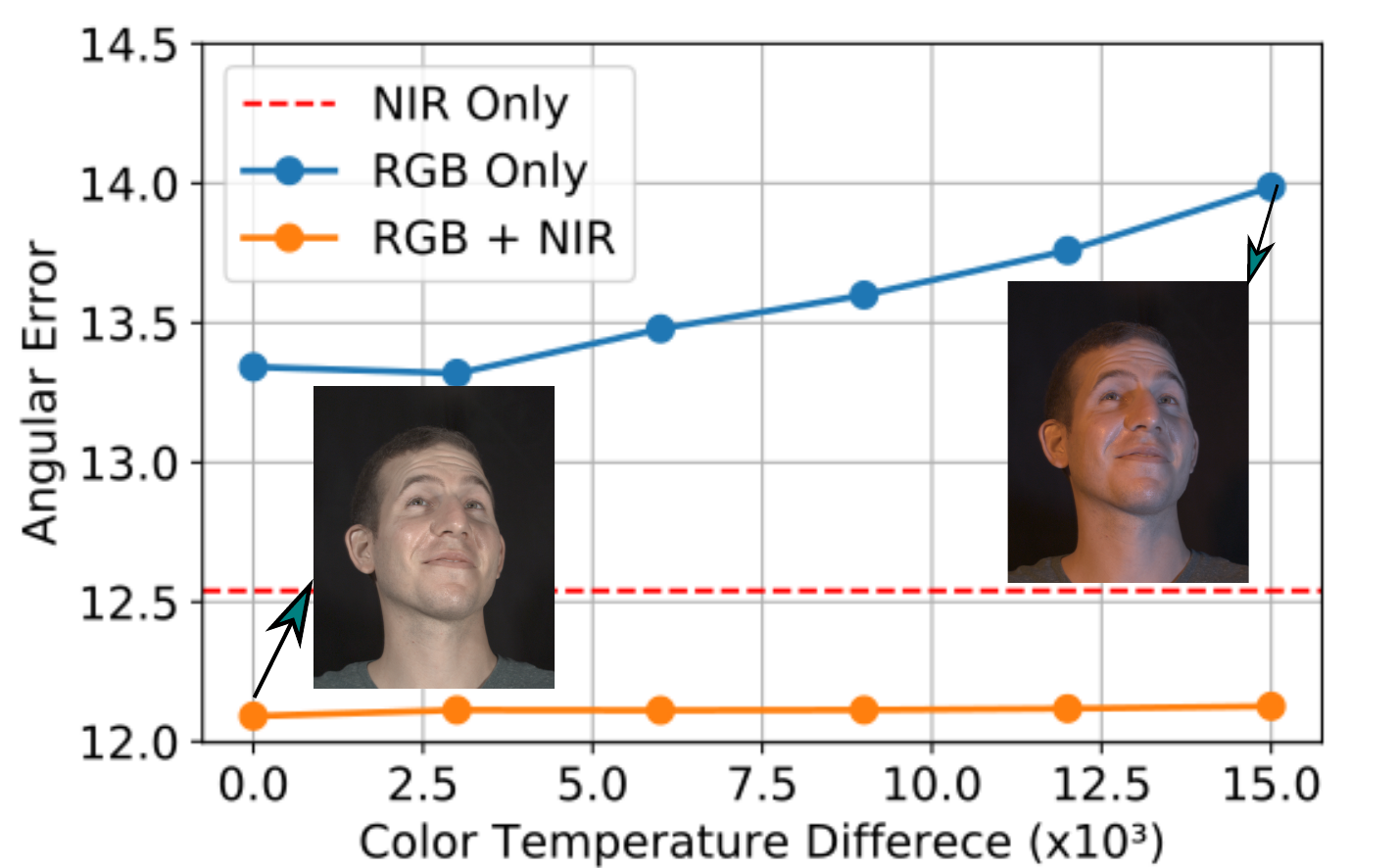} &
    \includegraphics[width=\imw]{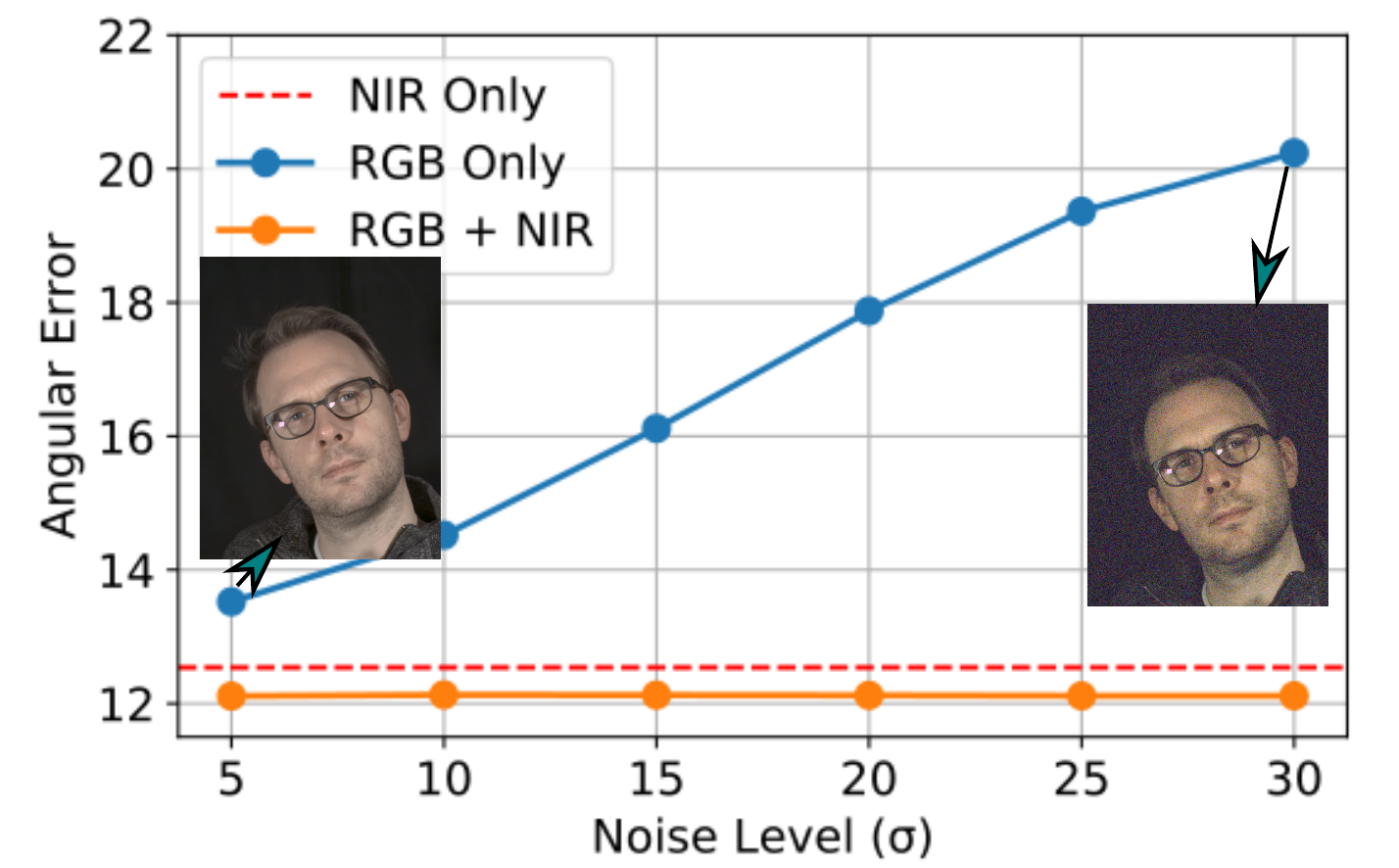}
    \end{tabular}
    \vspace{-3pt}    
\caption{Mean absolute angular errors in degrees of normal maps estimated by three different techniques in three challenging lighting conditions as the magnitude of each lighting issue increases from left to right. A similar network that uses only a single RGB image to estimate normals (RGB Only) and one that uses only a single NIR image (NIR Only) both perform worse that our method (RGB + NIR), with the RGB Only network deteriorating rapidly alongside the lighting issue. \emph{Left:} harsh lighting that produces cast shadows with increasing image exposures. \emph{Center:} mixing lights with different color temperatures. \emph{Right:} increasing levels of noise, which occurs in low-light conditions.}
\label{figure:lighting_invariance}
\end{figure*}

The graphs in Figure~\ref{figure:lighting_invariance} expand on the results shown in Figure~\ref{fig:lighting_invariance_examples} in the paper. They plot the mean angular error of the estimated normals against the baseline for three of our five lighting scenarios, and for three different technique{}s: our network modified to take only a single NIR input image (``NIR Only''), our network modified to take only a single RGB input image (``RGB Only''), and our full network, which considers both. In each graph the magnitude of the lighting issue increases from left to right.

Variations in illuminant color have a relatively small effect on the normals estimated by the ``RGB only'' network, but overexposed pixels and image noise cause large errors. As one would expect the ``NIR only'' network is invariant to these lighting changes (flat line in these graphs), and our proposed approach consistently outperforms both.

\section{Comparisons to Prior Single (RGB) Image Normal Estimation Methods}

To the best of our knowledge, our method is the first technique for estimating surface normals and RGB albedos from a \emph{single} RGB+NIR image. While the ``RGB Only'' baseline in the main paper demonstrates the effectiveness of supplementing the available visible illumination with a dark flash NIR image, we now provide a comparison to state-of-the-art single RGB image normal estimation methods from the literature. We compare our method to two recent techniques:

\paragraph{SfSNet~\cite{Sengupta2018SFSNet}.} For a fair comparison, we retrain SfSNet on our dataset. We use the same basic network architecture as our method, but with only a single RGB image as input and with an additional lighting estimation branch. This lighting estimation branch takes the output of the encoder network and generates an estimate of the lighting in the input RGB image. SFSNet uses second order spherical harmonics to represent the scene lighting, which isn't well suited for approximating the type of point lighting in our dataset. Therefore we instead represent the lighting as a blending weight vector over the four visible OLATs. The lighting estimation branch is supervised using the ground truth OLAT mixing weights of the input RGB image. The normal branch of SfSNet is supervised using the same stereo normals and same image reconstruction loss that we use for our method.

\paragraph{Directional Face Relighting Network~\cite{Nestmeyer2020FaceRelighting}.} We also compare our method to the intrinsic component estimation stage of Nestmeyer et al.~\cite{Nestmeyer2020FaceRelighting}, which uses a UNet to predict a normal and albedo map from a single RGB image. Since we assume that the lighting conditions in the input RGB image are unknown, we do not provide the source lighting direction to the network. We retrain their network on our dataset by supervising the normal estimation network path with the same stereo normals used to train our method, and by using one of the RGB OLAT images chosen at random as the relighting target image. Note that their instrinsic component estimation stage does not consider cast shadows.\\

Table~\ref{table:baselines} reports quantitative errors in the normals estimated using all three techniques for the different lighting conditions described in Section~\ref{sec:light}. Our method outperforms both techniques even in the well lit condition, which we attribute to our novel training strategy that combines shape information from complementary stereo and photometric signals, and the additional information provided by the NIR input. In challenging lighting conditions, the benefit of our method becomes more significant. Note that SfSNet~\cite{Sengupta2018SFSNet} uses a self-reconstruction loss that we found could not handle inputs with mixed color casts, saturated intensities, or a significant amount of noise and so it fails to produce plausible outputs in these cases (omitted from Table~\ref{table:baselines}).

\begin{table*}[!h]\small
  \begin{center}
  \setlength\tabcolsep{4pt}
  \begin{tabular}{|c|ccccc|}
  \hline
  Method & Well lit  & Shadows  & Mixed colors  & Overexposure & Low light \\ \hline
  SfSNet \cite{Sengupta2018SFSNet} & 14.10 & 18.32 & - & - & -\\ \hline
  Nestmeyer et al.\cite{Nestmeyer2020FaceRelighting} & 14.82 & 17.52 & 15.87 & 21.85 & 25.56 \\ \hline
  Ours & \textbf{12.29} & \textbf{12.27} & \textbf{12.28} & \textbf{12.30} & \textbf{12.33} \\ \hline
  
  \end{tabular}
  \end{center}
  \vspace{-10pt}
  \caption{Mean absolute angular errors in degrees for three normal estimation methods and five different lighting scenarios. Results for SfSNet~\cite{Sengupta2018SFSNet} are omitted in some cases where they fail to generate plausible normals.}
  \label{table:baselines}
\end{table*}

\section{Expanded Stereo Refinement Results}

\begin{figure*}
    \newcommand{\imw}{0.155\textwidth}
    \setlength{\tabcolsep}{1pt}
    \centering
    \begin{tabular}{cccccc}
    \includegraphics[width=\imw]{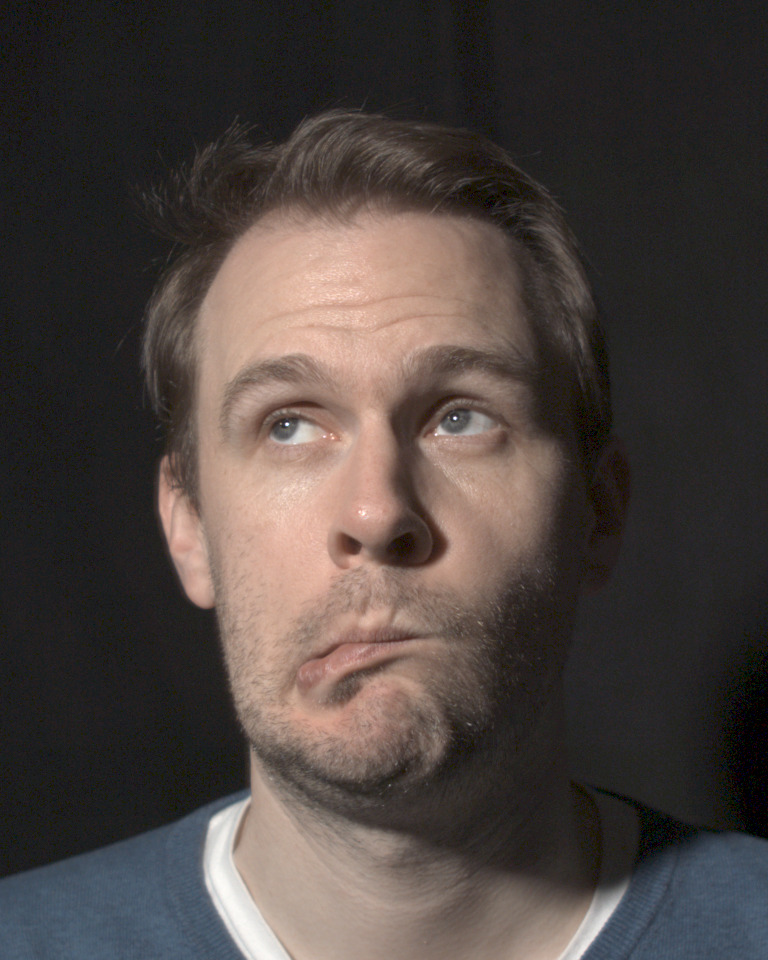} &
    \includegraphics[width=\imw]{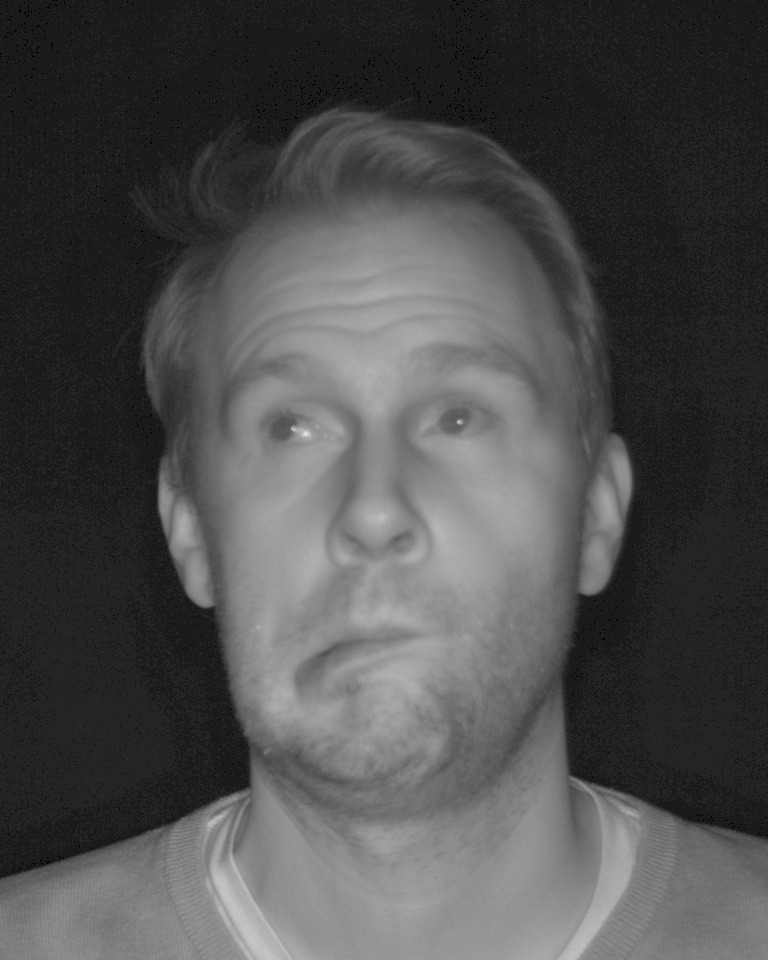} &
    \includegraphics[width=\imw]{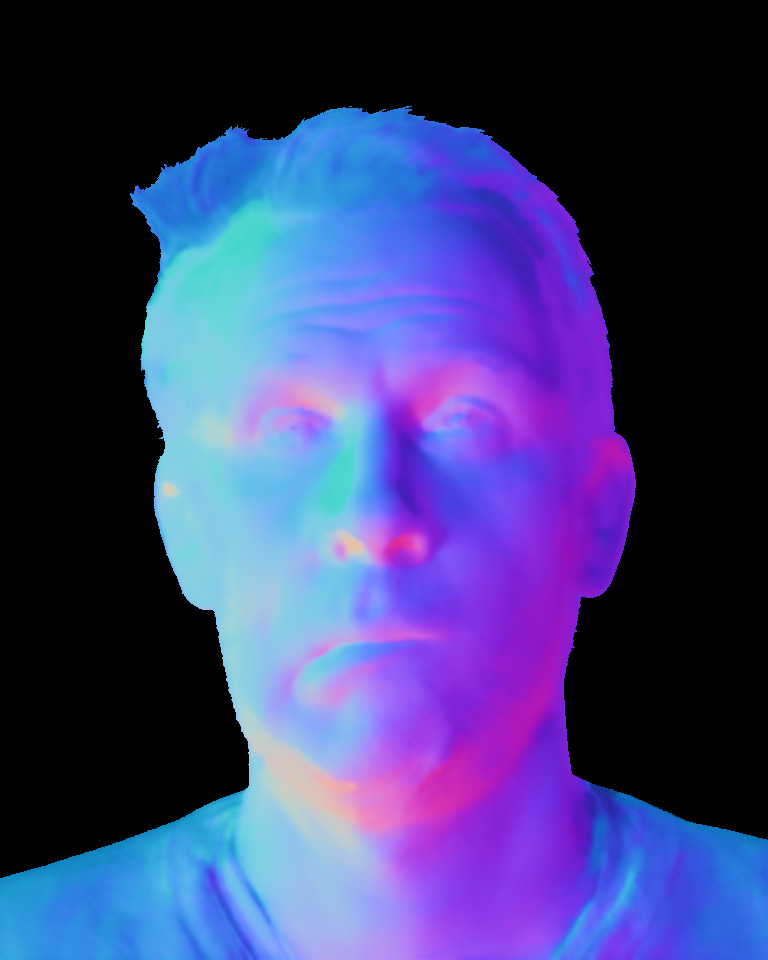} &
    \includegraphics[width=\imw]{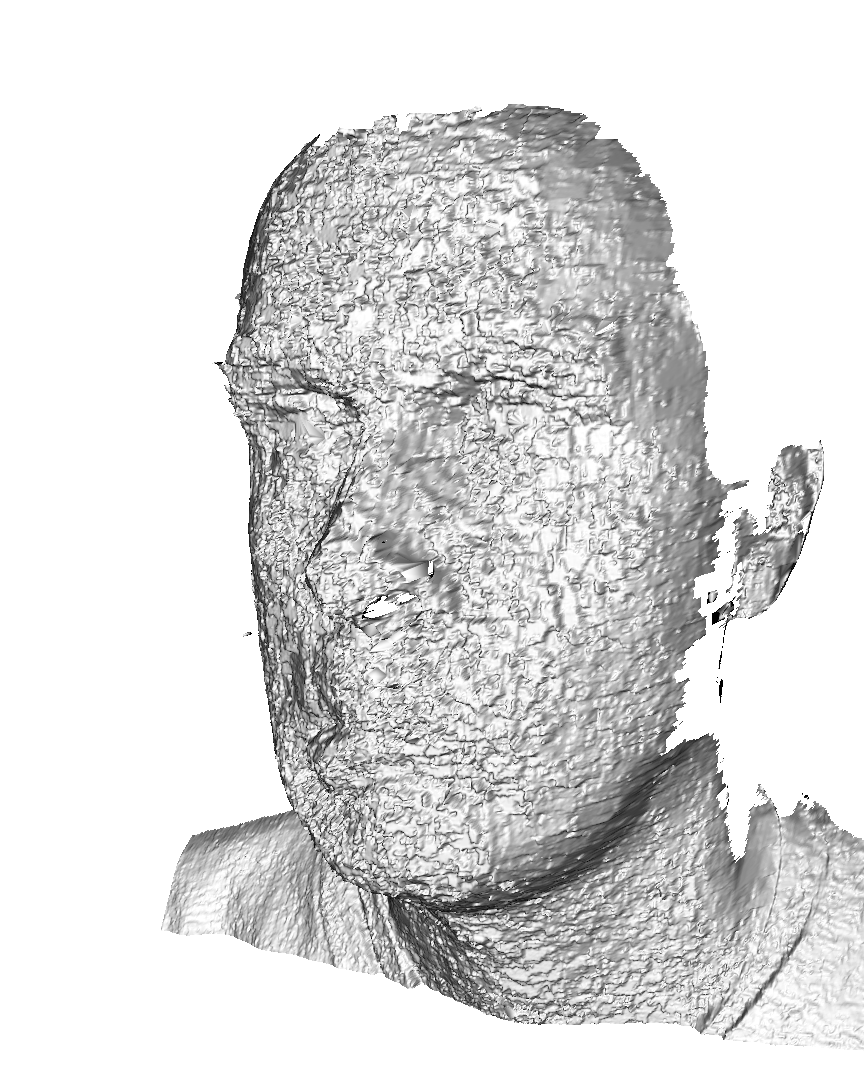} &
    \includegraphics[width=\imw]{figures/refinement/mesh_smooth01.png} &
    \includegraphics[width=\imw]{figures/refinement/mesh_refined01.png} \\
    \includegraphics[width=\imw]{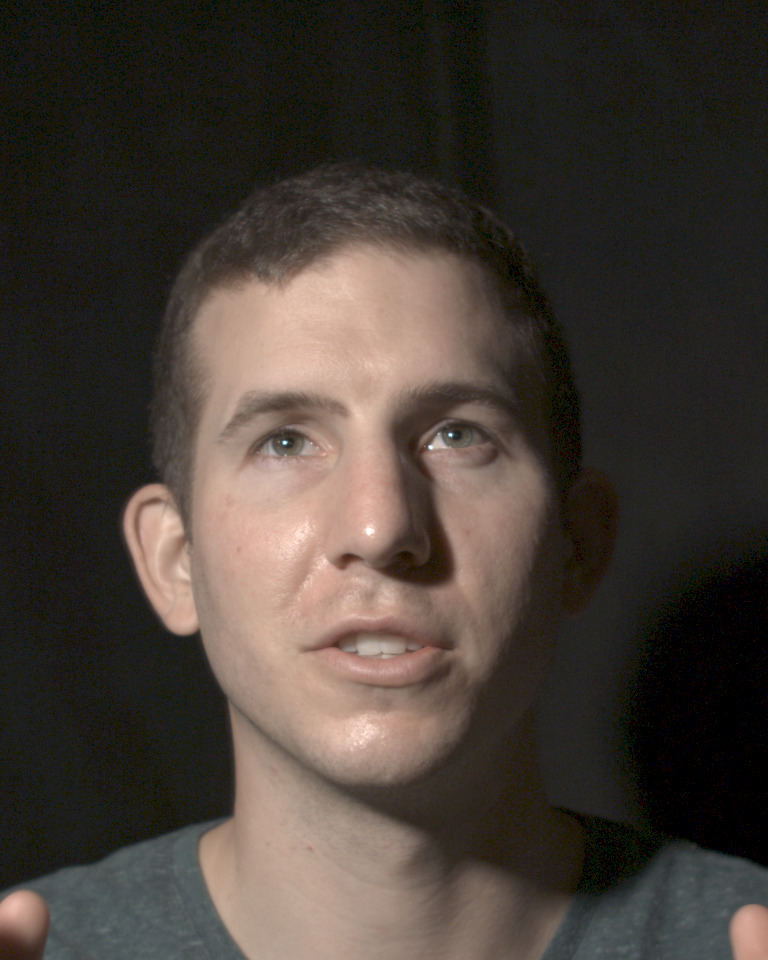} &
    \includegraphics[width=\imw]{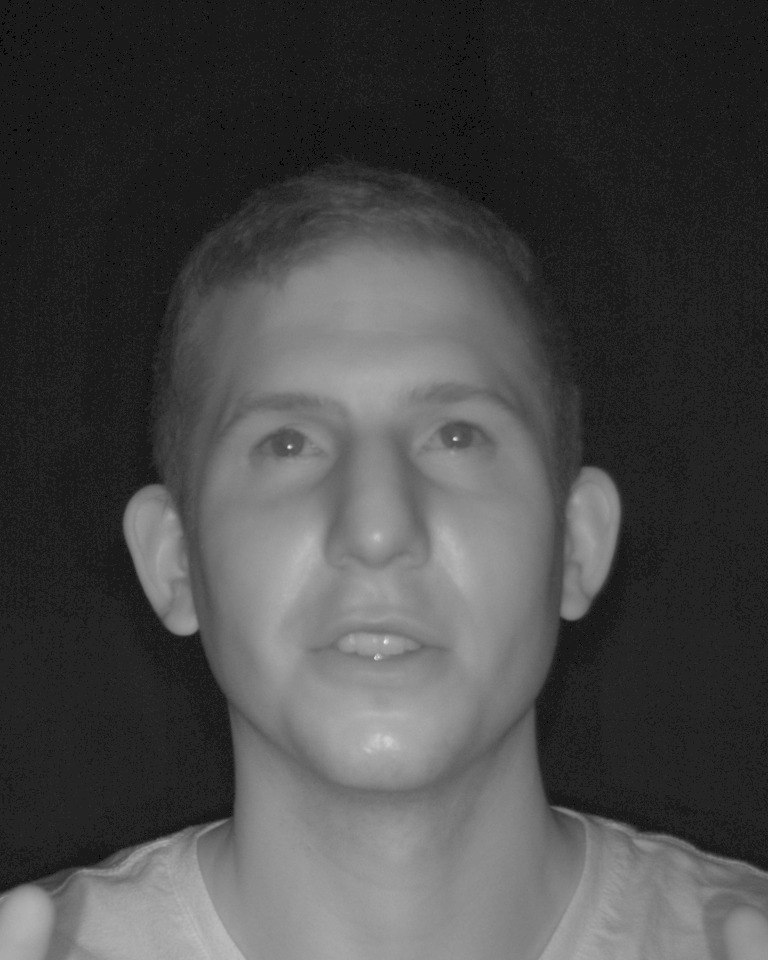} &
    \includegraphics[width=\imw]{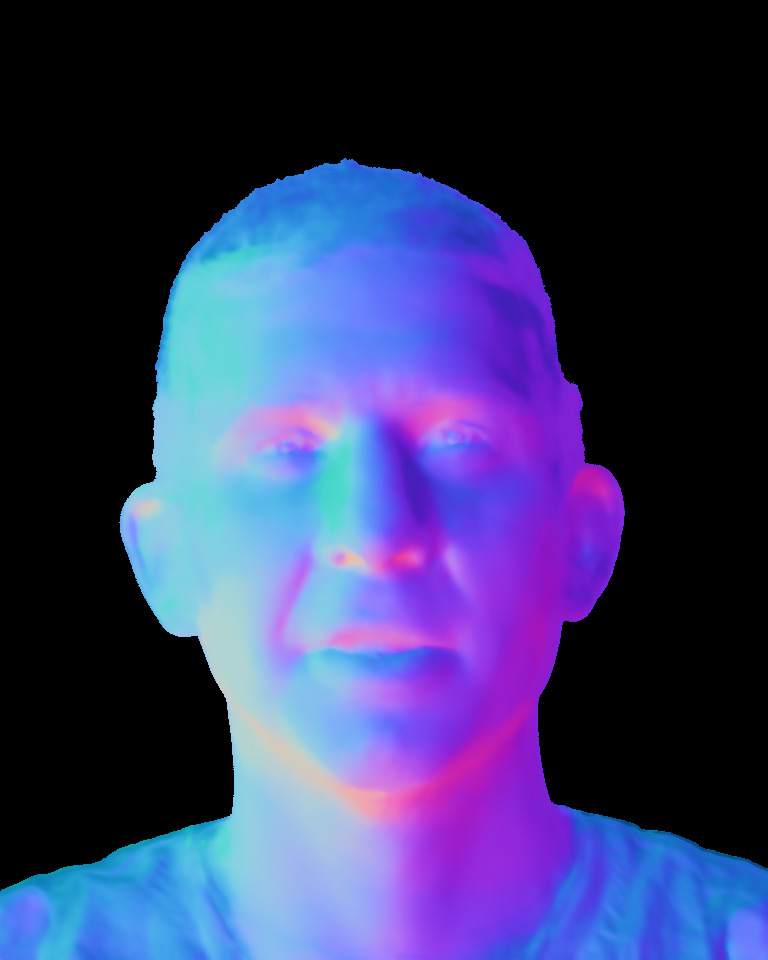} &
    \includegraphics[width=\imw]{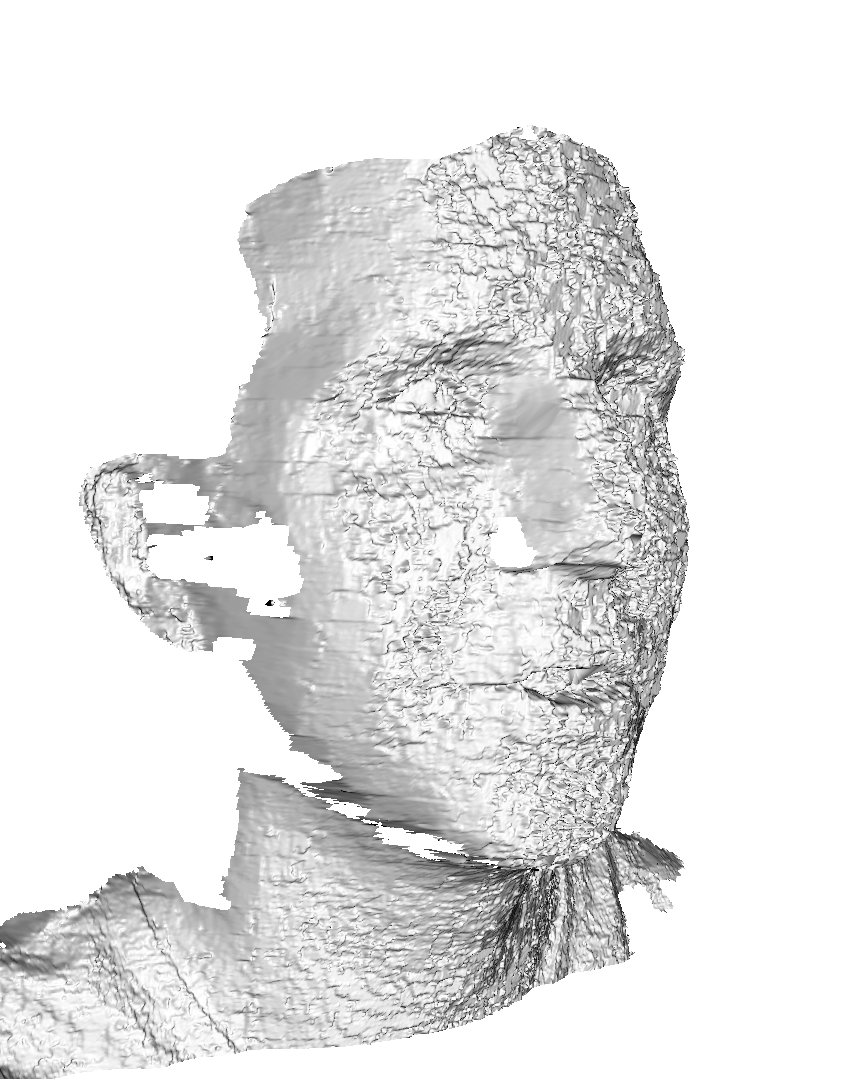} &
    \includegraphics[width=\imw]{figures/refinement/mesh_smooth02.png} &
    \includegraphics[width=\imw]{figures/refinement/mesh_refined02.png} \\
    \small{RGB Input} & 
    \small{NIR Input} & 
    \small{Normals} & 
    \small{Raw Stereo} & 
    \small{Smoothed} & 
    \small{Refined (ours)}
    \end{tabular}
    \vspace{-5pt}
    \caption[]{Stereo methods often struggle to recover fine-scale surface details. Applying a guided bilateral filter to raw stereo depths yields a smoother surface but with distorted features (e.g. the nose appears pinched and reduced). We use the method of Nehab et al.~\cite{Nehab2005PositionNormals} to compute a refined surface according to normals estimated with our method. Note how this better preserves details around the eyes, nose, and mouth, along with fine wrinkles and creases.}
    \label{fig:refinement_expand}
\end{figure*}

Figure~\ref{fig:refinement_expand} expands on Figure~\ref{fig:refinement} in the paper. Specifically it shows the input images and the normal map estimated by our method, along with the raw stereo depths prior to any smoothing or refinement.

\section{Expanded Lighting Adjustment Results}

\begin{figure*}[h]
    \newcommand{\imh}{3.4cm}
    \setlength{\tabcolsep}{2pt}    
    \centering
    \begin{tabular}{cccccc}
    \includegraphics[height=\imh]{figures/shadow_removal/input_rgb_84.jpg} &
    \includegraphics[height=\imh]{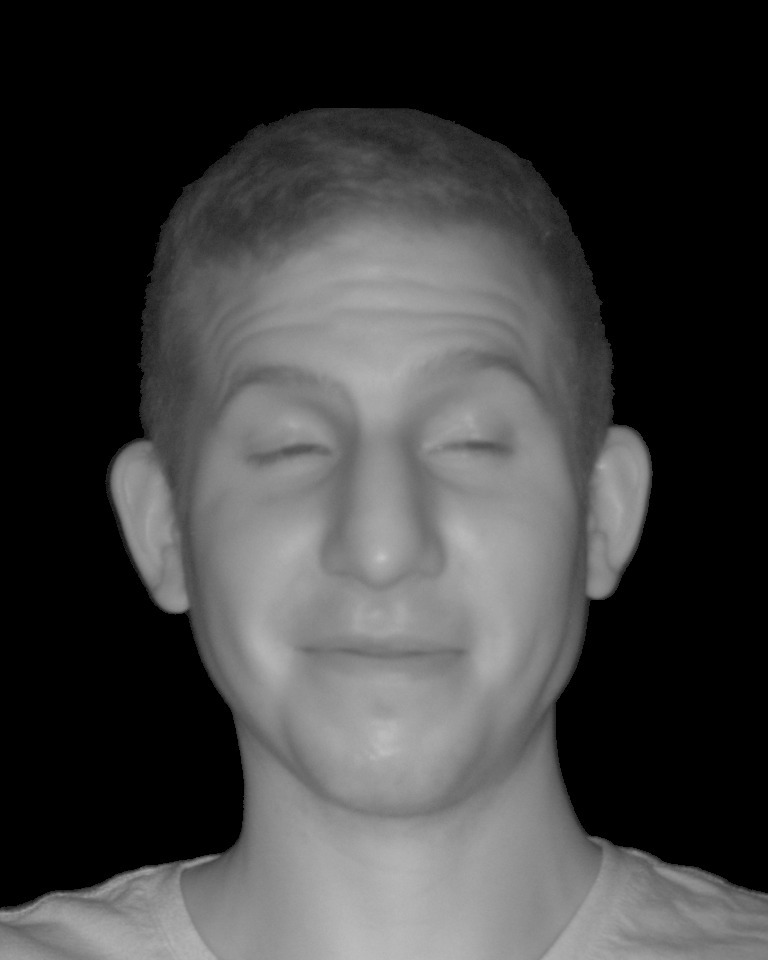} &
    \includegraphics[height=\imh]{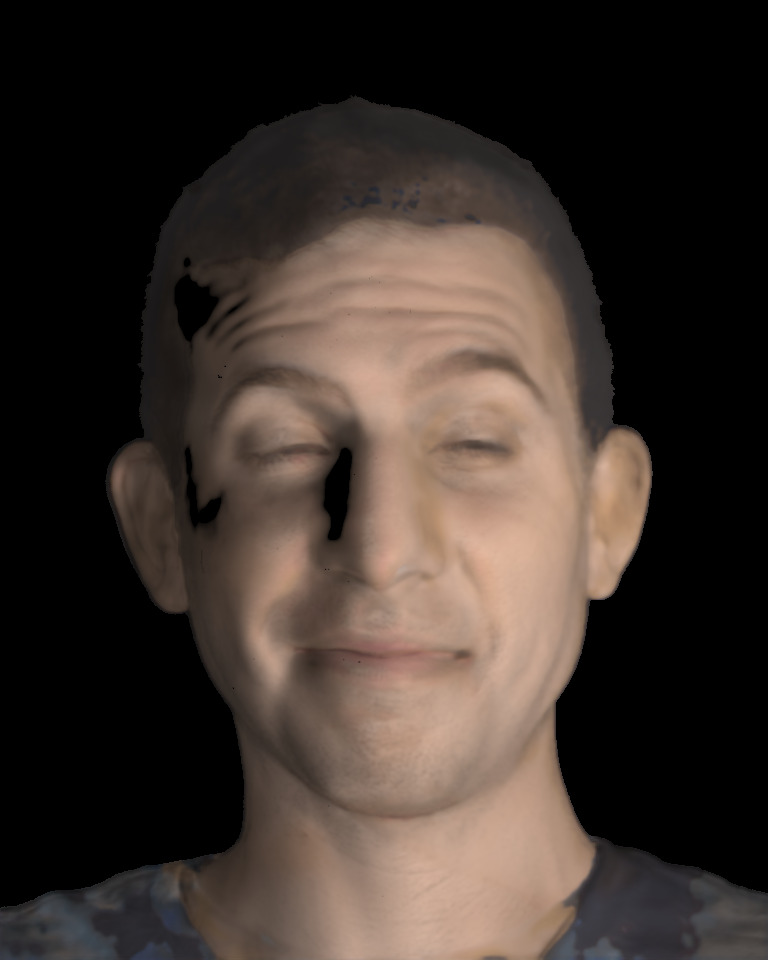} &
    \includegraphics[height=\imh]{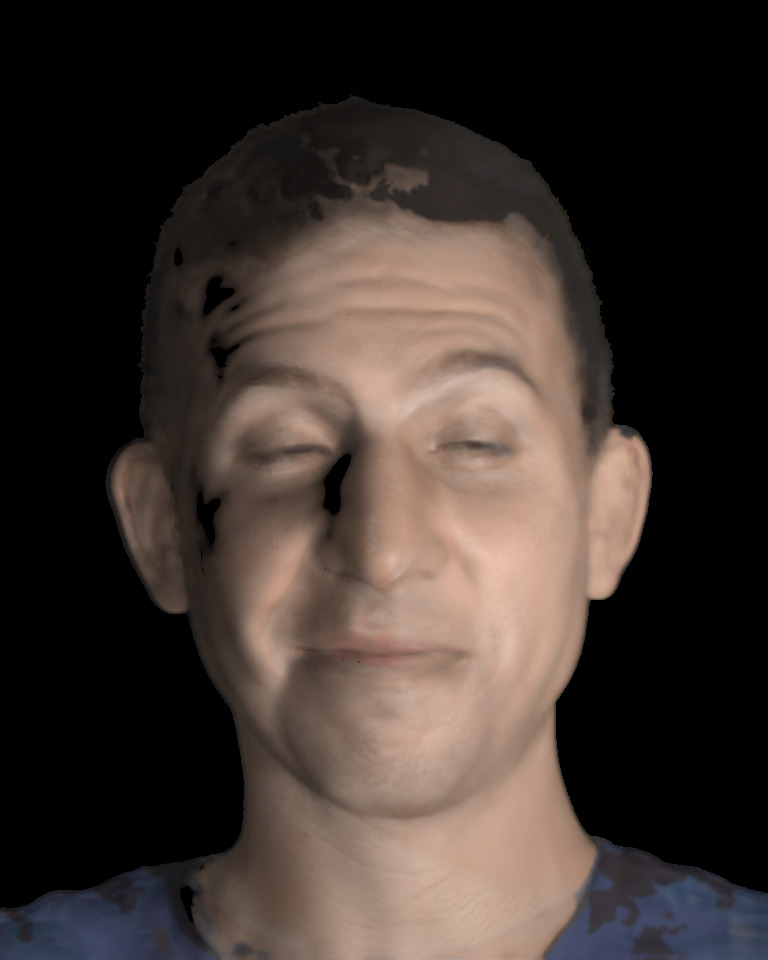} &
    \includegraphics[height=\imh]{figures/shadow_removal/bp_relit_84.jpg} &
    \includegraphics[height=\imh]{figures/shadow_removal/gt_rgb_84.jpg} \\
    \includegraphics[height=\imh]{figures/shadow_removal/input_rgb_138.jpg} &
    \includegraphics[height=\imh]{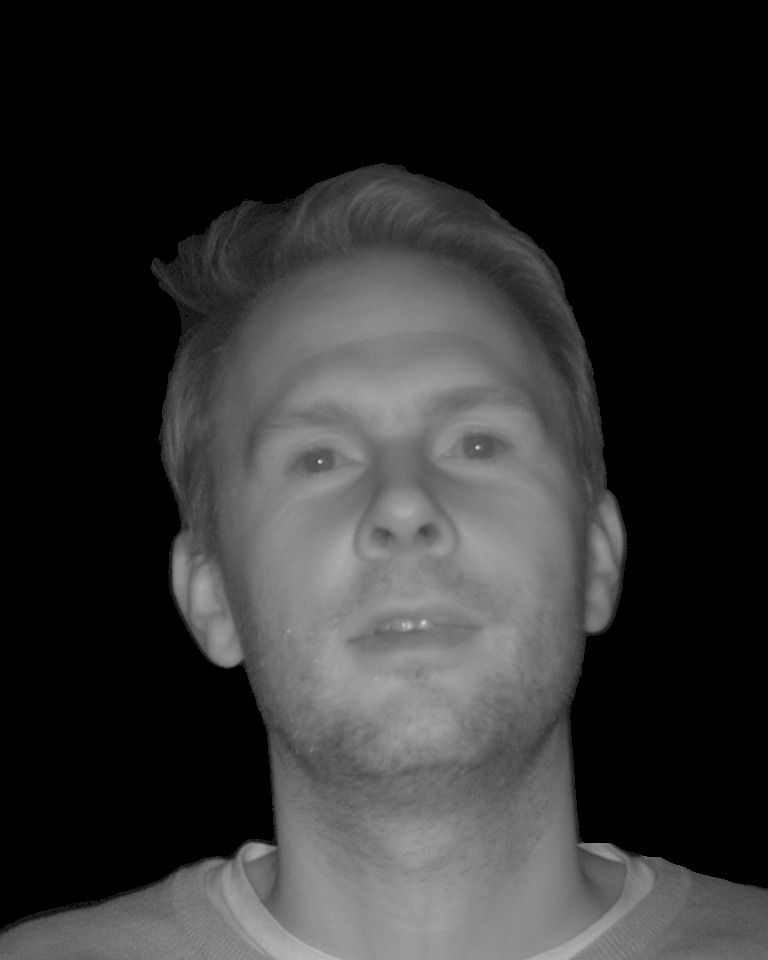} &
    \includegraphics[height=\imh]{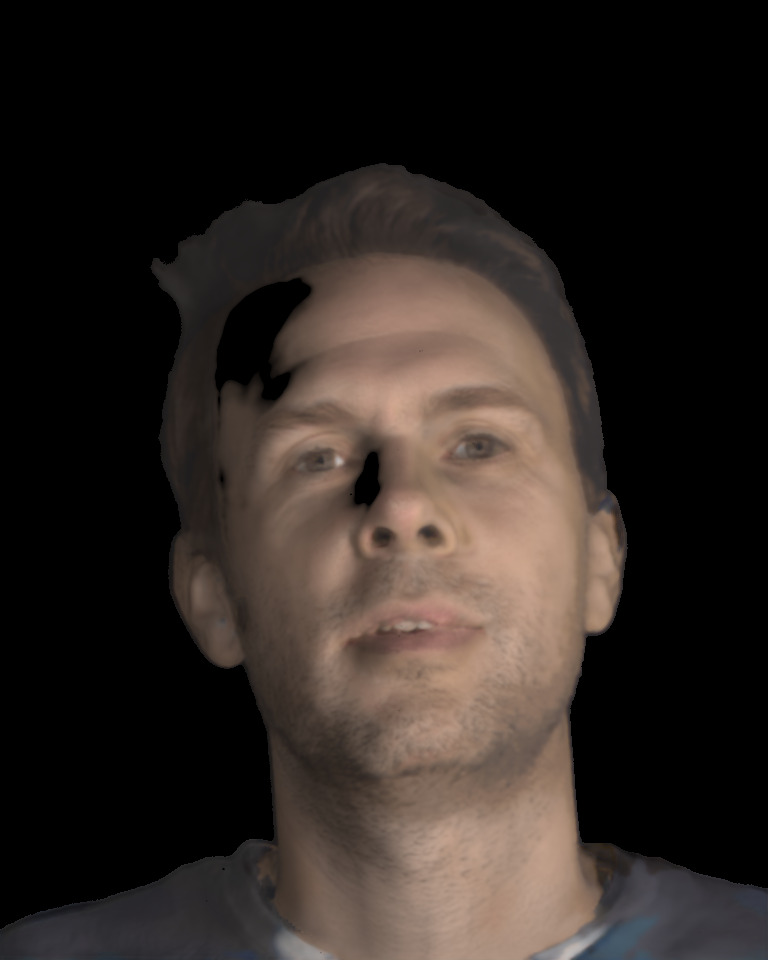} &
    \includegraphics[height=\imh]{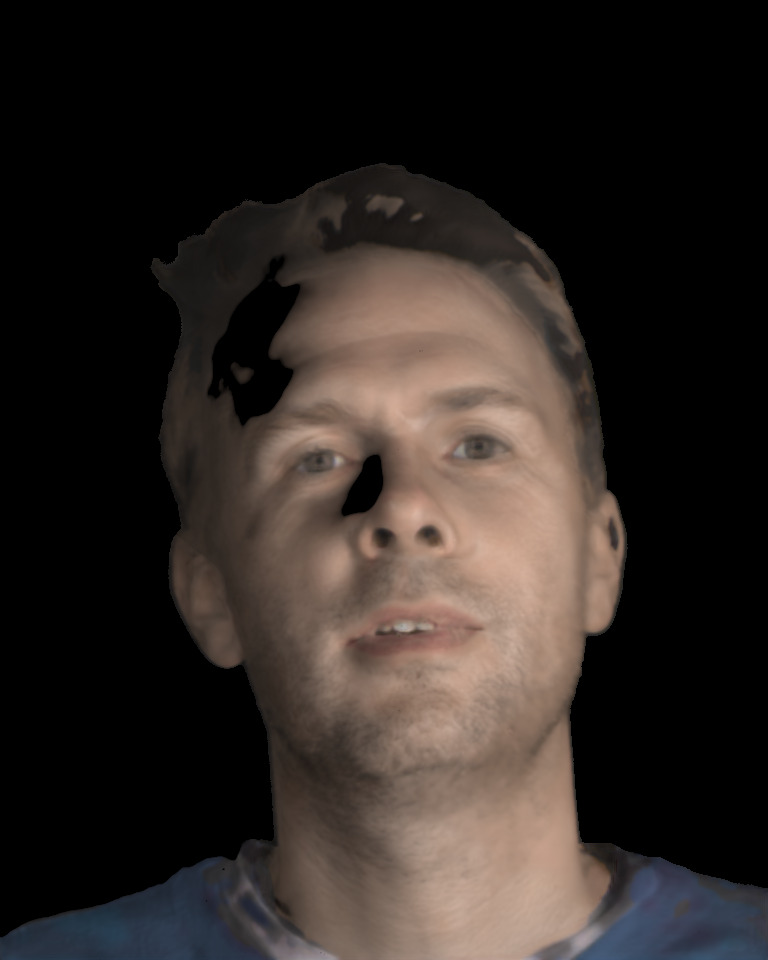} &
    \includegraphics[height=\imh]{figures/shadow_removal/bp_relit_138.jpg} &
    \includegraphics[height=\imh]{figures/shadow_removal/gt_rgb_138.jpg} \\
    \includegraphics[height=\imh]{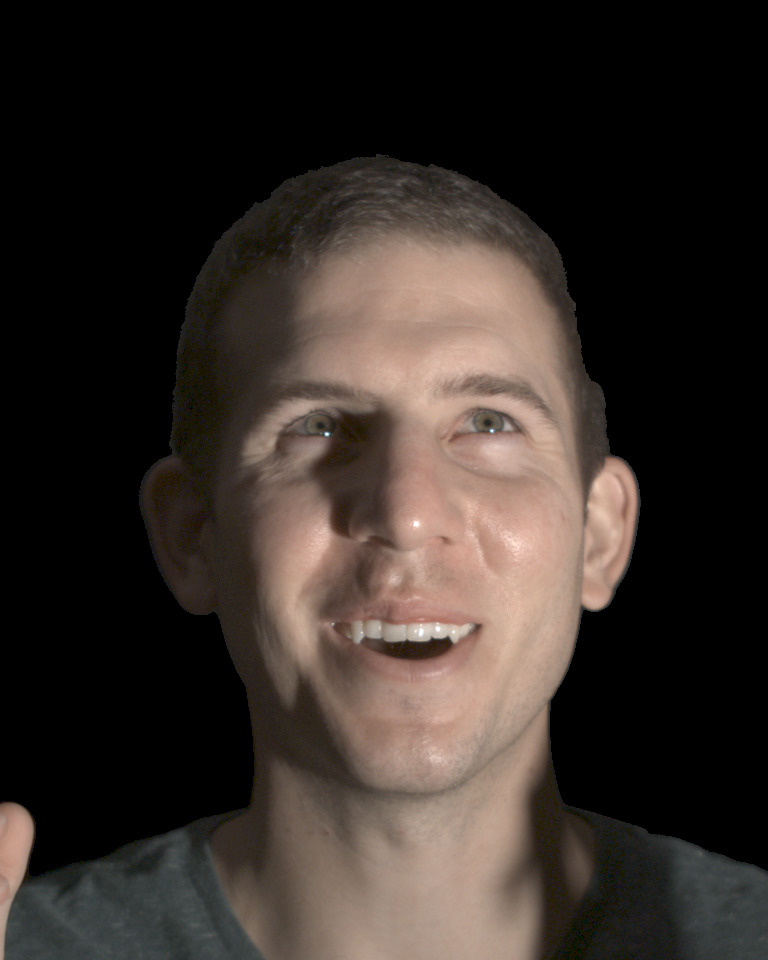} &
    \includegraphics[height=\imh]{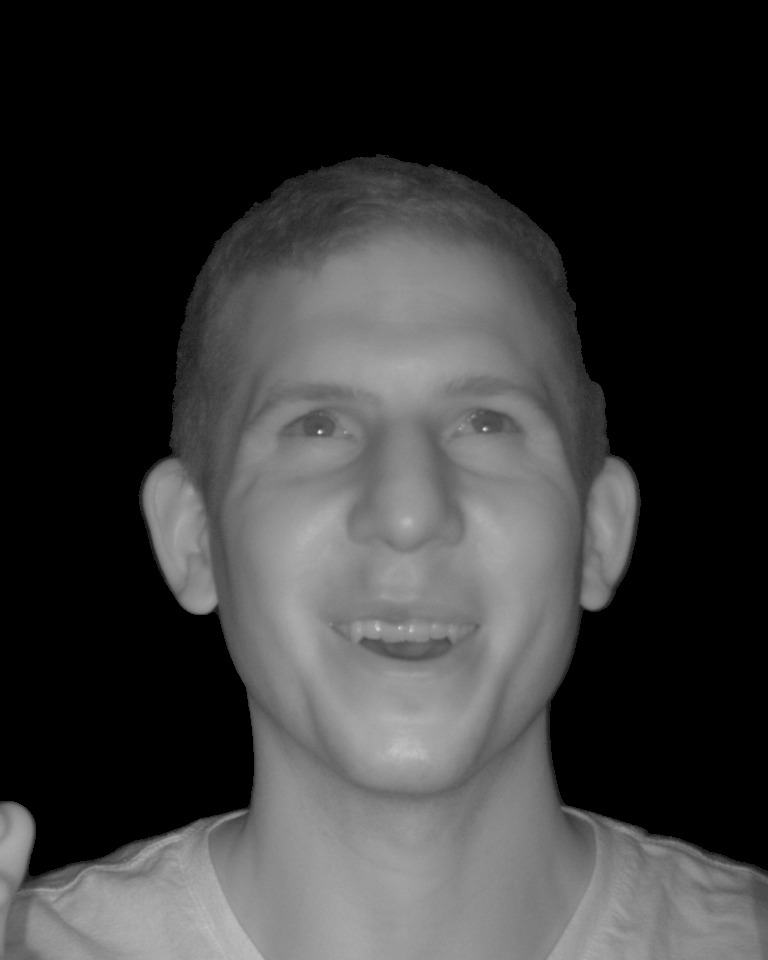} &
    \includegraphics[height=\imh]{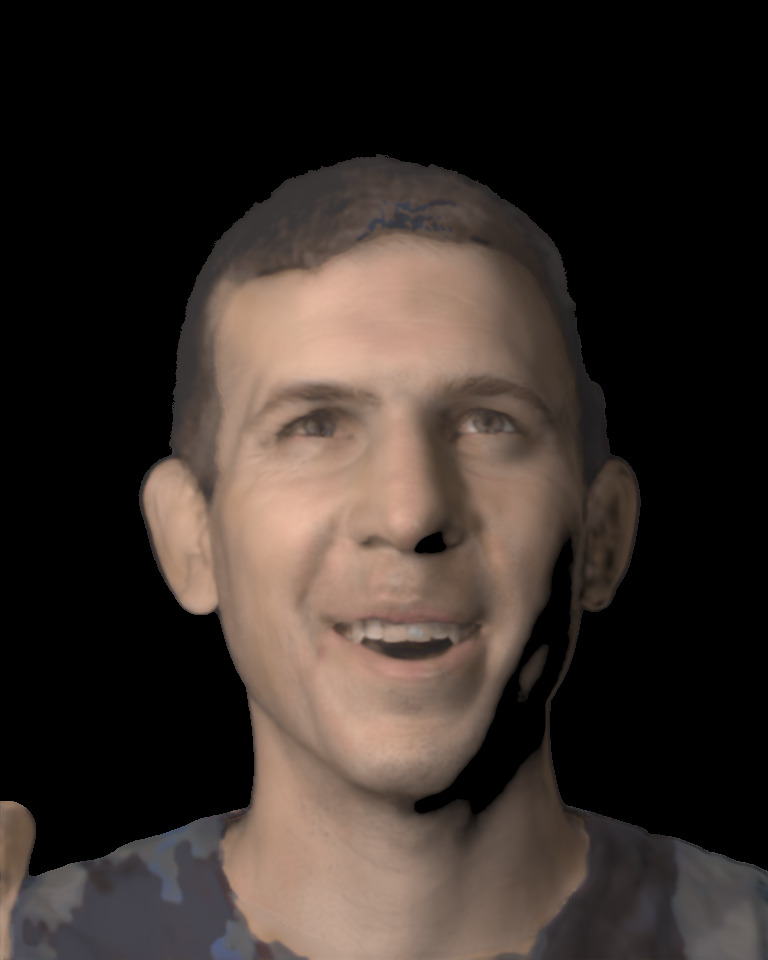} &
    \includegraphics[height=\imh]{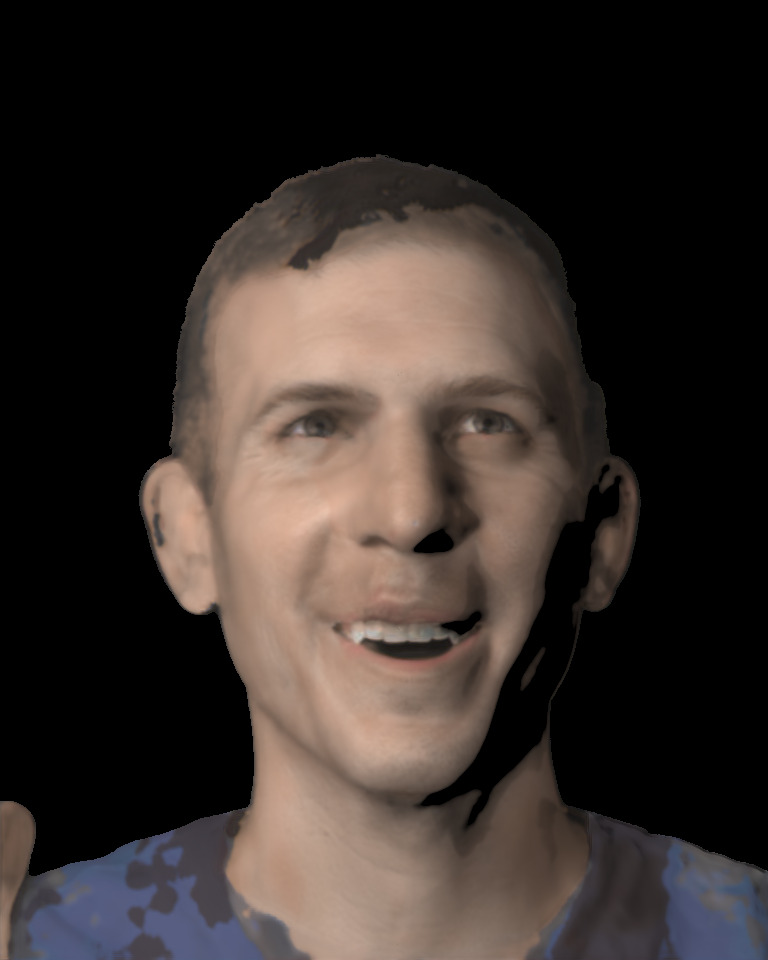} &
    \includegraphics[height=\imh]{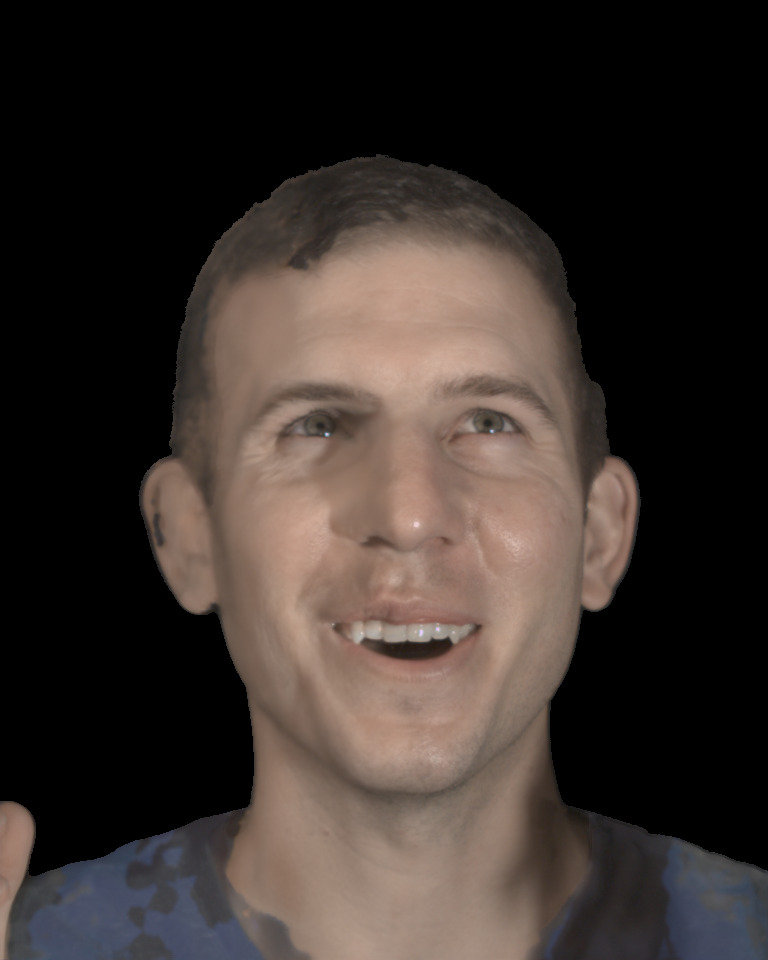} &
    \includegraphics[height=\imh]{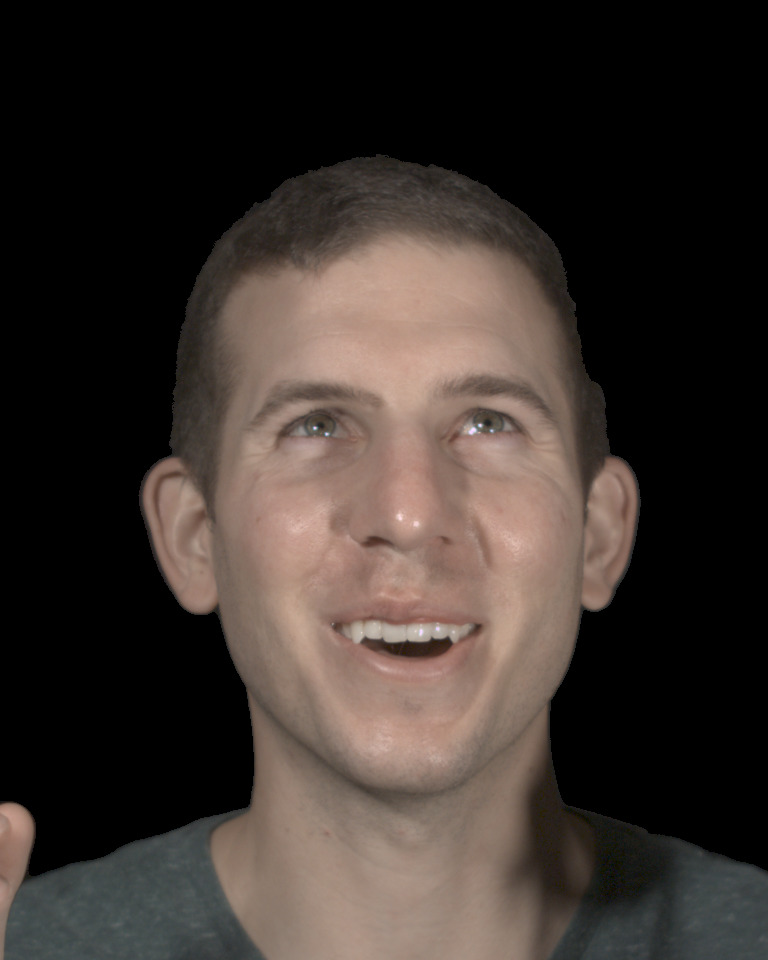} \\
    \small{RGB Input} & \small{NIR Input} & \small{Virtual Light} & \small{Virtual Light} & \small{Relit} & \small{Ground Truth} \\
    & & \small{(w/o Blinn-Phong)} & \small{(w/ Blinn-Phong)} & & \vspace{-5pt}
    \end{tabular}
    \vspace{-5pt}
    \caption{Our method can be used to simulate adding lights to a scene to fill in shadows. We show this virtual light image generated using a Lambertian reflectance model alone and using our full Lambertian + Blinn-Phong model, which produces more realistic highlights. When combined with the input RGB image (Relit) this approach compares favorably to ground truth.}
    \label{fig:shadow_removal_expand}
\end{figure*}

Figure~\ref{fig:shadow_removal_expand} expands on Figure~\ref{fig:shadow_removal} in the paper. Specifically it shows the supplemental rendered image that is combined with the input RGB image in order to brighten shadowed regions along the face. We also compare the difference between using a strictly Lambertian image formation model and our full Lambertian-plus-specular model in generating the supplemental image that is combined with the input. Note that our full model does a better job at reproducing specular highlights along the cheek and tip of the nose.

\section{Additional Video Results}

Please see our supplemental video for results and comparisons on image sequences.

\end{document}